\documentclass[twocolumn]{article}

\usepackage{geometry}
\usepackage{authblk}
\usepackage{cite}
\usepackage[utf8]{inputenc}
\usepackage{multirow}
\usepackage{graphicx}
\usepackage{outlines}
\usepackage[dvipsnames]{xcolor}
\usepackage{colortbl}
\usepackage{textcomp}
\usepackage{float}
\usepackage[normalem]{ulem}
\usepackage{comment}
\usepackage{amssymb}
\usepackage{mathtools}
\usepackage{booktabs}
\usepackage{tikz}
\usepackage{color}
\usepackage[titles]{tocloft}
\usepackage[mathscr]{eucal}

\definecolor{greyblue}{rgb}{0.5255,0.6039,0.6902}

\definecolor{lightgreyblue}{rgb}{0.6314,0.7373,0.7843}
\definecolor{lightgreygreen}{rgb}{0.6392,0.7255,0.6627}
\definecolor{lightgreyred}{rgb}{0.6824,0.6471,0.6863}
\definecolor{lightgreyyellow}{rgb}{0.8078,0.7529,0.6588}
\definecolor{lightgrey}{gray}{0.9}
\definecolor{lightestgrey}{gray}{0.95}

\newcommand{\code}[1]{\colorbox{lightgrey}{\texttt{#1}}}

\newcommand{\onlinecite}[1]{\hspace{-1 ex} \nocite{#1}\citenum{#1}} 

\let\OLDthebibliography\thebibliography
\renewcommand\thebibliography[1]{
  \OLDthebibliography{#1}
  \setlength{\parskip}{0pt}
  \setlength{\itemsep}{0pt plus 0.3ex}
}
  
\title{Phenomenological Model of \\Superconducting Optoelectronic Loop Neurons}
\author[1]{\Large{Jeffrey M. Shainline$^{*}$, Bryce A. Primavera, and Saeed Khan}
\\
\textit{\large{National Institute of Standards and Technology}}
\\
\vspace{-0.2em}
\textit{\large{325 Broadway, Boulder, CO, USA, 80305}}
\\
\small{$^*$jeffrey.shainline@nist.gov}
}
\date{October 17, 2022}

\begin{document}

\twocolumn[
\begin{@twocolumnfalse}
\maketitle
\begin{abstract}

Superconducting optoelectronic loop neurons are a class of circuits potentially conducive to networks for large-scale artificial cognition. These circuits employ superconducting components including single-photon detectors, Josephson junctions, and transformers to achieve neuromorphic functions. To date, all simulations of loop neurons have used first-principles circuit analysis to model the behavior of synapses, dendrites, and neurons. These circuit models are computationally inefficient and leave opaque the relationship between loop neurons and other complex systems. Here we introduce a modeling framework that captures the behavior of the relevant synaptic, dendritic, and neuronal circuits at a phenomenological level without resorting to full circuit equations. Within this compact model, each dendrite is discovered to obey a single nonlinear leaky-integrator ordinary differential equation, while a neuron is modeled as a dendrite with a thresholding element and an additional feedback mechanism for establishing a refractory period. A synapse is modeled as a single-photon detector coupled to a dendrite, where the response of the single-photon detector follows a closed-form expression. We quantify the accuracy of the phenomenological model relative to circuit simulations and find that the approach reduces computational time by a factor of ten thousand while maintaining accuracy of one part in ten thousand. We demonstrate the use of the model with several basic examples. The net increase in computational efficiency enables future simulation of large networks, while the formulation provides a connection to a large body of work in applied mathematics, computational neuroscience, and physical systems such as spin glasses.

\vspace{3em}
\end{abstract}
\end{@twocolumnfalse}
]

\section{\label{sec:introduction}Introduction}
A technological platform capable of realizing networks at the scale and complexity of the brains of intelligent organisms would be a tool of supreme scientific utility. Neuromorphic hardware based on conventional silicon microelectronics has a great deal to offer in this regard \cite{mead1989analog,indiveri2011neuromorphic,liu2014event,furber2016large,thakur2018large}. Yet challenges remain, primarily concerning bottlenecks in the shared communication infrastructure that must be employed to emulate the connectivity of biological neurons. Alternative hardware may bring new benefits, and we have argued elsewhere for the advantages of a superconducting optoelectronic approach \cite{shainline2017superconducting,shainline2018circuit,shainline2019superconducting,shainline2019fluxonic,shainline2021optoelectronic,primavera2021considerations}. In brief, light for communication enables high fan-out with low latency across spatial scales from a chip to a many-wafer system. Superconducting electronics provide single-photon detection coupled to high-speed, low-energy analog neuromorphic computational primitives.

While the components of these superconducting optoelectronic networks (SOENs) have been demonstrated \cite{buckley2017all,chiles2018design,mccaughan2019superconducting,khan2022superconducting}, full neurons have not. Prior to undertaking the effort and expense of realizing the requisite semiconductor-superconductor-photonic fabrication process, it is prudent to gain confidence that SOENs are indeed ripe for further investigation. This confidence can be gained through simulations of device, circuit, and system behavior using numerical simulations on digital computers. The constituent devices are most commonly modeled with circuit simulations carried out on a picosecond time scale to accurately capture the dynamics of Josephson junctions. Simulation of networks of large numbers of these neurons becomes computationally intensive. From the perspective of the neural system, the picosecond dynamics of the JJs are not of primary interest, and one would prefer to treat each synapse, dendrite, and neuron as an input-output device with a model that accurately captures the circuit dynamics on the nanosecond to microsecond time scales while not explicitly treating the picosecond behavior of the underlying circuit elements.
 
Here we introduce a phenomenological model of loop neurons and their constitutive elements that accurately captures the transfer characteristics of the circuits without solving the underlying circuit equations. Dendrites are revealed to be central to the system. Each dendrite is treated with a single, first-order ordinary differential equation (ODE) that describes the output of the element as a function of its instantaneous inputs and internal state. These equations take the form of a leaky integrator with a nonlinear driving term. The input to each dendrite is flux, and the output is an integrated current, which is coupled through a transformer into another dendrite. Synapses are dendrites with a closed-form expression for the input flux following each synapse event. The soma of a neuron is also modeled with the same equations as a dendrite with two modifications. First, when the output current reaches a specified level, threshold is reached, and an electrical signal is sent to a transmitter circuit, which produces a pulse of light. Second, a refractory dendrite triggered off this output is coupled back to the soma inhibitively to achieve a refractory period. 

By working at the phenomenological level, the time to simulate single dendrites is decreased by a factor of ten thousand while maintaining an accuracy of one part in ten thousand. The speed advantage grows with the size of the system being simulated and the duration of the simulation. The functional form of the speedup with size and duration has not been fully investigated, as solving the full system of circuit equations becomes very time consuming for even small systems. In addition, the model makes transparent the qualitative behavior of all components of the system, including the similarities to biological neurons as well as other physical systems such as Ising models, spin glasses, and pulse-coupled oscillators.

We begin by motivating the form of the model based on circuit considerations. We then describe the means by which the form of the driving term in the leaky integrator is obtained. Error is quantified by comparison with full circuit simulations, and convergence is investigated as a function of time step size. Numerical examples of dendrites, synapses, and neurons are presented. An example of a neuron with a dendritic arbor performing an image-classification task is given to illustrate the utility of the model. Further extensions to enable theoretical treatment of very large systems are discussed.

\section{\label{sec:overview_of_loop_neurons}Overview of Loop Neurons}
\begin{figure}[h!]
\includegraphics[width=8.6cm]{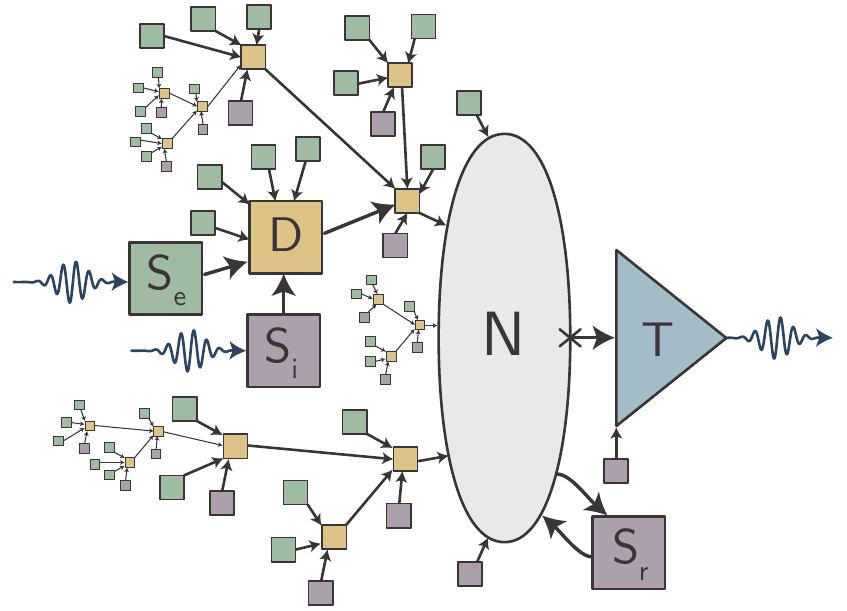}
\caption{\label{fig:schematic}Schematic of a loop neuron with an elaborate dendritic tree. The complex structure consists of excitatory and inhibitory synapses ($\mathsf{S_e}$ and $\mathsf{S_i}$) that feed into dendrites ($\mathsf{D}$). Each dendrite performs computations on the inputs and communicates the result to other dendrites for further processing or on to the cell body of the neuron ($\mathsf{N}$). The neuron cell body acts as the final thresholding stage, and when its threshold is reached, light is produced by the transmitter ($\mathsf{T}$), which is routed to downstream synaptic connections via photonic waveguides.}
\end{figure}
Research in superconducting optoelectronic networks of loop neurons aspires to realize artificial neural systems with scale and complexity comparable to the human brain. We have introduced the concepts of loop neurons in a number of papers \cite{shainline2017superconducting,shainline2018circuit,shainline2019superconducting,shainline2019fluxonic,shainline2021optoelectronic,primavera2021considerations}, and we have demonstrated many of the principles experimentally \cite{buckley2017all,chiles2017multi,chiles2018design,mccaughan2019superconducting,khan2022superconducting}. A schematic diagram of a loop neuron is shown in Fig.\,\ref{fig:schematic}, depicting the complex dendritic tree that appears central to the computations of loop neurons \cite{shainline2019fluxonic,primavera2021active}. In these neurons, integration, synaptic plasticity, and dendritic processing are implemented with inductively coupled loops of supercurrent. It is due to the prominent role of superconducting storage loops that we refer to devices of this type as loop neurons.

\begin{figure*}[h!]
\includegraphics[width=17.2cm]{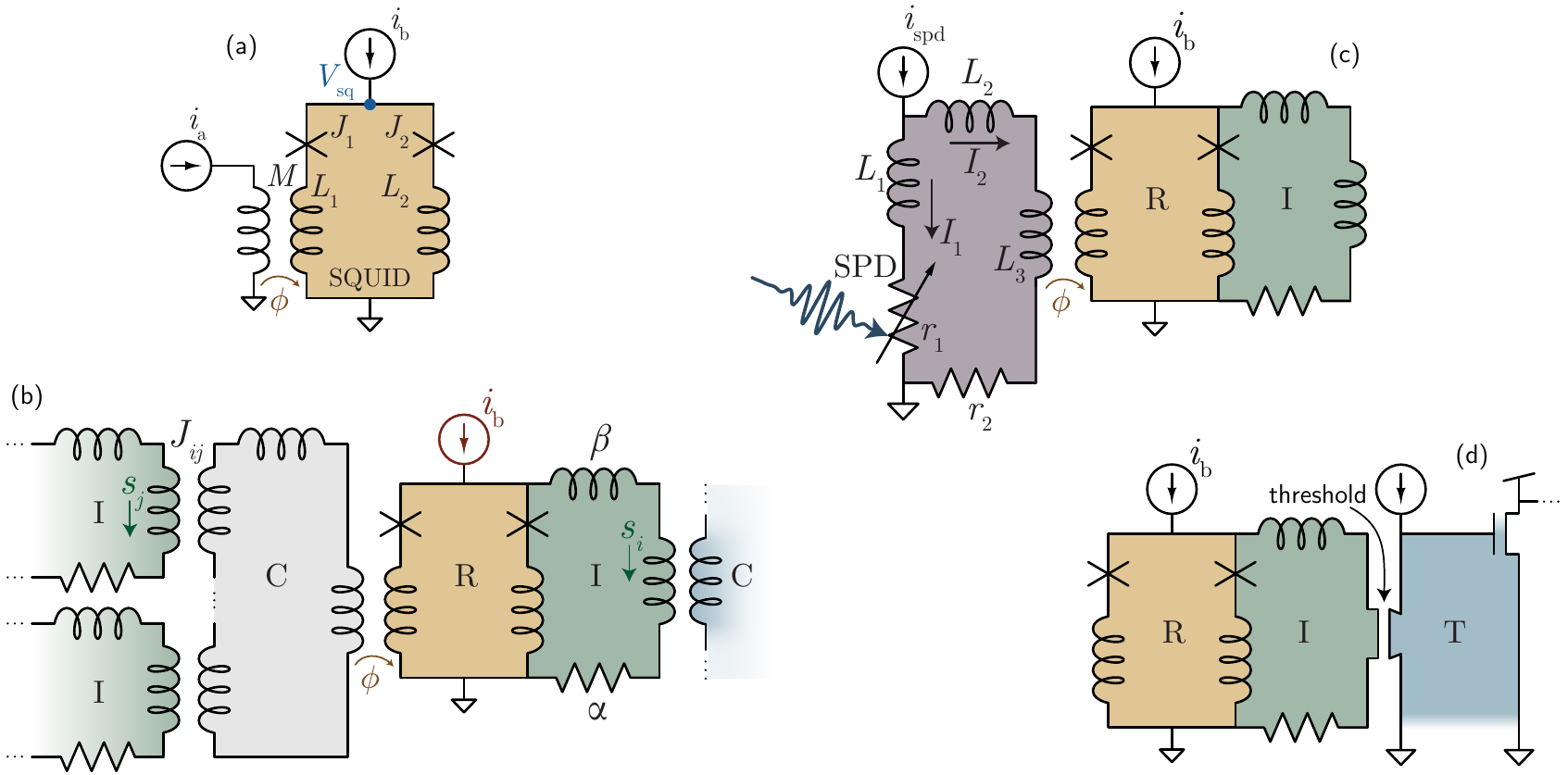}
\caption{\label{fig:circuits}Circuit diagrams. (a) A two-junction SQUID that forms the dendritic receiving loop. (b) A dendrite with receiving (R) and integrating (I) loop. The integrating loops of two other dendrites are input to a collection coil (C) that delivers flux to the receiving loop. (c) A synapse formed with an SPD input to a dendrite. (d) A soma realized as a dendrite with thresholding component in the integration loop and initial stage of the transmitter (T).}
\end{figure*} 
Operation of loop neurons is as follows. Photons from upstream neurons are received by a superconducting single-photon detector (SPD) at each synapse. Using a superconducting circuit comprising two Josephson junctions (JJs) coupled to the SPD, synaptic detection events are converted into an integrated supercurrent which is stored in a superconducting loop. The amount of current added to the integration loop during a photon detection event is determined by the synaptic weight. The synaptic weight is dynamically adjusted by another circuit combining SPDs and JJs, and all involved circuits are analog. When the integrated current of a given neuron reaches a (dynamically variable) threshold, an amplification cascade begins, resulting in the production of light from a waveguide-integrated semiconductor light emitter. The photons thus produced fan out through a network of dielectric waveguides and arrive at the synaptic terminals of other neurons where the process repeats.

The core active component of loop neurons is a circuit known as a superconducting quantum interference device (SQUID), which comprises two JJs in parallel. A circuit diagram is shown in Fig.\,\ref{fig:circuits}(a). The SQUID is a three-terminal device with a bias, ground, and an input which couples flux into the loop formed by the two JJs and inductors. For the present purpose, the flux input is the active signal. When the SQUID is current-biased below the critical current of the two JJs and the flux input is below a bias-dependent threshold, the SQUID remains superconducting, and the voltage across the device is zero. When the applied flux exceeds the threshold at a given bias point, the JJs will begin to emit a series of voltage pulses known as fluxons, and the time-averaged voltage across the device will become non-zero. 

To form a dendrite from a SQUID, we first ensure that it is biased below the critical current of the JJs so it is quiescent when no flux is applied. The output of the SQUID is captured by an $L$-$R$ loop that performs current integration with a leak [Fig.\,\ref{fig:circuits}(b)]. When sufficient flux is input to the SQUID to drive the JJs to begin producing voltage pulses, the pulses drive current into the $L$-$R$ loop, these pulses are summed in the inductor, and the accumulated signal leaks with the $L/R$ time constant of the loop. This configuration of a SQUID coupled to an $L$-$R$ loop is referred to as a dendrite, the SQUID that receives input flux is referred to as the receiving (R) loop, and the $L$-$R$ component is referred to as the integration (I) loop. The current stored in the integration loop produces the signal that will be communicated to other dendrites. To receive signal from multiple input dendrites, a passive collection coil (C) is used. A circuit diagram of a dendrite with two inputs, a collection coil, a receiving loop, and an integrating loop coupled to an output is shown in Fig.\,\ref{fig:circuits}(b). All coupling between dendrites is through magnetic flux communicated through mutual inductors. The use of transformers for this purpose mitigates cross talk and enables high fan-in \cite{primavera2021active}.

To form a synapse from a dendrite, we attach an SPD to the flux input to the receiving loop. This circuit is shown in Fig.\,\ref{fig:circuits}(c). When an SPD detects a photon, it rapidly switches from zero resistance to a large resistance, diverting the bias current to the other branch of the circuit. This current is coupled into the receiving loop of the dendrite as flux, driving the dendrite above threshold and adding current to the dendrite's integration loop.

To form a neuron from a dendrite, two modifications are required. First, the integration loop must be equipped with a thresholding element that drives a transmitter circuit to produce light when the integrated signal reaches this threshold. This thresholding element coupled to the transmitter is shown in Fig.\,\ref{fig:circuits}(d); it is referred to as a tron and is described in more detail in Sec.\,\ref{sec:neurons}. Second, an additional dendrite is attached to the neuron that provides negative feedback to achieve a refractory period (not shown in the circuit diagram for simplicity). The refractory period is a brief period of quiescence following a neuronal spike event. When the neuron's integration loop reaches threshold, this refractory dendrite is driven, accumulates signal in its integration loop, and this signal suppresses the state of the neuron's receiving loop. The time constant of the refractory dendrite establishes the refractory period of the neuron.

To summarize, a dendrite is a SQUID with a flux input adding current to a leaky integration loop. A synapse is a single-photon detector coupled to a dendrite. The soma of a neuron is a dendrite with a thresholding element in the integration loop as well as a second dendrite that provides feedback for refraction. To construct full loop neurons, many synapses are coupled into a dendritic arbor which feeds forward into the soma. The output of the soma is an optical pulse that couples light to a network of waveguides and delivers faint photonic signals to downstream synapses where they are received with single-photon detectors. This construction is illustrated schematically in Fig.\,\ref{fig:schematic}, and the basic circuits are shown in Fig.\,\ref{fig:circuits}. The current in a dendritic integration loop is analogous to the membrane potential of a biological dendrite \cite{dayan2005theoretical,gerstner2002spiking}, and these signals are the principal dynamical variables of the system. Inhibitory synapses can be achieved through mutual inductors with the opposite sign of coupling. Complex arbors with multiple levels of dendritic hierarchy can be implemented to perform various computations \cite{shainline2019fluxonic} as well as to facilitate a high degree of fan-in \cite{primavera2021active}.

We see that dendrites are central to loop neurons. The states of current in all dendritic integration loops specifies the state of the system. A phenomenological model of loop neurons must therefore capture the temporal evolution of these currents as well as the coupling between dendrites. With these concepts in mind we proceed to construct the model.

\section{\label{sec:dendrites}Dendrite Model}
\begin{figure}[tbh]
\includegraphics[width=8.6cm]{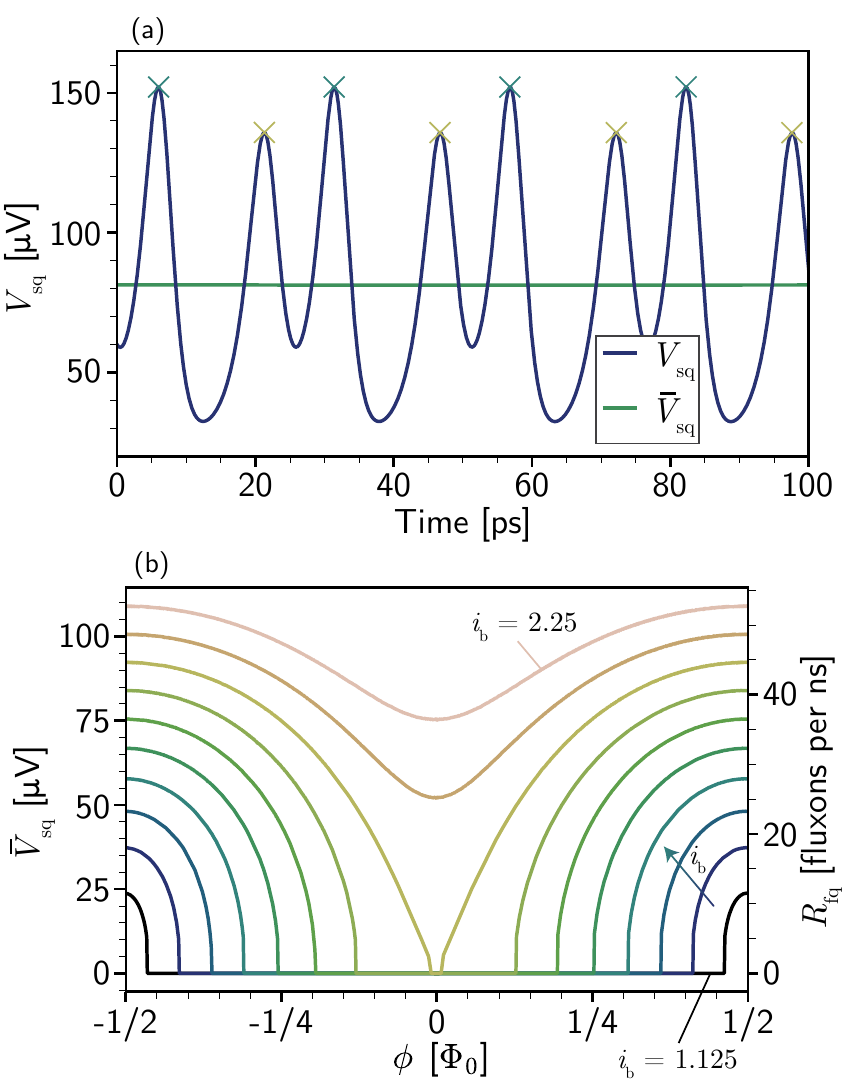}
\caption{\label{fig:squid}The response of a SQUID. (a) Time trace of a SQUID biased in the voltage state. Fluxon peaks are marked with crosses of different colors for the two JJs. The time average is also shown as calculated by taking the average of the time trace between two fluxons produced by the same junction. (b) The time-averaged voltage across the SQUID as a function of applied flux, $\phi$, normalized to the magnetic flux quantum, $\Phi_0$, for several values of normalized bias current, $i_\mathrm{b}$.}
\end{figure}
As mentioned in Sec.\,\ref{sec:overview_of_loop_neurons}, a SQUID is the primary active element of a dendrite. To motivate the phenomenological dendrite model we require a quantitative understanding of SQUID operation. The two-junction SQUID of Fig.\,\ref{fig:circuits}(a) with symmetrical inductances was modeled using a first-principles circuit model \cite{clarke2006squid}, and the results are shown in Fig.\,\ref{fig:squid}. Figure \ref{fig:squid}(a) shows a time trace of the voltage across the SQUID when it is in the voltage state. The peaks corresponding to fluxon production are evident, and the time-averaged voltage is also shown. While the voltage is a rapidly varying function of time on the picosecond scale, the time-averaged voltage is steady. 

This behavior is analyzed systematically in Fig.\,\ref{fig:squid}(b). The time-averaged voltage of a symmetric SQUID is plotted as a function of the applied flux for several values of the bias current, which has been normalized to the critical current of a single junction ($i_\mathrm{b} = I_\mathrm{b}/I_c$). The time-averaged voltage, plotted on the left $y$-axis, is proportional to the rate of flux-quantum production, plotted on the right $y$-axis. The relationship results from the single-valued nature of the superconducting wave function around the closed SQUID loop, which requires that
\begin{equation}
\label{eq:squid_voltage_integral}
\int_0^{t_\mathrm{fq}} V_\mathrm{sq}(t)\,dt = \Phi_0.
\end{equation}  
Equation \ref{eq:squid_voltage_integral} informs us that the time required to produce a single fluxon, $t_\mathrm{fq}$, is related to the voltage across the squid, $V_\mathrm{sq}$. For constant voltage, we obtain $r_\mathrm{fq} = 1/t_\mathrm{fq} = V_\mathrm{sq}/\Phi_0$. When viewed over time scales appreciably longer than $t_\mathrm{fq}$, it makes sense to speak of a rate of flux-quantum production, $r_\mathrm{fq}$, and this is the first element of our model: when driven to the active state, a dendrite will begin to produce fluxons, which carry current, and we can track this current by monitoring the rate of fluxon production while ignoring the picosecond dynamics by which the JJs produce the fluxons.

Several features of Fig.\,\ref{fig:squid}(b) are pertinent to the present study. First, for a given value of $i_\mathrm{b}$, a finite value of flux is required before the SQUID enters the voltage state. This provides a non-zero threshold that is relevant to dendritic computation. This threshold can be adjusted with the bias current. Second, the response is periodic in applied flux, with the period being $\Phi_0/2$, where $\Phi_0 = h/2e \approx 2\times 10^{-15}\mathrm{V}\cdot\mathrm{s} = 2\,\mathrm{mV}\cdot\,\mathrm{ps}$ is the magnetic flux quantum. To maintain a monotonic response, the applied flux must be limited to this value \cite{primavera2021active}. Third, if the inductors $L_1$ and $L_2$ are equal, the response of the SQUID is symmetric about $\Phi = 0$. These features will be discussed further as the study proceeds.

The second element of the model captures the integration and leak of the current generated when the SQUID is driven above threshold. These behaviors are accomplished by adding an $L$-$R$ loop to the output of the SQUID, as shown by the integration loop labeled I in Fig.\,\ref{fig:circuits}(b). The current integrated in this branch of the circuit is the quantity of interest for the dendrite. It is this quantitiy that will couple to other dendrites or the neuron cell body, and it is this quantity we wish to track with our phenomenological model. Dendrites comprising a receiving loop coupled to an integrating loop [Fig.\,\ref{fig:circuits}(b)] are referred to as RI dendrites. We know from elementary circuit theory that the $L$-$R$ loop will result in exponential decay of signal with a time constant of the dendritic integration loop given by $\tau_\mathrm{di} = L_\mathrm{di}/R_\mathrm{di}$. 

We can now write down a postulated expression for the signal $s$ stored in the integration loop of a dendrite:
\begin{equation}
\label{eq:main_equation__ode}
\beta \frac{d\,s}{d\tau} = r \left( \phi, s; i_\mathrm{b} \right) - \alpha\,s.
\end{equation}
Equation \ref{eq:main_equation__ode} states that the signal $s$ grows in time due to the driving term, which is the rate of flux quantum production, denoted by $r$. This function $r$ depends on the applied flux to the receiving loop of the dendrite, $\phi = \Phi/\Phi_0$, as well as of the signal present in the integration loop, $s$. The rate also depends on the bias current, $i_\mathrm{b}$, as a parameter that throughout this work is assumed to be held constant over times much longer than the inter-fluxon interval. In Eq.\,\ref{eq:main_equation__ode} we have formulated the model in dimensionless units, where $s = I_\mathrm{di}/I_c$, and $I_\mathrm{di}$ is the current present in the dendritic integration loop. $\beta = 2\,\pi\,I_c\,L_\mathrm{di}/\Phi_0$ is a dimensionless parameter that quantifies the inductance of the loop, and $\alpha = R_\mathrm{di}/r_\mathrm{j}$, where $r_\mathrm{j}$ is the shunt resistance of each JJ in the resistively and capacitively shunted junction model \cite{van1998principles,kadin1999introduction,tinkham2004introduction,clarke2006squid}. The signal $s$ decays at a rate related to $\alpha$ and $\beta$. Specifically, the time constant for decay is given by $\tau_\mathrm{di} = L_\mathrm{di}/R_\mathrm{di} = \beta/\omega_c\alpha$, where $\omega_c$ is the Josephson characteristic frequency discussed in Appendix A, and we include the subscript on $\tau_\mathrm{di}$ to refer to the dendritic integration loop and distinguish that quantity from the dimensionless time variable entering Eq.\,\ref{eq:main_equation__ode}. Equation \ref{eq:main_equation__ode} is a leaky integrator ODE. The drive term is the rate of flux quantum production, and the leak term gives simple exponential decay, as expected from an $L$-$R$ circuit.

\begin{figure*}[tbh]
\includegraphics[width=17.2cm]{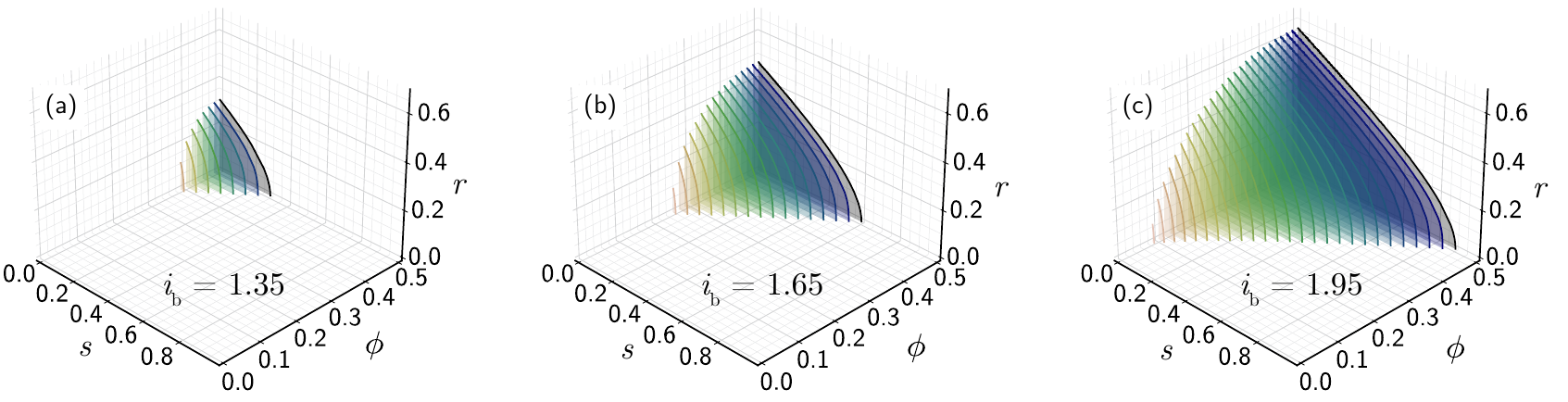}
\caption{\label{fig:rate_arrays__ri}Rate arrays for the dendrite with a receiving and integrating loop for three values of $i_\mathrm{b}$. (a) The normalized bias current $i_\mathrm{b} = 1.35$. (b) $i_\mathrm{b} = 1.65$. (c) $i_\mathrm{b} = 1.95$.}
\end{figure*}
Dendrites are coupled to each other through flux. The coupling flux from dendrites indexed by $j$ to dendrite $i$ is given by
\begin{equation}
\label{eq:main_equation__coupling}
\phi_i = \sum_{j=1}^n\,J_{ij}\,s_j,
\end{equation}
where $J_{ij}$ is a coupling term proportional to the mutual inductance that includes contributions from all the transformers present on the collection coil in Fig.\,\ref{fig:circuits}(b). Equation \ref{eq:main_equation__coupling} shows that coupling between dendrites is due to the signal in the integration loop of one dendrite being communicated as flux into the receiving loop of a subsequent dendrite. The signal in the subsequent dendrite is then obtained through the evolution of Eq.\,\ref{eq:main_equation__ode} with the flux from the first dendrite providing the flux $\phi$ in the function $r$ and the signal from the second dendrite providing the $s$ term. The $i_\mathrm{b}$ term entering $r$ refers to the bias on the second dendrite and is treated here as a parameter rather than a dynamical variable. Further details regarding the derivation of these expressions is given in Appendix \ref{apx:phenomenological_model_dimensionless_units}.

Equations \ref{eq:main_equation__ode} and \ref{eq:main_equation__coupling} constitute the phenomenological model of a dendrite. A neuron or network can be simulated by solving these coupled equations for all dendrites in the system. However, we have not specified the form for the rate function, $r \left( \phi, s; i_\mathrm{b} \right)$, which is central to the model.

We have arrived at Eq.\,\ref{eq:main_equation__ode} as a postulate; this expression is not directly obtained from the underlying circuit equations. The postulate is that there will be a function $r \left( \phi, s; i_b \right)$ such that Eq.\,\ref{eq:main_equation__ode} provides an accurate account of the signal present in a dendrite's integration loop under the circumstances of interest for loop neurons, provided we only inquire about the signal over time scales appreciably longer than the inter-fluxon interval, which is on the order of 10\,ps. We aim to interrogate dendrites on time scales of 100\,ps or longer, with neuron and network activity of interest on time scales from nanoseconds to the longest timescales that can be simulated under the limits of computational resources. 

Our procedure for obtaining $r \left( \phi, s; i_\mathrm{b} \right)$ is as follows. A dendrite with a SQUID as a receiving loop and an $L$-$R$ branch as an integrating loop is numerically modeled with the circuit equations given in Appendix \ref{apx:circuit_equations_RI}. A constant value of flux is applied to the receiving loop, and the rate of flux-quantum production is monitored as a function of time while current accumulates in the integration loop. For these simulations, the resistance of the integration loop is set to zero. This procedure is repeated for many values of $\phi$ and $i_\mathrm{b}$ to obtain what we refer to as ``rate arrays'', which are shown in Fig.\,\ref{fig:rate_arrays__ri}, where $r \left( \phi, s; i_\mathrm{b} \right)$ is plotted as a function of $\phi$ and $s$ for three values of $i_\mathrm{b}$. Here we work in dimensionless units, so the units of $r$ are fluxons generated per unit of dimensionless time, $\tau$, which is related to the JJ characteristic frequency (Appendix \ref{apx:circuit_equations_RI}). It can be seen that the value of $r \left( \phi, s; i_\mathrm{b} \right)$ is monotonically increasing with $\phi$ over the range considered here, while accumulation of $s$ decreases the rate of flux-quantum production. This decrease is because addition of current to the integration loop diverts the bias away from the SQUID, so the voltage is decreased, and the rate is reduced in accordance with Eq.\,\ref{eq:squid_voltage_integral}. When sufficient signal is accumulated in the integration loop, the rate of flux quantum production drops to zero, and we say the loop is saturated. 

For a small value of $i_\mathrm{b}$ [Fig.\,\ref{fig:rate_arrays__ri}(a)] a large amount of flux is required to drive the dendrite above threshold to the active state, and a small signal $s$ is present at saturation. As $i_\mathrm{b}$ is increased [Figs.\,\ref{fig:rate_arrays__ri}(b) and (c)] the threshold is reduced, and the saturation level is increased. The qualitative shape of $r$ for different values of $i_\mathrm{b}$ is similar, and the surfaces for smaller $i_\mathrm{b}$ are seen to fit inside those for larger $i_\mathrm{b}$. For this reason we refer to these surfaces as $r$-shells, with each shell corresponding to a fixed value of $i_\mathrm{b}$. Network adaptation, training, and learning consist in finding the values of $i_\mathrm{b}$ that achieve the desired input-output relationships for a given computational or control task.

While Fig.\,\ref{fig:rate_arrays__ri} shows $r$ for $0 \le \phi \le 1/2$, these surfaces are symmetric about zero for the circuit of Fig.\,\ref{fig:circuits}(b) with symmetric inductances in the SQUID [as is also the case in the SQUID response of Fig.\,\ref{fig:squid}(b)], and they are periodic in normalized flux $\phi$ with period of unity (period of $\Phi_0$ in SI units). One consequence of this response is that it may be necessary to restrict the applied flux to $\phi \le 1/2$ to retain a monotonic response \cite{primavera2021active}, and if inhibition is applied to a dendrite ($\phi < 0$) it may be necessary to limit this flux below the threshold for activity so that inhibition does not drive the dendrite to the active state. These concepts are discussed further in Sec.\,\ref{sec:discussion}.

\begin{figure}[tbh]
\includegraphics[width=8.6cm]{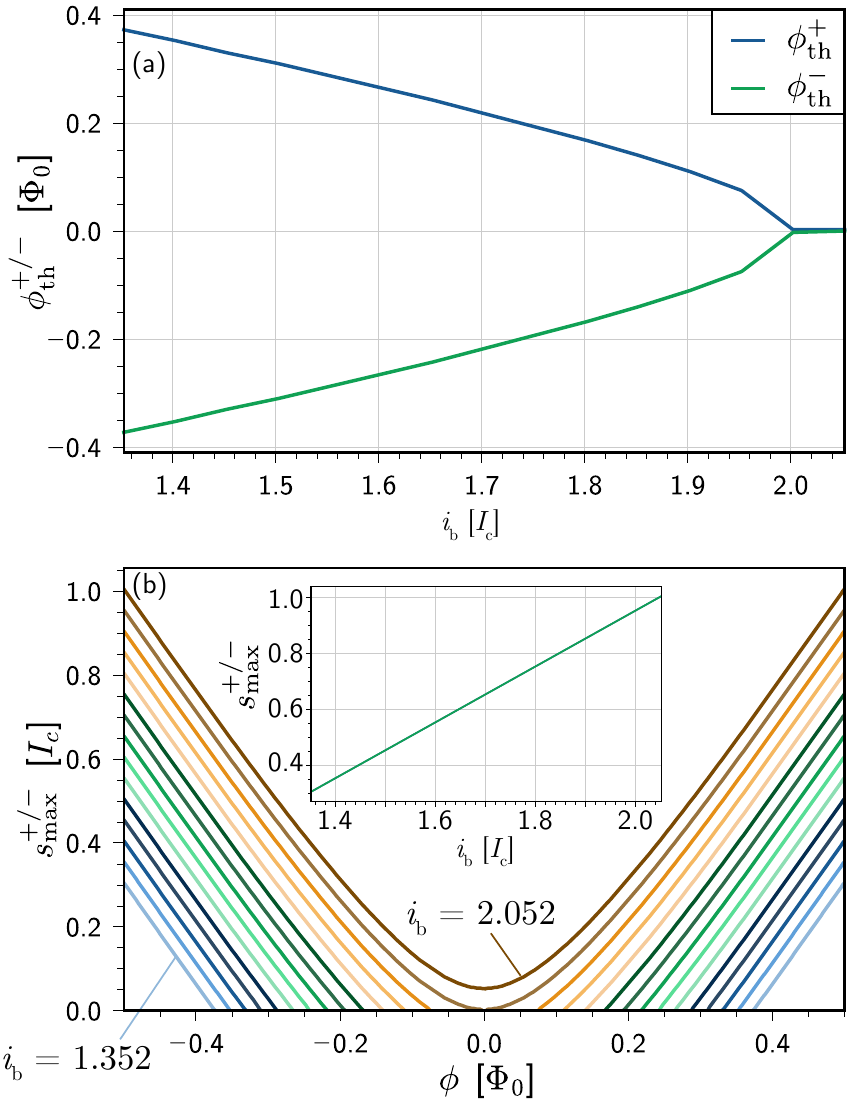}
\caption{\label{fig:dend__ri__thresholds__saturations}Response parameters of the RI dendrite extracted from the rate arrays. (a) The thresholds to positive and negative applied flux as a function of the dendritic bias current. (b) The value of saturation signal as a function of the normalized flux applied to the receiving loop for several values of the normalized bias current, $i_\mathrm{b}$.}
\end{figure}
From the rate arrays we obtain several quantities of interest that facilitate dendrite design. In particular, the value of $\phi$ for which $r$ becomes non-zero as a function of $i_\mathrm{b}$ determines the dendrite threshold. This function is shown in Fig.\,\ref{fig:dend__ri__thresholds__saturations}(a) for both positive and negative flux. These traces can be leveraged to anticipate when a dendrite will become active as well as the maximum value of inhibitory flux that can be applied before activity results, which is necessary in design of the refractory dendrite that quenches activity of a soma. The saturation values of $s$ as a function of applied flux and bias current are also important for anticipating the maximum value of signal that a dendrite will accumulate, and therefore the maximum value of flux it will couple into another dendrite. We refer to this value as $s_\mathrm{max}$, and Fig.\,\ref{fig:dend__ri__thresholds__saturations}(b) shows its behavior as a function of both positive and negative applied flux ($s_\mathrm{max}^{+/-}$). The inset shows the value versus $i_\mathrm{b}$ for the maximum absolute value of applied flux, $|\phi| = 1/2$.

Obtaining the rate arrays is somewhat computationally intensive, requiring a large number of numerical simulations of the dendrite circuit. However, these computations only need to be carried out once. The rate arrays are then accessed by the phenomenological model as a look-up table, and the same arrays are used for all values of the integration loop inductance and leak rate. Using the same arrays for all values of inductance is possible because in all cases the inductance of the integration loop is much larger than that of the receiving loop, so the fraction of current bias that initially goes into the integration branch of the circuit is small in all cases of interest here. If properties of the receiving loop---the SQUID---are changed, new rate arrays will need to be calculated. Examples include changing the value of $\beta_\mathrm{c}$ of the JJs or changing the asymmetry of the inductances. However, there appears to be little advantage from employing SQUIDs with a variety of $\beta_\mathrm{c}$ values. Leveraging different asymmetry designs may be fruitful, but a small number of variants will likely suffice. We anticipate a symmetric SQUID and a highly asymmetric SQUID will be the two cases of primary interest.

\begin{figure}[tbh]
\includegraphics[width=8.6cm]{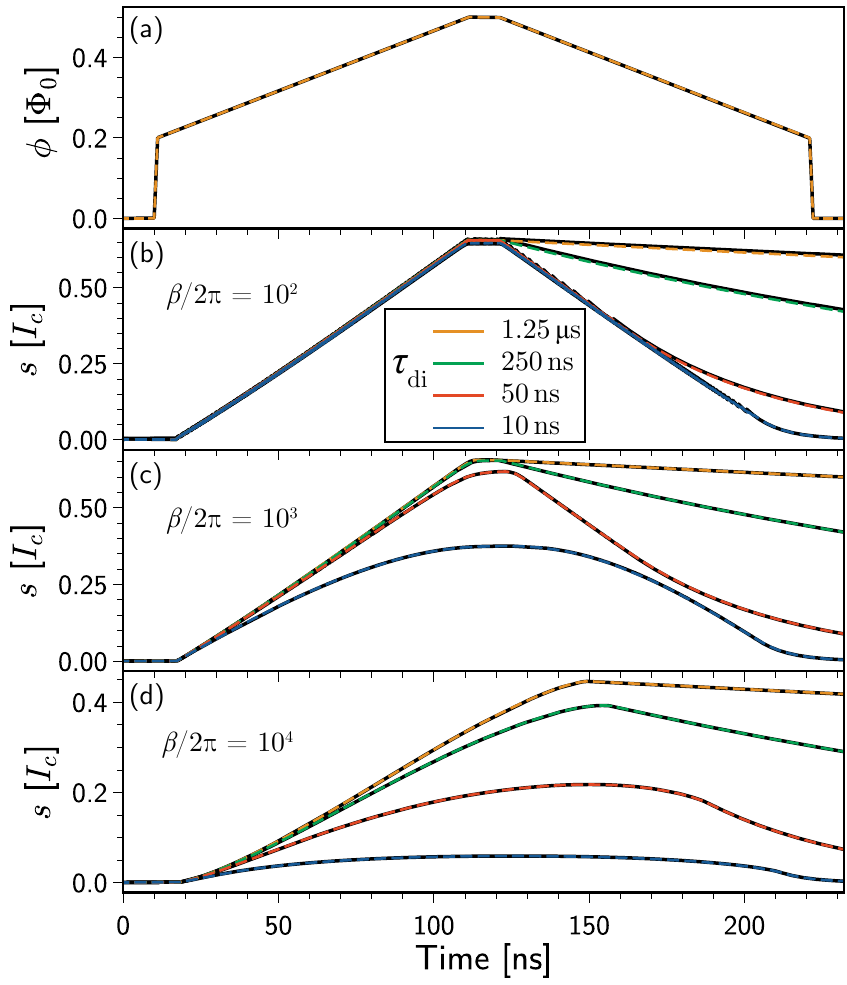}
\caption{\label{fig:dend__ode_comparison__lin_ramp__100ps}Comparing the phenomenological model to circuit equations in the case of a linearly ramping drive. The time step for the phenomenological model was 100\,ps. (a) The flux drive signal. (b) The smallest integration loop with $\beta/2\pi = 10^2$. (c) Integration loop with $\beta/2\pi = 10^3$. (d) $\beta/2\pi = 10^4$. In (a)-(d) the black traces show the values for the circuit simulations while the dashed colored traces show the values for the phenomenological model. The error of the drive signal was $\chi^2_\mathrm{drive} < 5\times 10^{-11}$ for all cases shown here.}
\end{figure}
Equipped with the phenomenological model (Eqs.\,\ref{eq:main_equation__ode} and \ref{eq:main_equation__coupling}) as well as the rate arrays, we can compare the model performance to explicit solution of the circuit equations. Throughout this work, we numerically solve systems described by Eqs.\,\ref{eq:main_equation__ode} and \ref{eq:main_equation__coupling} using a forward Euler method to facilitate computational speed. In discrete form, Eq. \ref{eq:main_equation__ode} becomes
\begin{equation}
\label{eq:main_equation__ode__discrete}
s_{\tau+1} = s_\tau\left(1-\Delta\tau\frac{\alpha}{\beta}\right) + \frac{\Delta\tau}{\beta}r(\phi,s;i_\mathrm{b}).
\end{equation}
We solve the circuit equations given in Appendix \ref{apx:circuit_equations_RI} using a Runge-Kutta method implemented with the Python SciPy function \code{solve\_ivp} for initial-value problems, which uses an adaptive time mesh.

As a first test, Fig.\,\ref{fig:dend__ode_comparison__lin_ramp__100ps} compares the phenomenological model to the circuit equations in the presence of a linear ramp function as the applied flux. The form of the applied flux into the receiving loop ($\phi$) is shown in Fig.\,\ref{fig:dend__ode_comparison__lin_ramp__100ps}(a). The rate arrays are obtained only for discrete values of applied flux (200 values between 0 and $1/2$), so this applied drive is relevant to quantify performance when the applied flux takes intermediate values between those explicitly present in the rate arrays. No interpolation was used. At each time step, the three inputs to the rate array ($\phi$, $s$, and $i_\mathrm{b}$) are rounded to the closest value for which the rate array has been evaluated, and that value of $r$ is used in Eq.\,\ref{eq:main_equation__ode__discrete} at that time step. The dendrite signals are shown in Fig.\,\ref{fig:dend__ode_comparison__lin_ramp__100ps}(b)-(d) for three values of the inductance parameter, $\beta$, from $\beta/2\pi = 10^2 - 10^4$. For each value of $\beta$, four time constants are considered from 10\,ns to 1.25\,\textmu s. The value of $\beta/2\pi$ quantifies the number of fluxons that can be accommodated in the storage loop, and the time constant is converted to the dimensionless parameter $\alpha$ for simulation in the dimensionless model. All data from the phenomenological model in Fig.\,\ref{fig:dend__ode_comparison__lin_ramp__100ps} were obtained using a time step of 100\,ps (converted to dimensionless time for Eq.\,\ref{eq:main_equation__ode__discrete}). The circuit model time traces are shown as black lines, while the traces obtained with the phenomenological model are shown as various dashed colored lines.

\begin{figure}[tbh]
\includegraphics[width=8.6cm]{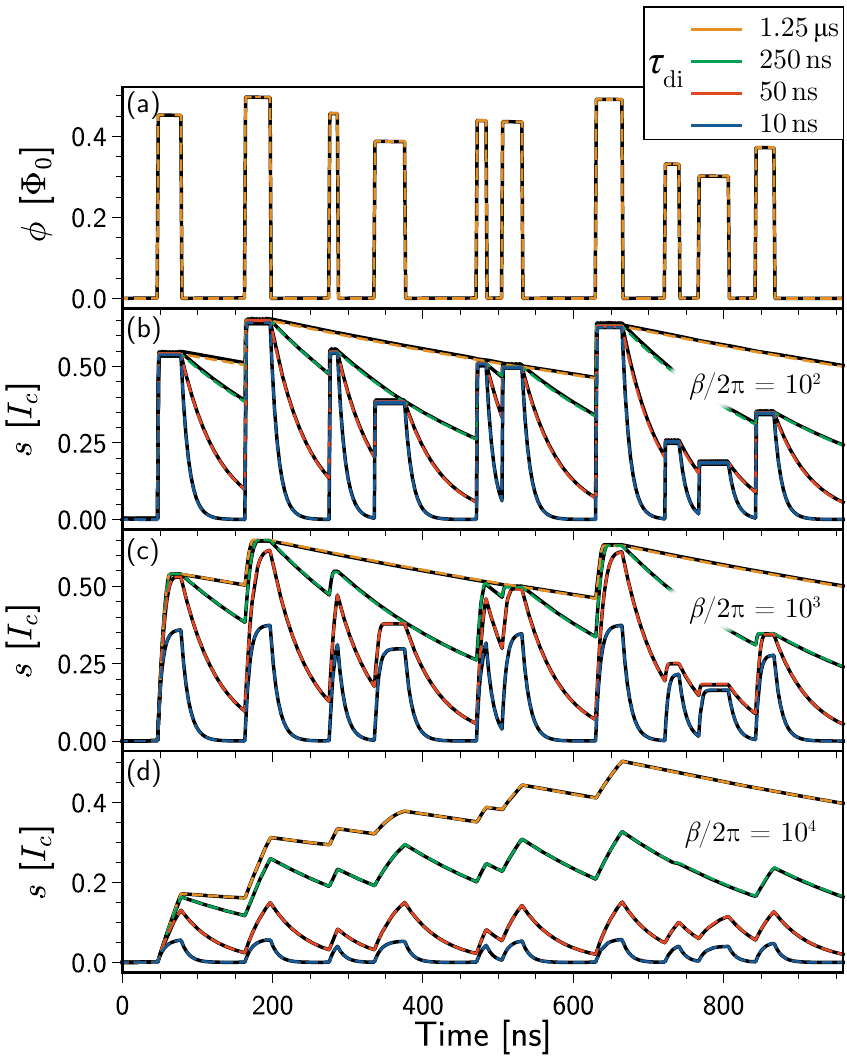}
\caption{\label{fig:dend__ode_comparison__sqr_pulse__100ps}Comparing the phenomenological model to circuit equations in the case of a random sequence of square pulses. The time step for the phenomenological model was 100\,ps. (a) The flux drive signal. (b) The smallest integration loop with $\beta/2\pi = 10^2$. (c) Integration loop with $\beta/2\pi = 10^3$. (d) $\beta/2\pi = 10^4$. The error of the drive signal was $\chi^2_\mathrm{drive} = 5.03\times 10^{-5}$ for all cases shown here.}
\end{figure}
Accuracy of the phenomenological model is quantified with a $\chi^2$ of the form
\begin{equation}
\label{eq:chi_squared}
\chi^2 = \frac{\sum_{i = 1}^{n_t^\mathrm{soen}-1} \left| s_\mathrm{soen}\left( t_i^\mathrm{soen} \right) - s_\mathrm{ode}^\mathrm{interp} \left( t_i^\mathrm{soen} \right) \right|^2 \Delta t_i^\mathrm{soen}}{\sum_{i = 1}^{n_t^\mathrm{ode}-1} \left|  s_\mathrm{ode} \left( t_i^\mathrm{ode} \right) \right|^2 \Delta t_i^\mathrm{ode}}.
\end{equation}
The subscript or superscript ``soen'' refers to the phenomenological model, while ``ode'' refers to the first-principles circuit model of Appendix \ref{apx:circuit_equations_RI}. In Eq.\,\ref{eq:chi_squared}, $n_t^\mathrm{soen}$ is the number of time steps in a given SOEN simulation, $s_\mathrm{soen}$ is the state quantity obtained with the SOEN phenomenological model, $t_i^\mathrm{soen}$ is the time of the given simulation at time step $i$, $s_\mathrm{ode}^\mathrm{interp}$ is the state quantity obtained with the ODE circuit model solved with an adaptive time mesh and interpolated to the coarser phenomenological model time mesh, and $\Delta t_i^\mathrm{soen} = t_{i+1}^\mathrm{soen} - t_i^\mathrm{soen}$. In the denominator, $n_t^\mathrm{ode}$ is the number of time steps in the ODE solution with the adaptive mesh, $s_\mathrm{ode}(t_i^\mathrm{ode})$ is the ODE solution on the non-uniform time grid without interpolation, and $\Delta t_i^\mathrm{ode} = t_{i+1}^\mathrm{ode} - t_i^\mathrm{ode}$. In all cases shown in Fig.\,\ref{fig:dend__ode_comparison__lin_ramp__100ps}, the value of $\chi^2$ is less than $7\times 10^{-5}$, and the time required to run the ODE model exceeded that to run the SOEN model by a factor of more than one thousand. The accuracy is worst for the smallest storage loop ($\beta/2\pi = 10^2$) and the shortest time constant ($\tau_\mathrm{di} = 10$\,ns). All values of $\chi^2$ and ratios of simulation times are given in Table \ref{tab:comparison__lin_ramp__100ps} in Appendix \ref{apx:soen_ode_comparison_additional_data}.

As another test case, a series of square flux pulses was input to a dendrite for the same values of $\beta$ and $\tau_\mathrm{di}$, as shown in Fig.\,\ref{fig:dend__ode_comparison__sqr_pulse__100ps}. The heights of these pulses were drawn randomly from the interval between the flux threshold [Fig.\,\ref{fig:dend__ri__thresholds__saturations}(a)] and the maximum value of 1/2. The pulse durations were drawn randomly from the interval between 5\,ns and 40\,ns, while the pauses between pulses were drawn randomly from the interval between 10\,ns and 100\,ns. In Fig.\,\ref{fig:dend__ode_comparison__sqr_pulse__100ps}, 10 such pulses were applied. The largest value of $\chi^2$ was $3.32\times 10^{-4}$, again in the case of smallest $\beta$ and $\tau_\mathrm{di}$, while the time to complete the ODE simulations was longer than that for the SOEN simulations by at least $10^3$ in all cases. Complete data for the square pulses is given in Table \ref{tab:comparison__sqr_pulse__100ps} in Appendix \ref{apx:soen_ode_comparison_additional_data}.

\begin{figure}[h!]
\includegraphics[width=8.6cm]{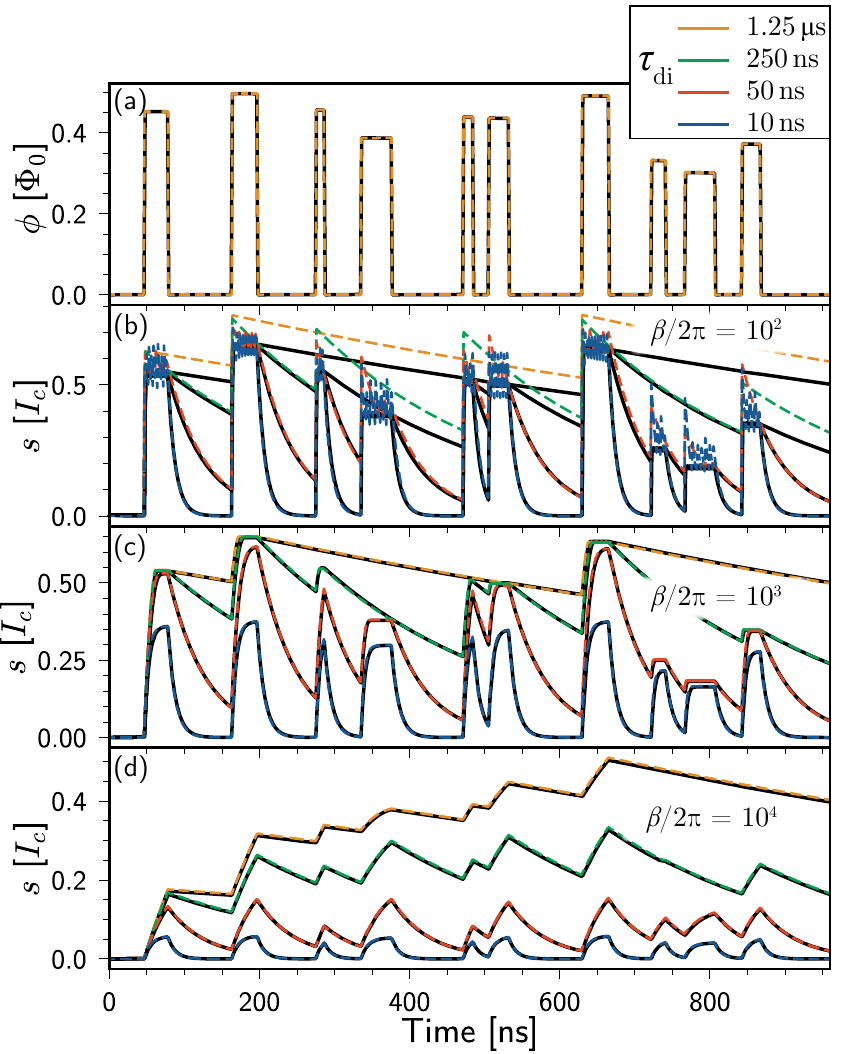}
\caption{\label{fig:dend__ode_comparison__sqr_pulse__1ns}Similar comparison as Fig.\,\ref{fig:dend__ode_comparison__sqr_pulse__100ps} except the time step was 1\,ns. The error in reproducing the drive signal was $\chi^2_\mathrm{drive} = 7.35\times 10^{-3}$.}
\end{figure}
\begin{figure}[h!]
\includegraphics[width=8.6cm]{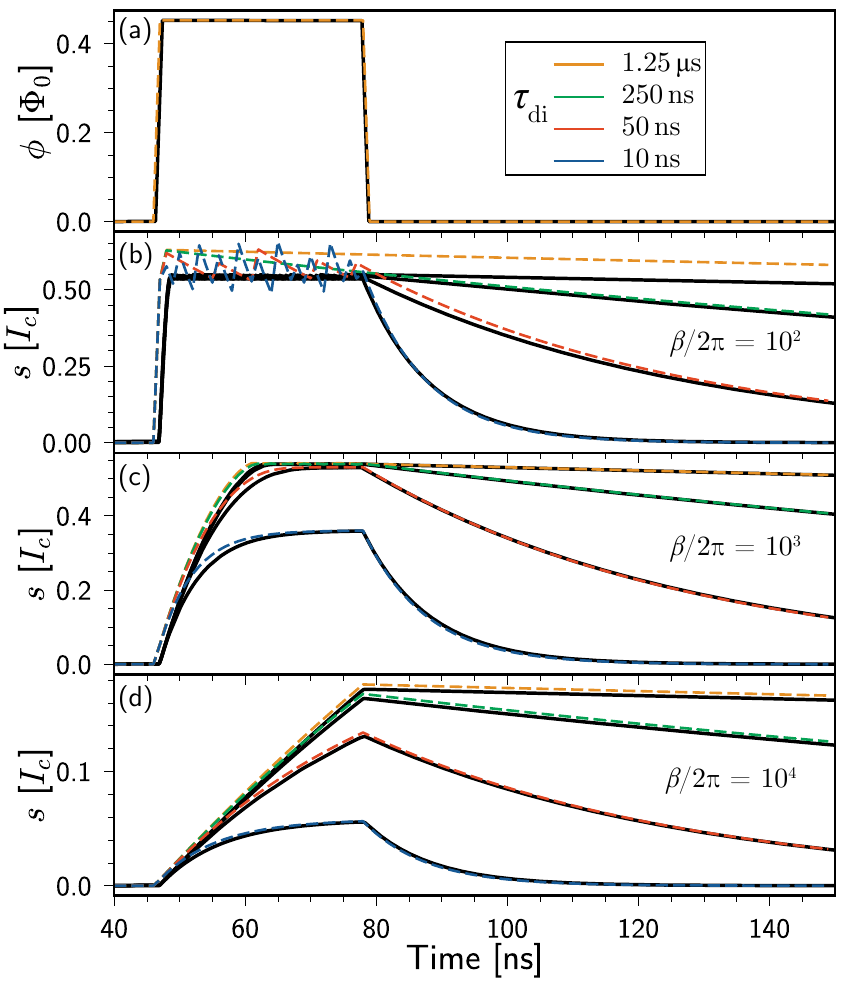}
\caption{\label{fig:dend__ode_comparison__sqr_pulse__1ns__zoom}The same simulations as Fig,\,\ref{fig:dend__ode_comparison__sqr_pulse__1ns} except zoomed on the time window around the first input pulse.}
\end{figure}
To push the limits of speed, the same linear ramp and square pulse cases were carried out using a time step of 1\,ns in the phenomenological model. The results for the square-pulse drive are shown in Fig.\,\ref{fig:dend__ode_comparison__sqr_pulse__1ns} for the same values of $\beta$ and $\tau_\mathrm{di}$. In all cases, the agreement between the two models is good, except for the smallest inductance of $\beta/2\pi = 10^2$. Poor performance in these cases results from the fact that the term $\Delta\tau\frac{\alpha}{\beta}$ in Eq.\,\ref{eq:main_equation__ode__discrete} becomes large, leading to numerical instability. More detail is shown in Fig.\,\ref{fig:dend__ode_comparison__sqr_pulse__1ns__zoom} where a temporal zoom on just the first input square pulse is presented. The numerical performance is seen to be unacceptable for the smallest value of $\beta$ and only marginally acceptable for values above that.

\begin{figure}[t!]
\includegraphics[width=8.6cm]{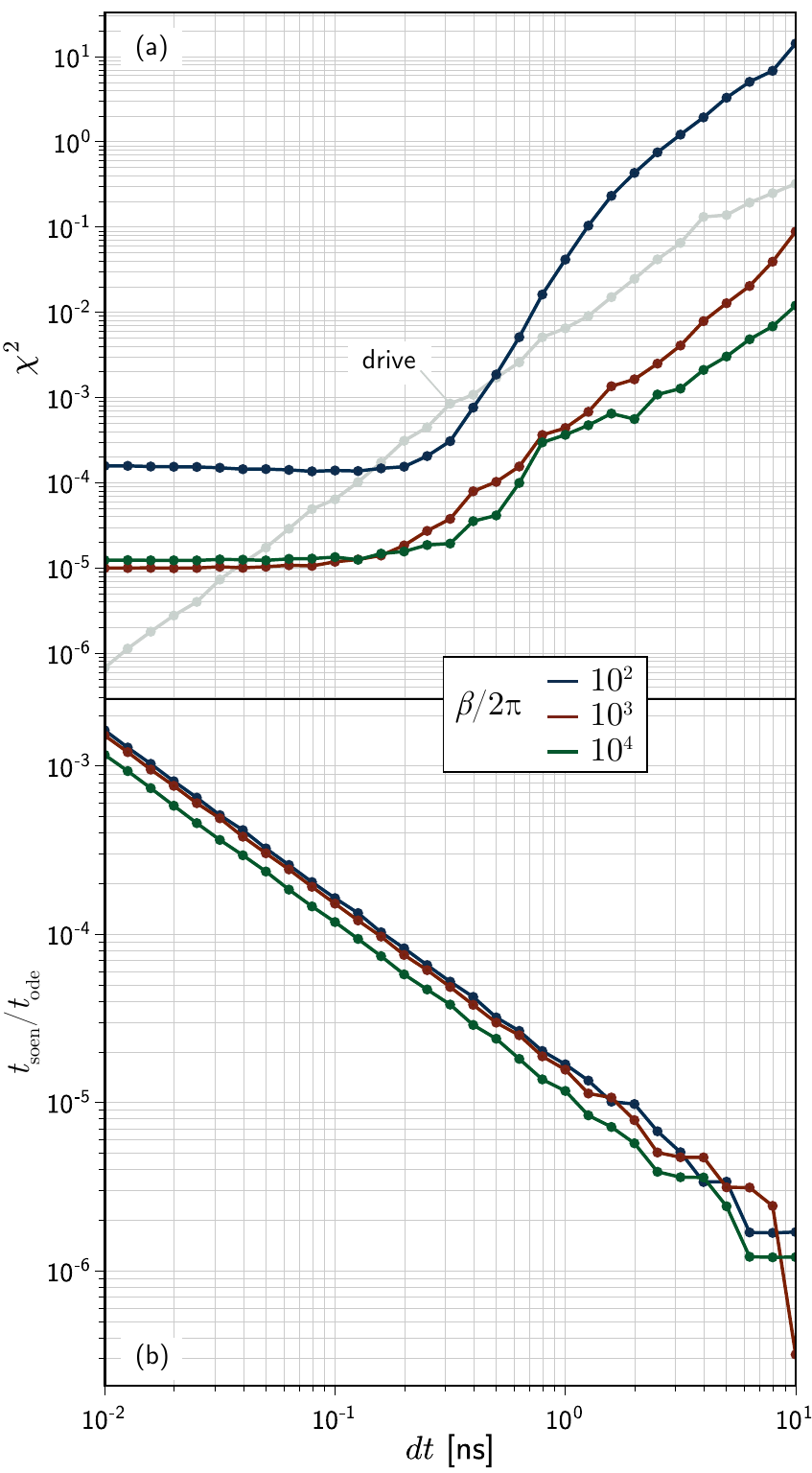}
\caption{\label{fig:dend__error_vs_dt}Quantification of the phenomenological model versus the circuit model. For the dendrite considered here, $\tau_\mathrm{di} = 250$\,ns and $i_\mathrm{b} = 1.702$. These simulations included 160 random square pulses, and the circuit model was set to converge to a relative and absolute accuracy of $10^{-5}$ in the \code{solve\_ivp} function. (a) The values of $\chi^2$ as a function of time step $dt$ for three values of $\beta$. (b) The ratio of the time required to run the simulations for the two numerical approaches as a function of time step $dt$.}
\end{figure}
To determine the largest acceptable time step, square-pulse-input simulations were conducted for numbers of input pulses ranging from 10 to 160 with time steps ranging from 10\,ps to 10\,ns for the same values of $\beta$ as above. To keep the computation time manageable, only $\tau_\mathrm{di} = 250$\,ns was considered, but individual test cases were conducted with other values to ensure the conclusions were insensitive to this number. The results of the $\chi^2$ values for each simulation as well as the ratio of computational run times were stored, and the results for the case of 160 random square pulse inputs is shown in Fig.\,\ref{fig:dend__error_vs_dt}. Further data on the other cases is given in Fig.\,\ref{fig:apx__error_vs_dt__summary} in Appendix \ref{apx:soen_ode_comparison_additional_data}. The $\chi^2$ values again show that the lowest value of $\beta$ is the most difficult to match, and the $\chi^2$ converges close to $10^{-4}$ with a time step of 200\,ps. All other $\beta$ values approach $\chi^2 = 10^{-5}$ with this same time step. The simulated time interval in this case was 12\,\textmu s, and with a time step of 200\,ps it took over ten thousand times longer to solve the system of circuit ODEs as to step through the phenomenological model. Based on this analysis, a time step of 200\,ps appears an optimal compromise between speed and accuracy. This time step is likely the convergence point in these simulations as 200\,ps is the rise and fall time of the square pulses. To simulate 12\,\textmu s of activity of a single dendrite with the phenomenological model required 1.02\,s, while simulating the same time interval with the ODE model required 2.78\, hours. Attempting to simulate longer time intervals with the circuit model becomes impracticable. 

\begin{figure}[tbh]
\includegraphics[width=8.6cm]{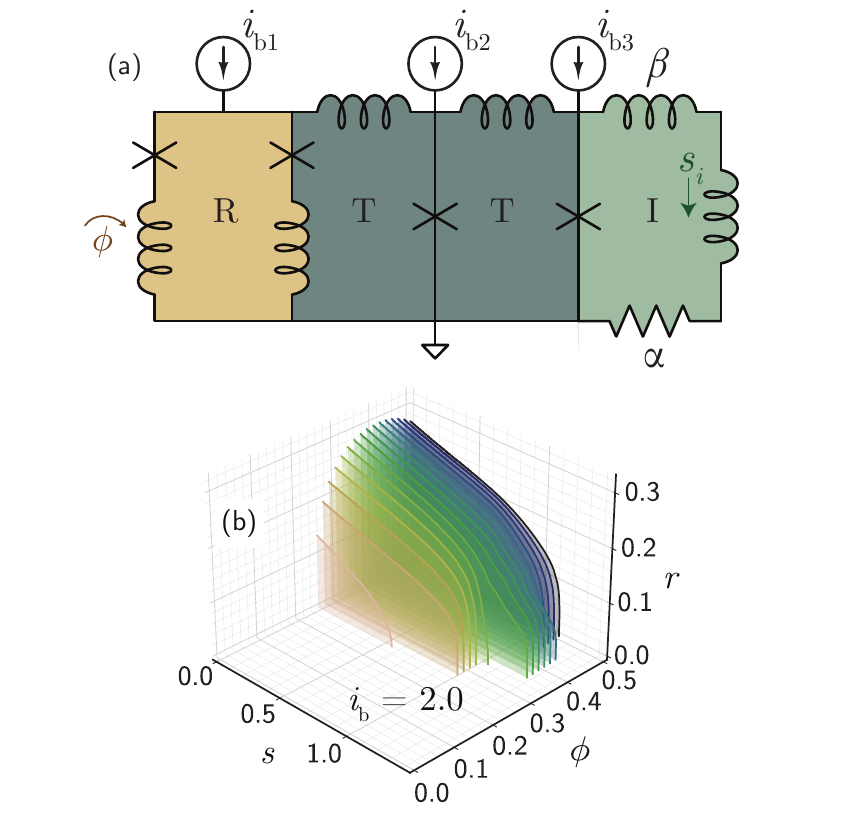}
\caption{\label{fig:dend__rtti__circuit__rate_arrays}The RTTI dendrite. (a) Circuit diagram showing the addition of two transmission loops. (b) An example $r$-shell for the RTTI dendrite.}
\end{figure}
Before proceeding to treat synaptic inputs to dendrites, we note an extension of the model to dendrites with additional circuit complexity. The response functions of the two-JJ dendrite is useful for many operations, but in certain cases it is desirable for the value of the saturation current to be less sensitive to the applied flux. This can be accomplished by adding a Josephson transmission line between the receiving and integrating loops of the dendrite, as shown in Fig.\,\ref{fig:dend__rtti__circuit__rate_arrays}(a). Due to the presence of two transfer loops between the receiving and integrating loops, we refer to this as the RTTI dendrite (receive-transfer-transfer-integrate). An example rate array for the RTTI dendrite is shown in Fig.\,\ref{fig:dend__rtti__circuit__rate_arrays}(b). While it is similar to the RI dendrite in that it has a threshold of applied flux and decreases with integrated $s$, the overall shape is more complicated, with appreciable structure close to saturation. The details of this structure can be used for various computations. In particular, the fact that the response is relatively flat for certain ranges of $\phi_\mathrm{r}$ can be used to achieve a digital response in which the output is insensitive to the exact input value of $\phi_\mathrm{r}$. This attribute will be exploited below to achieve the OR gate.

\begin{figure}[tbh]
\includegraphics[width=8.6cm]{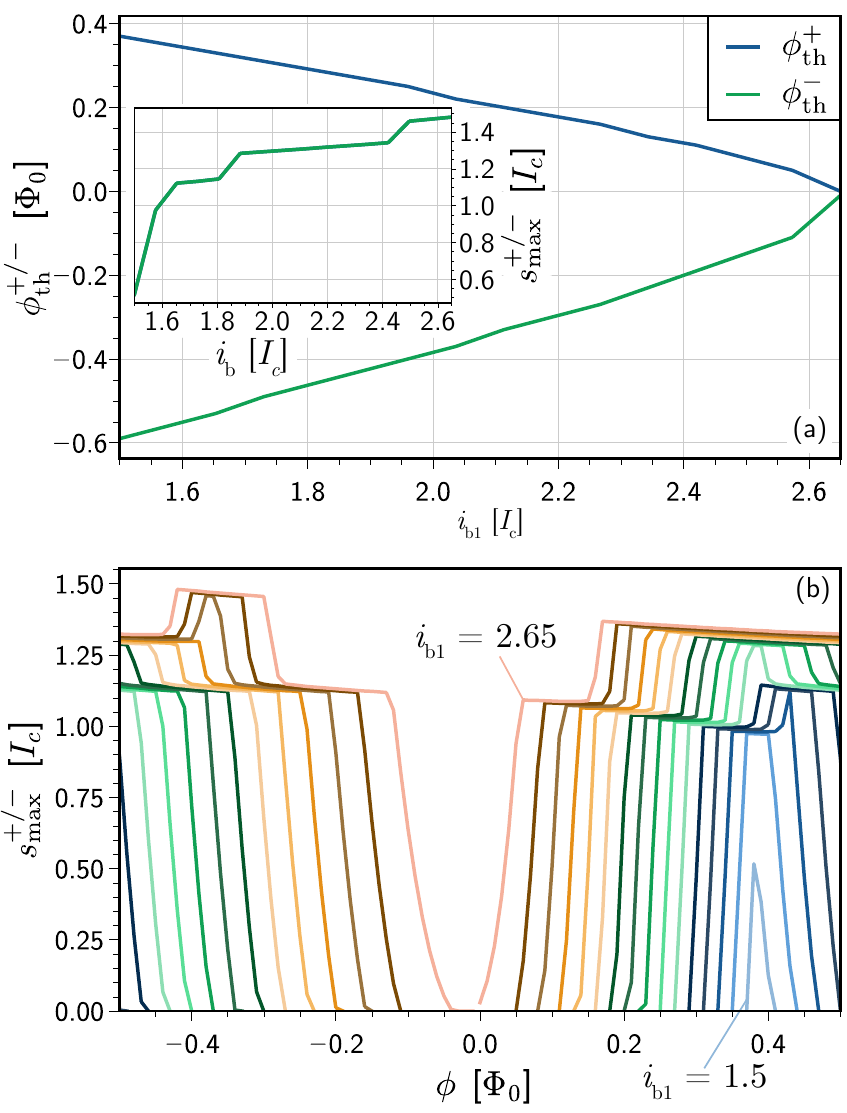}
\caption{\label{fig:dend__rtti__thresholds__saturations}The values of flux thresholds (a) and saturation levels (b) for the RTTI dendrite extracted from the rate arrays, analogous to these quantities shown in Fig.\,\ref{fig:dend__ri__thresholds__saturations} for the RI dendrite.}
\end{figure}
As for the case of the RI dendrite, we extract the thresholds and saturation values from the rate arrays. These functions are shown in Fig.\,\ref{fig:dend__rtti__thresholds__saturations}. In Fig.\,\ref{fig:dend__rtti__thresholds__saturations}(a) it can be observed that the thresholds for positive and negative applied flux are now slightly asymmetric due to the breaking of the symmetry of the circuit with the low-inductance transfer loop to the right of the receiving loop. More importantly, the values of $s_\mathrm{max}^{+/-}$ shown in Fig.\,\ref{fig:dend__rtti__thresholds__saturations}(b) are seen to have several regions of nearly flat response where the saturation level of the dendritic integration loop does not depend on applied flux. Such a response is useful for obtaining digital behaviors, as we show below when discussing logic gates in Sec.\,\ref{sec:synapses}. The peak value of $s_\mathrm{max}^{+\-}$ as a function of $i_\mathrm{b}$ is shown in the inset to Fig.\,\ref{fig:dend__rtti__thresholds__saturations}(a).

To simulate the RTTI dendrite with a conventional model, nine coupled ODEs are used, making the simulations even slower than the RI dendrite with its five coupled equations. With the phenomenological model, the RTTI dendrite reduces to a single ODE, just as the RI dendrite. The only difference is the form of the $r$-shells that provide the driving term. 

\section{\label{sec:synapses}Synapses}
As described in Secs.\,\ref{sec:overview_of_loop_neurons} and \ref{sec:dendrites}, the dendritic receiving loop is a SQUID. In many applications in science and technology, a SQUID is used as a measurement device with unmatched sensitivity for detecting magnetic flux. Based on the response curves of Fig.\,\ref{fig:squid}(b), it can be seen that a SQUID can be used as a flux-to-voltage transducer. Such a device can also be used to measure very low current levels when the current is coupled into the SQUID as flux using a transformer. It is in this mode of operation that we convert a dendrite into a synapse.  

When an SPD detects a photon, a current pulse is diverted out of the circuit, and that current returns with an $L/R$ time of around 35\,ns, which sets the recovery time and maximum rate of synapse events. To couple an SPD into a dendrite to form a synapse, this current output can be coupled into the SQUID that forms the receiving loop through a transformer, as shown in Fig.\,\ref{fig:circuits}(c). The equations of motion for the SPD can be solved exactly under a simple model \cite{yang2007modeling} to obtain the currents as a function of time. In this model, the resistance of the SPD is zero until a photon is detected, at which point it switches to a finite resistance for a finite duration. We are interested in the flux applied to a dendrite by the SPD receiver circuit; this flux is the product of the current $I_2$ with the mutual inductance of the transformer, which is a quantity we can select in design. Therefore, obtaining the currents is sufficient to obtain the flux, which we write as
\begin{equation}
\label{eq:spd_response}
\begin{split}
\phi_\mathrm{r} &= \phi_\mathrm{peak} \, \left( 1 - \frac{\tau_\mathrm{rise}}{\tau_\mathrm{fall}} \right) \\
&\times 
\begin{cases} \left[ 1 - e^{-t/\tau_\mathrm{rise}} \right],\text{ for}\,0\le t\le t_0 \\
 \left[ 1 - e^{-t_0/\tau_\mathrm{rise}} \right]\,e^{-(t-t_0)/\tau_\mathrm{fall}},\text{ for}\,t > t_0.
\end{cases}
\end{split}
\end{equation}
In Eq.\,\ref{eq:spd_response} it is assumed a synapse event occurred at time $t = 0$, which is the time at which the SPD entered the resistive state, and $t_0$ is the duration for which the SPD stays in the resistive state following the absorption of a photon, which we take to be 200\,ps, following the model of Ref.\,\onlinecite{yang2007modeling}. This duration is commensurate with the 200\,ps time step that we found to be optimal in Sec.\,\ref{sec:dendrites}. The resistance $r_1$ [Fig.\,\ref{fig:circuits}(c)] is the resistance of the SPD in the time window $t_0$ after absorbing a photon, while $r_2$ is a fixed resistance chosen to obtain a sufficiently long $L/R$ recovery time to prevent the SPD from latching \cite{natarajan2012superconducting}. The total inductance of the circuit is $L_\mathrm{tot} = L_1 + L_2 + L_3$, which includes the kinetic inductance of the SPD itself ($L_1$), the inductance of the transformer input coil ($L_3$), and any additional parasitic inductance in the circuit $L_2$. In terms of these parameters, $\tau_\mathrm{rise} = L_\mathrm{tot}/(r_1+r_2)$ and $\tau_\mathrm{fall} = L_\mathrm{tot}/r_2$. Typically $\tau_\mathrm{rise}$ is a few tens of picoseconds and $\tau_\mathrm{fall}$ is 30\,ns - 50\,ns \cite{khan2022superconducting}. The quantity $\phi_\mathrm{peak}$ can be chosen in design and set by the transformer between the SPD and the dendrite. Throughout this work we take $\phi_\mathrm{peak} = 0.5$ in units of $\Phi_0$.

\begin{figure}[tbh]
\includegraphics[width=8.6cm]{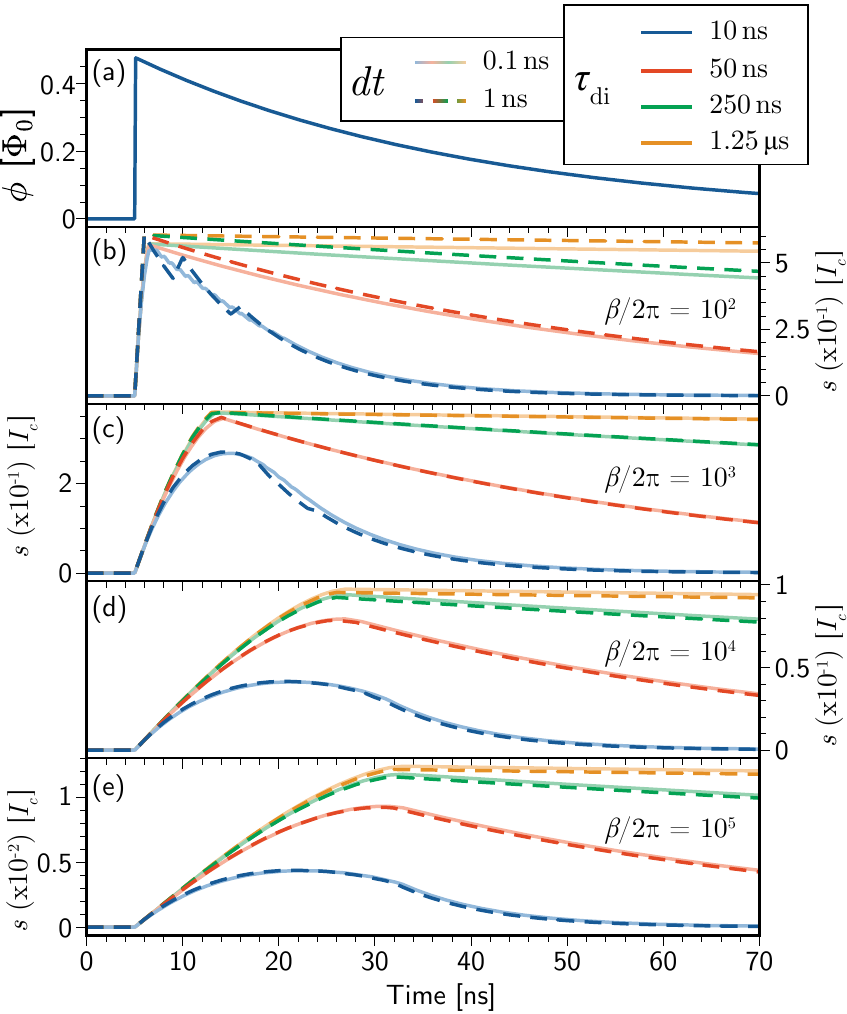}
\caption{\label{fig:synapses__one_pulse}Output of the phenomenological model for the case of a synapse. (a) Flux output from the SPD. (b) Signal $s$ in the dendritic integration loop for the case of $\beta/2\pi = 10^2$. The solid lines result from a time step of 100\,ps, while the dashed lines used 1\,ns. (c)-(e) $\beta/2\pi = 10^3$, $10^4$, and $10^5$, respectively.}
\end{figure}
We now consider several examples of the phenomenological model with the output of an SPD coupled as input to a dendrite. In Fig.\,\ref{fig:synapses__one_pulse} we show the response of the dendrite to a single synaptic pulse. The time course of the total applied flux to the receiving loop, $\phi$, is shown in Fig.\,\ref{fig:synapses__one_pulse}(a), and four values of $\beta$ are shown in Fig.\,\ref{fig:synapses__one_pulse}(b)-(e), each for four values of $\tau_\mathrm{di}$. In each case, the dashed line corresponds to using a time step of 1\,ns in the phenomenological model, while the solid line corresponds to 100\,ps. Again we see that the worst performance occurs with small $\beta$ and small $\tau_\mathrm{di}$.

\begin{figure}[tbh]
\includegraphics[width=8.6cm]{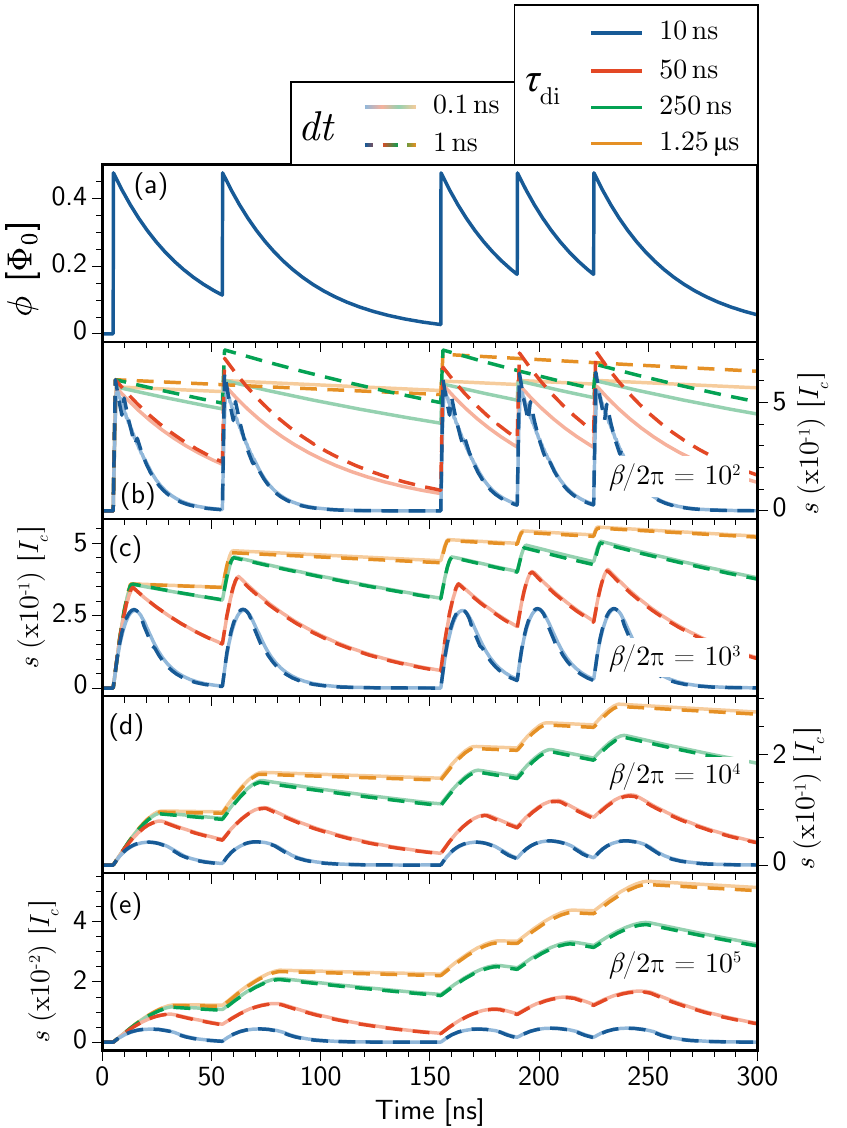}
\caption{\label{fig:synapses__pulse_sequence}Similar data to Fig.\,\ref{fig:synapses__one_pulse} except signals are in response to a train of five input synapse events.}
\end{figure}
Similar data is shown in Fig.\,\ref{fig:synapses__pulse_sequence} for an arbitrary sequence of five synapse events. Dendritic integration is evident, and the effects of $\beta$ and $\tau_\mathrm{di}$ can be seen: for smaller $\beta$ and larger $\tau_\mathrm{di}$, the integration loop reaches saturation after a small number of input pulses, while for larger $\beta$ and/or smaller $\tau_\mathrm{di}$, the integrated signal does not reach saturation even by the end of the train of five pulses. 

Using synapses and dendrites to perform various temporal extensions of basic logic operations is a primary function of biological neural computation. Here we consider four basic logic operations: AND, OR, AND-NOT, and XOR. Each of these operations is accomplished with a single dendrite receiving input from two synapses.

\begin{figure}[t!]
\includegraphics[width=8.6cm]{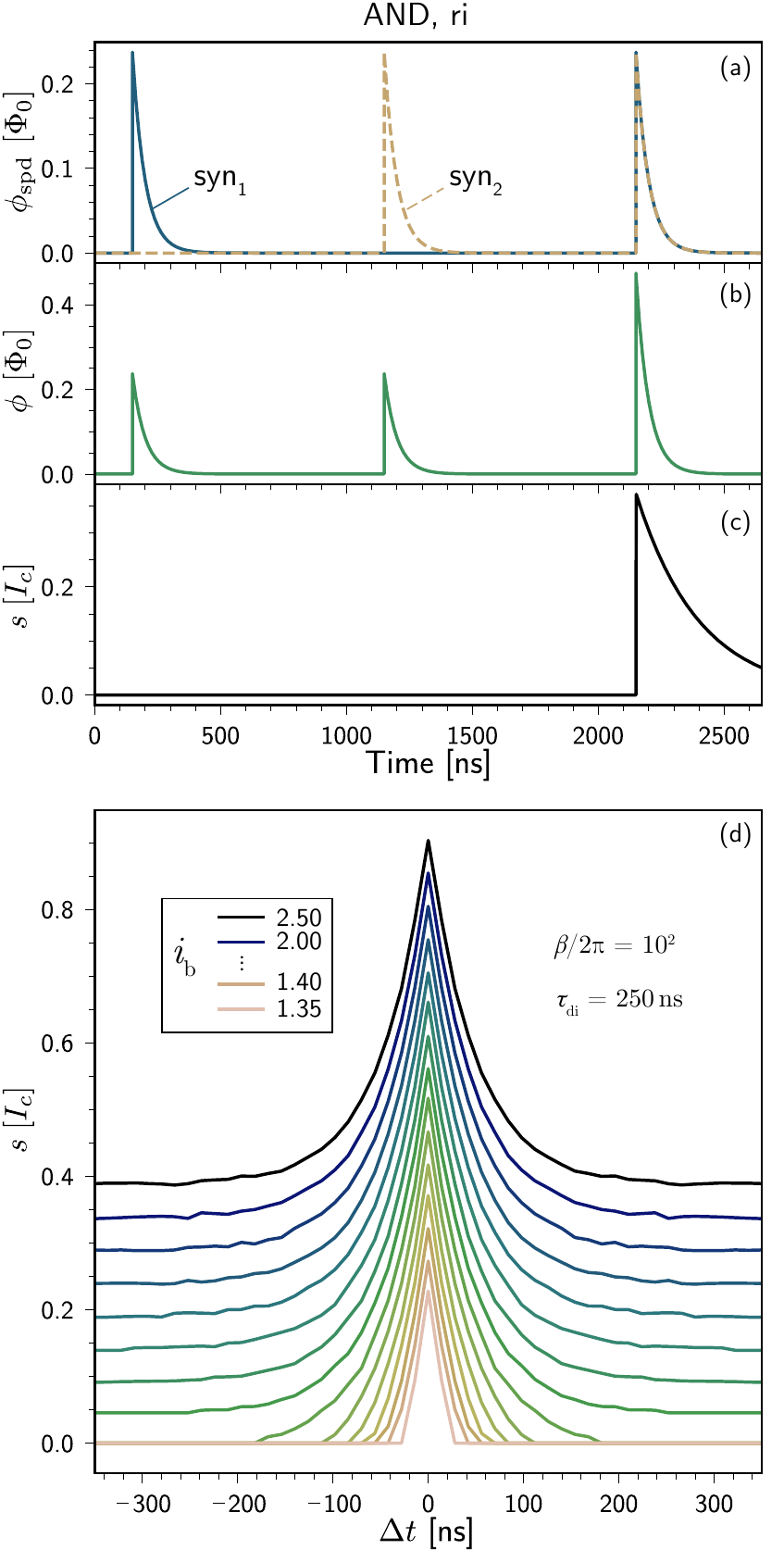}
\caption{\label{fig:dend__logic__AND}Operation of the AND gate. (a) Input flux from the two synapses separately. (b) Combined input flux. (c) Dendrite output signal. (d) Analysis of the gate response as a function of time delay between the two synapse events for several values of $i_\mathrm{b}$.}
\end{figure}
\begin{figure}[t!]
\includegraphics[width=8.6cm]{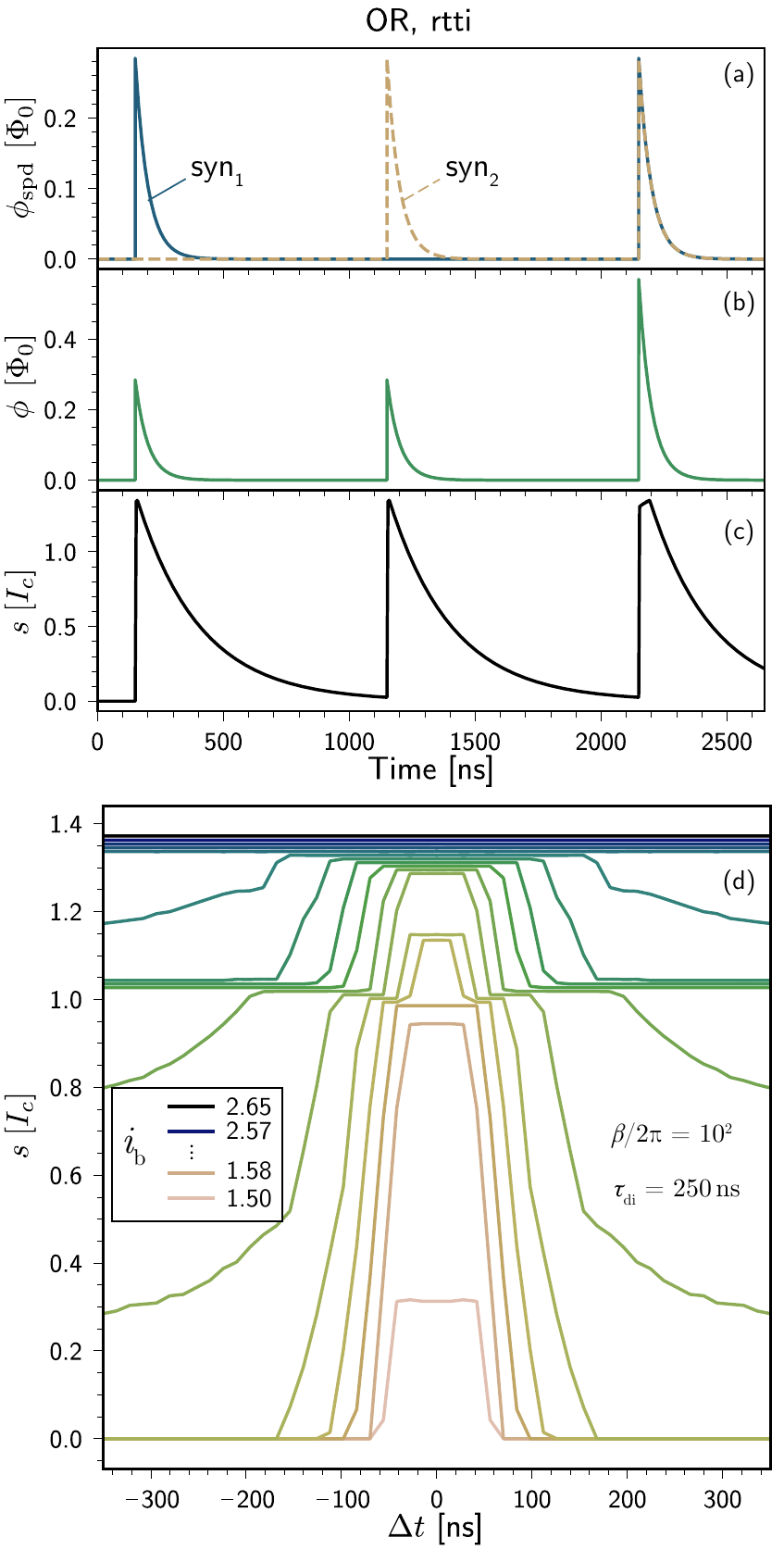}
\caption{\label{fig:dend__logic__OR}Operation of the OR gate. (a) Input flux from the two synapses separately. (b) Combined input flux. (c) Dendrite output signal. (d) Analysis of the gate response as a function of time delay between the two synapse events for several values of $i_\mathrm{b}$.}
\end{figure}
\begin{figure}[t!]
\includegraphics[width=8.6cm]{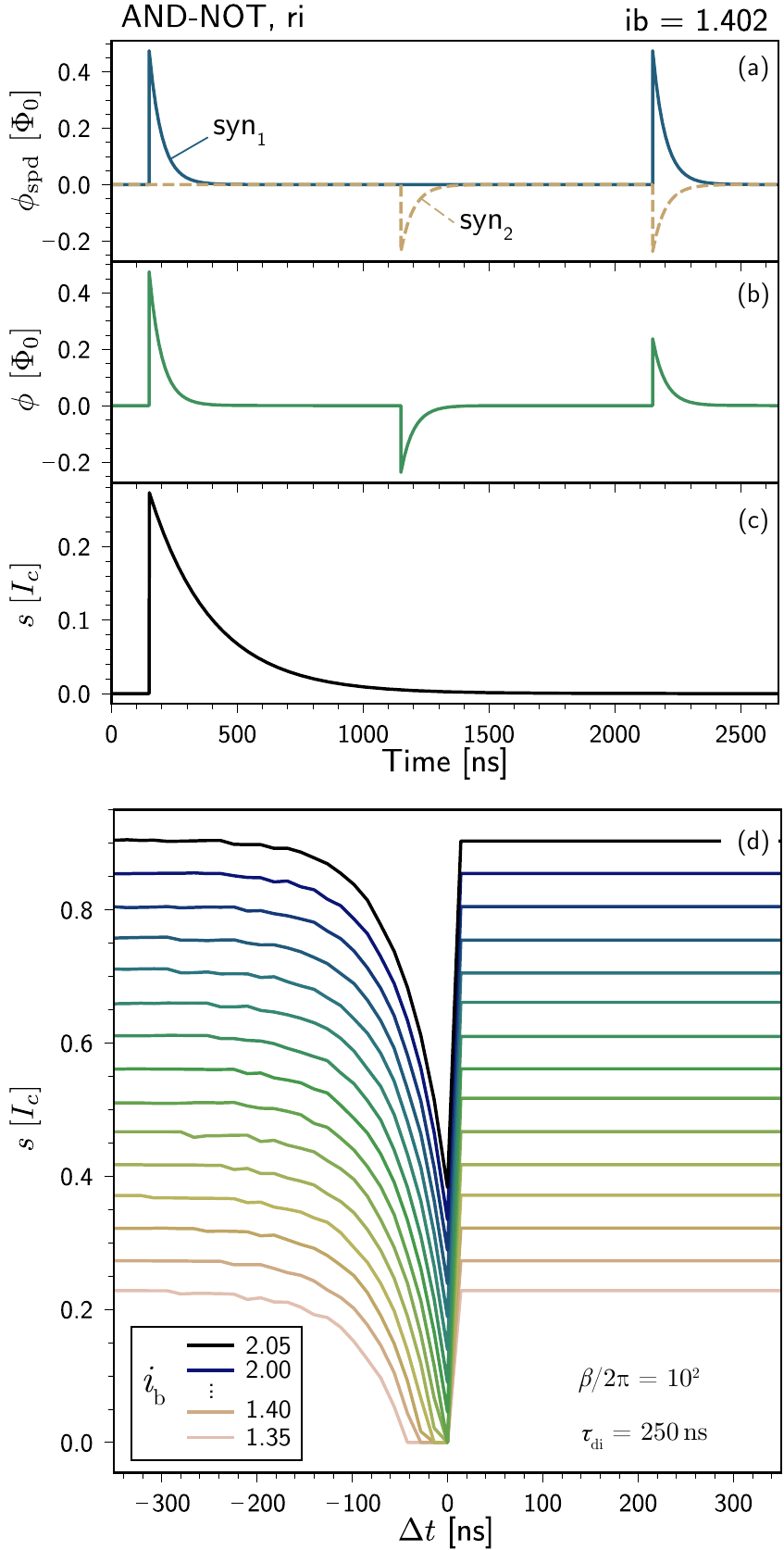}
\caption{\label{fig:dend__logic__ANDNOT}Operation of the AND-NOT gate. (a) Input flux from the two synapses separately. (b) Combined input flux. (c) Dendrite output signal. (d) Analysis of the gate response as a function of time delay between the two synapse events for several values of $i_\mathrm{b}$.}
\end{figure}
\begin{figure}[t!]
\includegraphics[width=8.6cm]{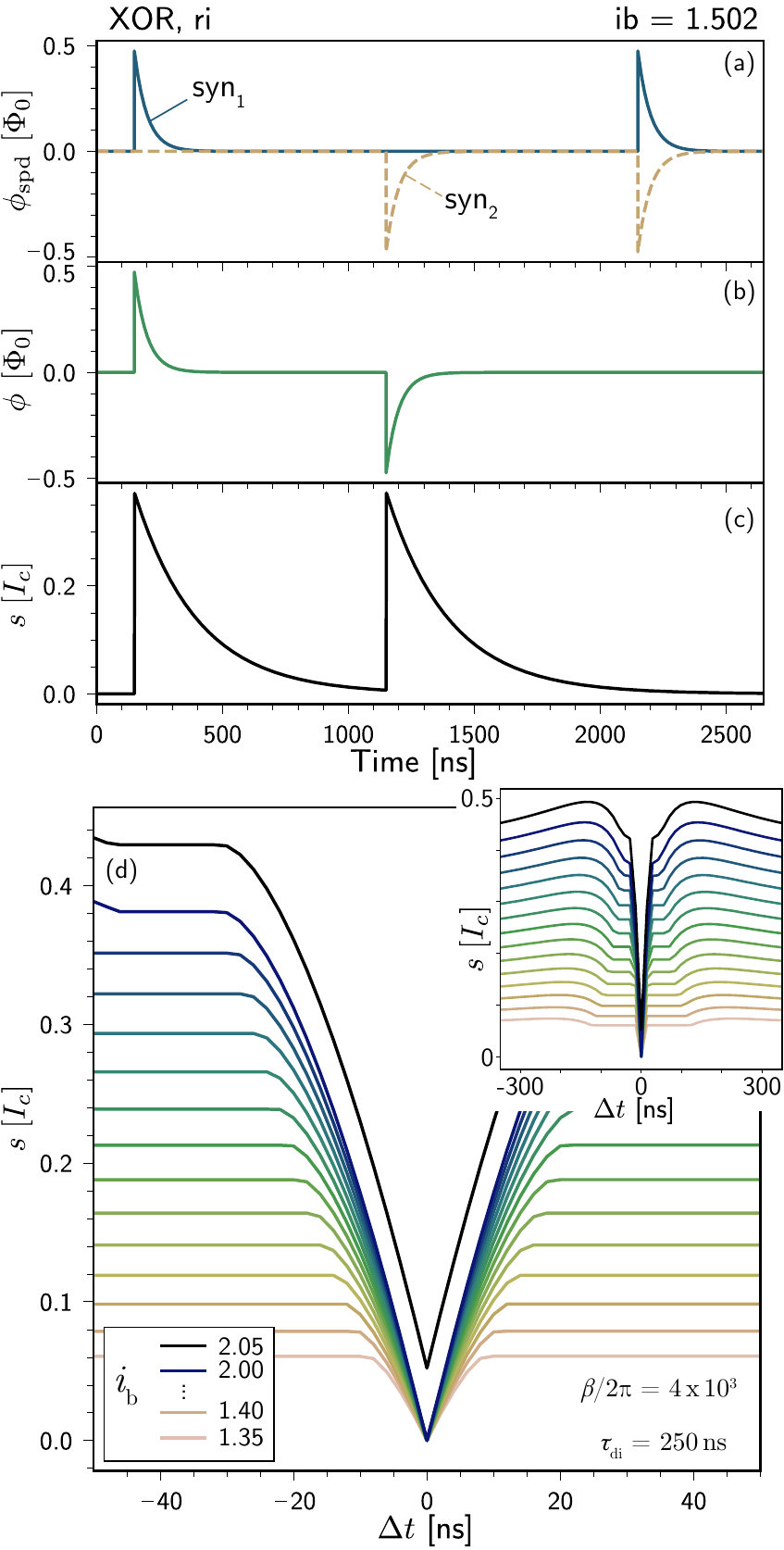}
\caption{\label{fig:dend__logic__XOR}Operation of the XOR gate. (a) Input flux from the two synapses separately. (b) Combined input flux. (c) Dendrite output signal. (d) Analysis of the gate response as a function of time delay between the two synapse events for several values of $i_\mathrm{b}$. Here the time axis is narrower than Figs.\,\ref{fig:dend__logic__AND}, \ref{fig:dend__logic__OR}, and \ref{fig:dend__logic__ANDNOT}. The inset shows the full time axis of the other logic gate figures.}
\end{figure}

The AND operation is shown in Fig.\,\ref{fig:dend__logic__AND}. The flux from the two synapses independently is shown in Fig.\,\ref{fig:dend__logic__AND}(a), their sum is shown in Fig.\,\ref{fig:dend__logic__AND}(b), and the signal in the dendritic integration loop is shown in Fig.\,\ref{fig:dend__logic__AND}(c). In Figs.\,\ref{fig:dend__logic__AND}-\ref{fig:dend__logic__XOR} the flux from a single SPD to the dendrite is referred to as $\phi_\mathrm{spd}$, while the combined flux from both SPDs to the receiving loop of the dendrite is labeled $\phi_\mathrm{r}$. To achieve the AND response, the dendrite is configured such that the flux from a single synapse is insufficient to drive the dendrite above threshold, but the flux from two synapses together is adequate, and signal is added to the integration loop only when the pulses are coincident. 

Because the synaptic flux and the dendritic signal are both functions of time, it is important to know how the accumulated signal will vary with the delay between the arrival of the two synapse events. This data is shown in Fig.\,\ref{fig:dend__logic__AND}(d) for multiple values of the bias to the dendrite, $i_\mathrm{b}$. Such a circuit can be used as a temporal coincidence detector, as described in Ref.\,\onlinecite{shainline2019fluxonic}. 

The OR operation is shown in Fig.\,\ref{fig:dend__logic__OR}. The challenge with OR is that we would like the amplitude of the signal in the dendritic integration loop to be identical whether one or both synapses are active. Because both the rate of fluxon production and the value of saturation depend on the applied flux for the RI dendrite, this is difficult to accomplish. However, the flat response of the RTTI dendrite can generate this input-output relationship, and that type of dendrite has been used to generate Fig.\,\ref{fig:dend__logic__OR}. In Fig,\,\ref{fig:dend__logic__OR}(b) we can see that when both synapses are active the applied flux to the receiving loop has twice the amplitude as when only a single synapse is active, yet the integrated signal in Fig.\,\ref{fig:dend__logic__OR}(c) is nearly identical whether one or both synapses are active. The response is possible due to the flat-top behavior of the RTTI dendrite. We have biased the dendrite at a point such that when the flux from one synapse is present the dendrite just enters the plateau of the response, and when the flux from both is present it remains on the plateau.

The response as a function of the delay between the two synapse events is shown in Fig.\,\ref{fig:dend__logic__OR}(d). For lower bias points the flat-top response only occurs for a relatively short coincidence window, providing a response more like the AND gate. Yet for higher bias points the response is very flat, independent of temporal delay between the synapse events, just as we desire for the OR gate. We will see in Sec.\,\ref{sec:neurons} that this response is useful in achieving the desired classification task. 

The AND-NOT operation is shown in Fig.\,\ref{fig:dend__logic__ANDNOT}. This logic gate produces a logical one at the output if and only if the first input is a logical one and the second input is a logical zero. This is distinct from the NAND gate. For AND-NOT we return to the RI dendrite. In this case, synapse one is excitatory, while synapse two is inhibitory with the opposite sign of flux applied as well as a reduced amplitude. Fig.\,\ref{fig:dend__logic__ANDNOT}(c) shows that the dendrite only becomes active when synapse one receives a pulse in the absence of synapse two receiving a pulse. Extension to various temporal delays in Fig.\,\ref{fig:dend__logic__ANDNOT}(d) shows the effect of the asymmetric coupling of the two synapses.

The XOR operation is shown in Fig.\,\ref{fig:dend__logic__XOR}. XOR is accomplished in much the same manner as AND-NOT, except the coupling between the two synapses is equal. Thus, when either synapse is active in the absence of the other the dendrite becomes active. In this instance we are making use of the fact that the RI dendrite response is symmetric with respect to positive or negative flux. The response as a function of relative delay between the synapse events is shown in Fig.\,\ref{fig:dend__logic__XOR}(d), and it is evident that cancellation of the two pulses requires relatively high timing precision. The response over longer delays is shown as an inset. The requisite timing precision can be adjusted with the $L/R$ time of the SPD. This implementation of XOR in a single dendrite is analogous to that observed in human pyramidal neurons in cortical layers two and three \cite{gidon2020dendritic}.

\begin{figure}[tbh]
\includegraphics[width=8.6cm]{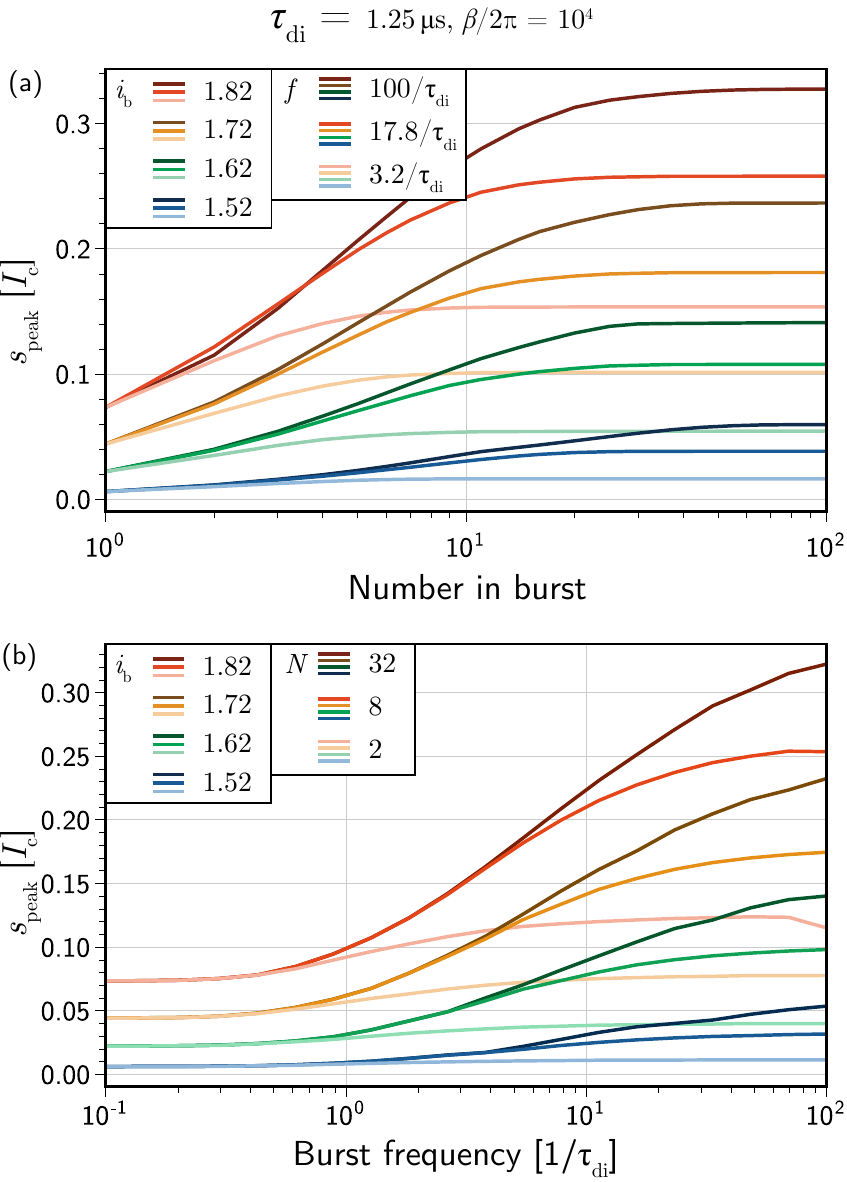}
\caption{\label{fig:synapses__transfer_functions}Transfer functions for the RI dendrite. (a) Peak of integrated signal as a function of the number of synapse events in a burst for several values of $i_\mathrm{b}$ and frequency of burst inputs. (b) Peak of integrated signal as a function of the frequency of input pulse train for several values of $i_\mathrm{b}$ and number of pulses in the train.}
\end{figure}
In addition to discrete responses to few synapse events, the phenomenological model can help us understand the transfer functions performed by synapses coupled to dendrites when many pulses are incident. In Fig.\,\ref{fig:synapses__transfer_functions} we analyze these responses for various bias conditions, revealing the role of $i_\mathrm{b}$ for sculpting dendritic behavior. Figure \ref{fig:synapses__transfer_functions}(a) shows the peak of the accumulated signal in the dendritic integration loop as a function of the number of synapse events input in a burst. Traces from the same color family correspond to the same value of $i_\mathrm{b}$. Lighter traces correspond to pulses input at a slower rate relative to the time constant of the dendrite, while darker traces show the response for higher-frequency inputs. The response is unsurprisingly larger for higher input frequencies and can be broadly adjusted with the bias current. The sigmoid-like response with threshold and saturation has been useful in various machine learning applications. Similar responses are seen in Fig.\,\ref{fig:synapses__transfer_functions}(b) where the frequency of the burst is now the quasi-continuous variable on the $x$-axis, and various curves have different numbers of input pulses. Similar thresholding and saturation behavior is evident, and the same qualitative behavior has been demonstrated in experiments of related circuits \cite{khan2022superconducting}.

If generated with conventional circuit models, the plots of Fig.\,\ref{fig:synapses__transfer_functions} would have required so much computational time as to be impracticable with the computer used for this study. With the phenomenological model they are produced in minutes so that many scenarios may be investigated and insight across parameter space can be quickly gained. 

\section{\label{sec:neurons}Neurons}
Whereas a synapse was shown to be a formed where a single-photon detector couples to a dendrite, a loop neuron can be as simple as a dendrite with a transmitter circuit providing output. Such a soma circuit is shown in Fig.\,\ref{fig:circuits}(d). While all dendrites have a bias-dependent threshold for activation in the receiving loop below which applied flux elicits no response, the soma is distinguished in that it also has a threshold in the integration loop at its output. Embedded in the integration loop is a superconducting thresholding element referred to as a tron that switches from a zero-resistance state below threshold to a high-resistance state when the current in the integration loop reaches a critical value. Such comparators have been demonstrated as an interface between superconductor and semiconductor electronics \cite{zhao2017nanocryotron,mccaughan2019superconducting}. In the present case, accumulation of sufficient signal in the soma's integration loop will drive the tron to the voltage state, at which point a semiconductor-based transmitter circuit will generate photons from a light-emitting diode (LED). This is a spike event or action potential of a superconducting optoelectronic loop neuron. When this threshold is reached and the tron switches, the current in the somatic integration loop is purged, and integration must begin again. In most dendrites in the system, any accumulated signal in the integration loop is immediately coupled as flux into other dendrites. However, the soma has no output other than the transmitter circuit, so its state is only communicated to other elements of the network when threshold is reached and light is produced.

Several approaches to transmitter circuits for loop neurons have been explored \cite{shainline2019superconducting}, and here we base the phenomenological model on a circuit concept that leverages MOSFETs in an essentially digital configuration. Upon reaching threshold, activity is transferred to these semiconductor circuits, and to understand their operation we must work backward from the light source itself. To maintain ``brevity'' in the main thread of this article, we relegate the details to appendices. Appendix \ref{apx:source} describes a rate-equation model to capture the primary behavior of the light emitters, while Appendix \ref{apx:transmitter} treats the transmitter circuit from the thresholding element through the LED. Because the subject of this work is to establish a phenomenological model, here we simply extract the key message from those appendices: when a soma reaches threshold, photons are produced with a few nanosecond delay followed by an exponential probability distribution. The distribution is obtained numerically through simulation of the transmitter circuit combined with the source rate equation model. In the reduced phenomenological model that is the subject of this work, each time a somatic integration current reaches threshold, a specified number of photon-production times are drawn randomly from this distribution of a delay followed by an exponential decay. These photon times are then assigned randomly to the neuron's downstream synapses and added to a list of times of input synapse events. With this approach, the number of ODEs which must be treated remains unchanged, and the added computational burden of treating the transmitter circuit and sources is reduced to dealing with random number generation only on the time steps at which the soma's threshold is reached. The indefatigable reader is directed to Appendices \ref{apx:source} and \ref{apx:transmitter} for more details.

\begin{figure}[h!]
\includegraphics[width=8.6cm]{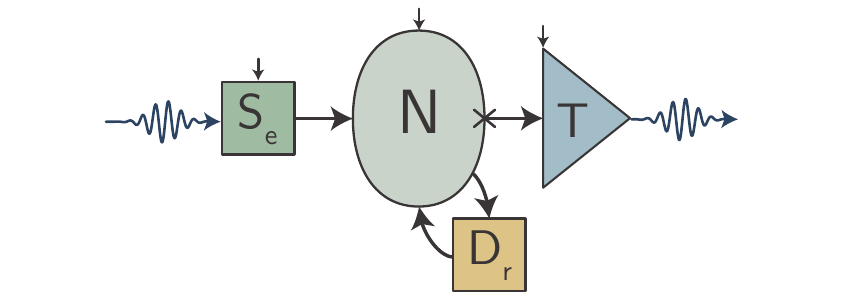}
\caption{\label{fig:neuron__one_synapse__schematic__circuit}Schematic of a monosynaptic point neuron with excitatory synapse ($\mathsf{S_e}$), neuron cell body ($\mathsf{N}$), transmitter ($\mathsf{T}$) and refractory dendrite ($\mathsf{D_r}$).}
\end{figure}
In addition to this output thresholding behavior, a loop neuron requires a means of establishing a refractory period so that spike events are indeed discrete and their rate does not exceed the roughly 20\,MHz at which the synaptic SPDs can respond. A refractory period is accomplished by adding a second dendrite to the soma. A schematic of this minimal point neuron is shown in Fig.\,\ref{fig:neuron__one_synapse__schematic__circuit}, where thie refractory dendrite is labeled $\mathsf{D_r}$. This refractory dendrite is driven by a flux pulse identical in form to the SPD response (Eq.\,\ref{eq:spd_response}) when the soma reaches threshold, and the signal accumulated in the refractory integration loop is coupled back to the soma's receiving loop as inhibition. The time constant of the refractory dendrite's integration loop therefore sets the refractory period and maximum neuronal firing rate. To achieve robust refraction, the coupling from the refractory dendrite back to the soma is designed so that the inhibitory flux is as strong as possible without driving activity on the negative-flux side of the $r$-shell. This can be accomplished with knowledge of the values of $s_\mathrm{max}$ for the refractory dendrite at its bias point as well as the values of threshold flux $\phi_\mathrm{th}^+$ and $\phi_\mathrm{th}^-$. The refractory dendrite is chosen to have a relatively small value of $\beta$ so that it saturates quickly at $s_\mathrm{max}$ each time the soma fires. The coupling from the refractory to the somatic dendrite is then given by $J = ( \phi_\mathrm{th}^+ \, - \, \phi_\mathrm{th}^-)/s_\mathrm{max}$.

The role of the refractory dendrite is to inhibit the neuron for a brief duration after it spikes. Yet similar inhibition in response to activity is often desirable on longer timescales. Homeostatic plasticity \cite{cooper2012bcm,pozo2010unraveling,turrigiano2012homeostatic} is one means by which the activity of neurons can be maintained in a useful dynamic range. This self-regulatory behavior can be accomplished with feedback analogous to refraction if further dendrites receive input when the soma fires and couple inhibitively to the soma's receiving loop. Such a homeostatic dendrite is likely to have a longer time constant and also a larger integration loop inductance than the refractory dendrite. In this way, each time the neuron fires, a small inhibitory signal is fed back to the soma, effectively increasing the threshold for the next firing. This signal will decay over a longer time period, on the order of many interspike intervals. Such a plasticity mechanism can be treated with the same phenomenological model presented here.

\begin{figure}[h!]
\includegraphics[width=8.6cm]{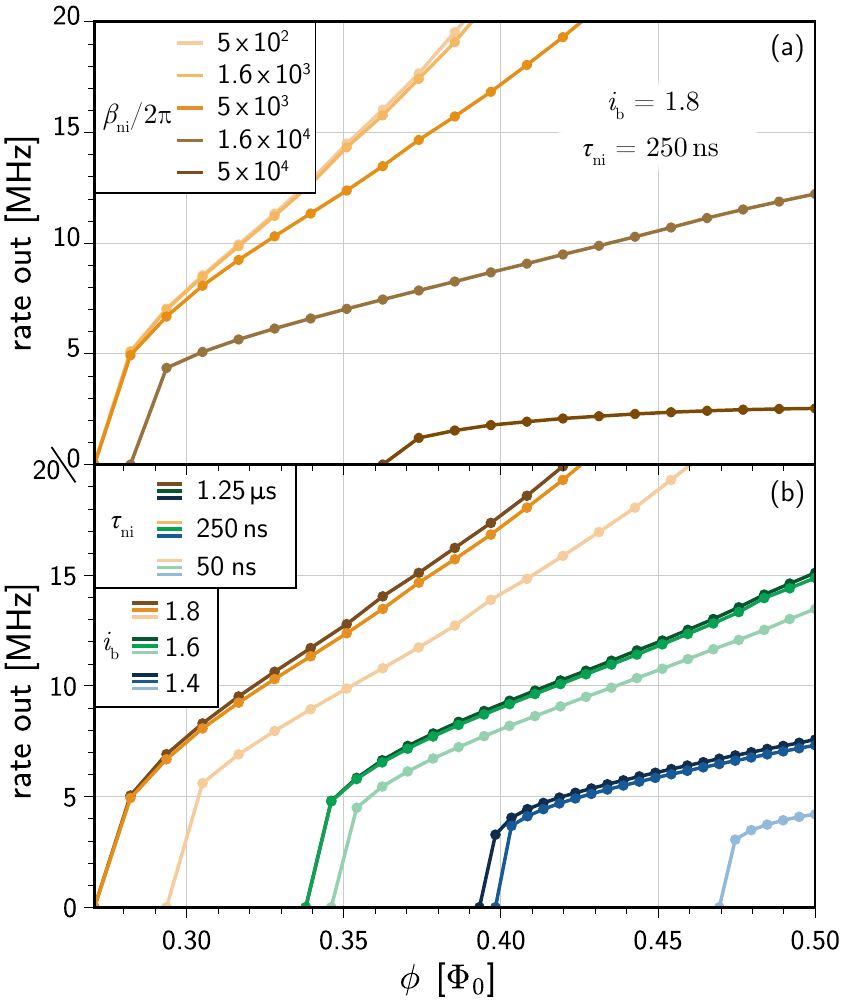}
\caption{\label{fig:neuron__rate_out_vs_applied_flux}Neuron firing rate versus applied flux. (a) Comparing several values of the neuronal integration loop inductance parameter, $\beta_\mathrm{ni}$. (b) Comparing several values of neuronal integration loop time constant and bias current. For $i_\mathrm{b} = 1.6$ and 1.8, $\beta_\mathrm{ni}/2\pi = 5\times 10^3$. For $i_\mathrm{b} = 1.4$, $\beta_\mathrm{ni}/2\pi = 1.6\times 10^4$. }
\end{figure}
To begin analysis of neurons with this framework, we consider a neuron with no dendritic tree in which all inputs are received directly at the soma. Such a neuron is referred to as a point neuron. As a first means of gaining intuition about the design of a point neuron, we consider the spiking behavior of a simple loop neuron with steady-state flux coupled directly into the soma and a single refractory dendrite to induce pulsatile behavior at the transmitter. In Fig.\,\ref{fig:neuron__rate_out_vs_applied_flux} we show the rate of neuron spiking as a function of the steady-state applied flux, $\phi$. Figure \ref{fig:neuron__rate_out_vs_applied_flux}(a) shows the response for five values of the neuronal integration loop inductance parameter, $\beta_\mathrm{ni}$. In all cases the time constant of the refractory integration loop was 50\,ns and the value of the somatic integration loop threshold was set to $0.3\,s_\mathrm{max}$. Smaller values of $\beta_\mathrm{ni}$ lead to rapidly refilling of the integrated signal after it is purged upon reaching threshold, while larger values require more time to accumulate signal back to threshold, leading to slower firing activity for the same flux drive. 

By contrast, in Fig.\,\ref{fig:neuron__rate_out_vs_applied_flux}(b) the somatic integration loop inductance was fixed at an intermediate value, while the operating bias point and somatic integration loop time constant were adjusted. Lower biasing requires higher input flux to achieve threshold and also leads to slower firing rates, while higher bias reduces threshold and increases the output rate. Such a control parameter is the primary means by which the soma can be reconfigured dynamically during network operation, either in an unsupervised manner by activity within the system or in a supervised manner by a user or control system. The time constant cannot be changed dynamically, as it is set by the resistance and inductance of the loop, which are fixed in fabrication. However, the time constant can have an appreciable effect on the transfer characteristics, as is most pronounced for the low-bias case where reducing the time constant to 50\,ns dramatically increases the flux threshold and reduces the output rate at a given value of applied flux. All of these parameters must be considered when designing neurons to play various roles in a network.

As in the transfer functions of Fig.\,\ref{fig:synapses__transfer_functions}, the responses in Fig.\,\ref{fig:neuron__rate_out_vs_applied_flux} would have been quite cumbersome to obtain with the conventional circuit equation approach. Even at this basic level, the phenomenological model provides utility in guiding the design principles of loop neurons across a broad parameter space.

\begin{figure}[h!]
\includegraphics[width=8.6cm]{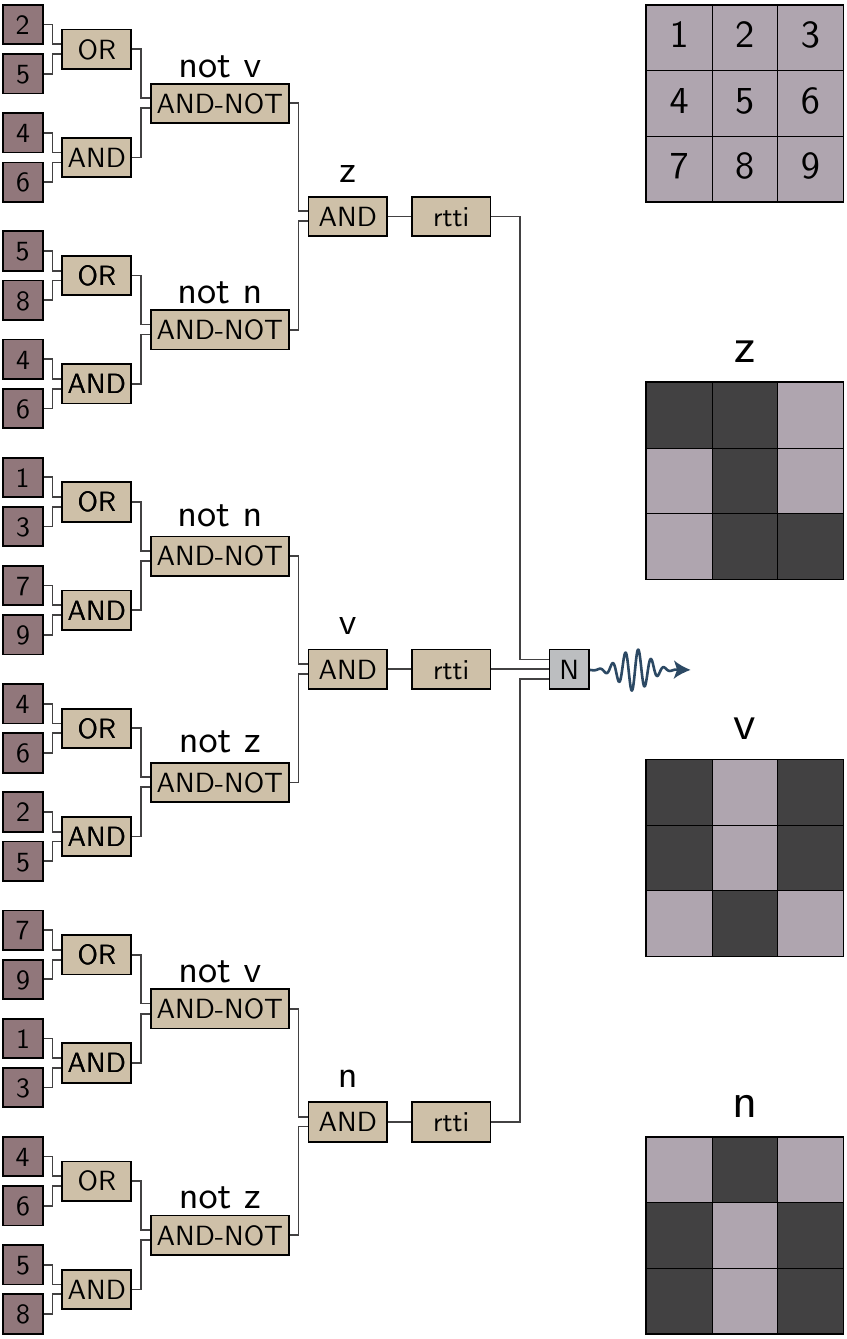}
\caption{\label{fig:neuron__nine_pixel_schematic}Schematic of the nine pixel classifier. The dendritic tree is illustrated as are the three classes of inputs, $z$, $v$, and $n$.}
\end{figure}
We now consider an example of employing the phenomenological model to treat a loop neuron with a more elaborate dendritic arbor to solve the standard benchmark problem of the nine-pixel image classifier \cite{prezioso2015training}. Figure \ref{fig:neuron__nine_pixel_schematic} shows the nine-pixel input and the manner in which the pixels can be argued to depict three letters: `$z$', `$v$', and `$n$'. The problem as formulated for small neural networks is to train the network to identify the image as the correct letter even when the state of any one pixel is allowed to switch. This problem is often solved with small neural networks trained with conventional techniques such as backpropagation. Here we solve the problem with a single neuron with a small dendritic tree, showing that such a tree can serve a similar role to a feed-forward neural network \cite{poirazi2003pyramidal} and to illustrate how even relatively simple loop neurons become sophisticated computational processors, as inspired by their biological counterparts \cite{koch2000role}. We also achieve the desired neural operation through inspection without a training algorithm, in the spirit of biological vision systems that ``hard-code'' basic computational primitives, such as Gabor filters matched to common spatial frequencies in natural scenes \cite{sterling2015principles}.

The arbor that solves the problem is shown in Fig.\,\ref{fig:neuron__nine_pixel_schematic}. Each pixel is input to a synapse as a single spike event: a spike event occurs at synapse $i$ if pixel $i$ is active in the image under consideration, and no spike event is input at that synapse if the corresponding pixel is not active. The various dendrites are labeled by the logic operations they perform (Sec.\,\ref{sec:synapses}). The knowledge accessible at various stages of the tree is written above the dendrites whose signal represents that information. The basic reasoning that solves the problem can followed by stepping through the tree. For example, starting at the top left, if pixel two or five is active but pixels four and six are not both active, the letter cannot be $v$. Similar reasoning can be applied to rule out $n$. If it is not $v$ and it is not $n$, it must be $z$. At the last stage of the tree before the soma, three different dendrites know whether their letter is present. By assigning these three dendrites different time constants ($\alpha$ parameters in Eq.\,\ref{eq:main_equation__ode}), they each evoke different numbers of pulses from the neuron. An output of one spike event from the neuron informs us a $z$ has been presented, two correspond to $v$, and three to $n$. The neuron under consideration is shown to accomplish the task even in the presence of 1\,ns timing jitter due to the transmitter circuit and quantum dot light source (Appendices \ref{apx:source} and \ref{apx:transmitter}).

\begin{figure}[h!]
\includegraphics[width=8.6cm]{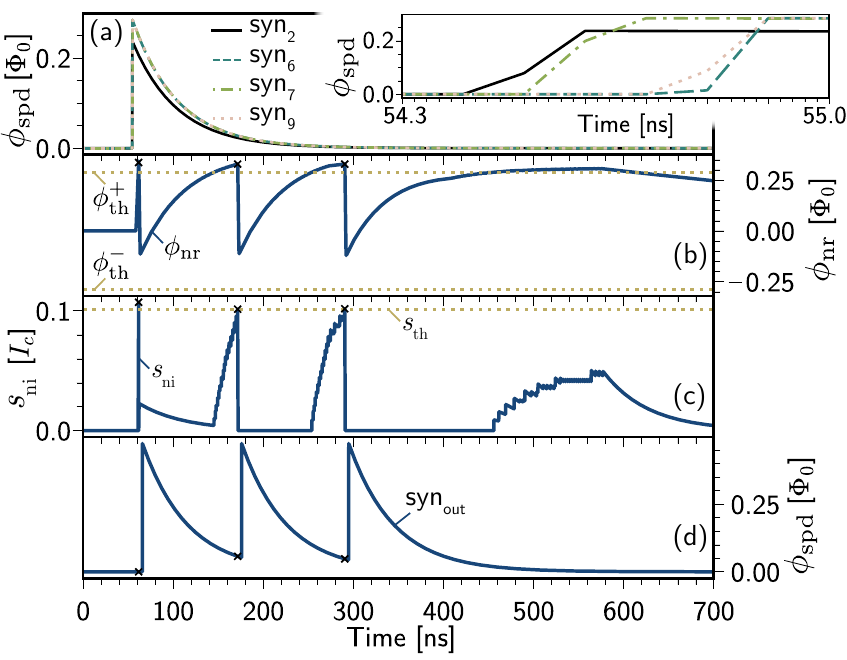}
\caption{\label{fig:neuron__nine_pixel_time_traces}Time traces for the nine-pixel classifier neuron for the case of $n_4$, the image of $n$ with pixel four switched. (a) The input synaptic flux from all four active synapses. The inset shows higher temporal resolution to display the timing jitter due to the model of the quantum-dot light source. (b) The total applied flux to the neuronal receiving loop, including the refractory contribution. (c) The integrated current in the neuronal integration loop. (d) The flux generated by the downstream SPD receiving photons upon neuronal firing.}
\end{figure}
Time traces from a single classification instance ($n_4$) are shown in Fig.\,\ref{fig:neuron__nine_pixel_time_traces}. The four input synapse events are shown in Fig.\,\ref{fig:neuron__nine_pixel_time_traces}(a), and the timing jitter due to the transmitter circuit and light source is evident in the inset. The total flux input to the neuronal receiving loop is shown in Fig.\,\ref{fig:neuron__nine_pixel_time_traces}(b), as are the thresholds for activity due to positive and negative flux. The accumulated signal in the neuronal integration loop is shown in Fig.\,\ref{fig:neuron__nine_pixel_time_traces}(c), along with the value of the neuronal integration loop threshold. The times of neuronal spike events are shown as black crosses. In this model, the neuron is given one output synapse, and the flux present at the output of that synapse due to photon detection events from neuronal firing are shown in Fig.\,\ref{fig:neuron__nine_pixel_time_traces}(d).

\begin{figure}[h!]
\includegraphics[width=8.6cm]{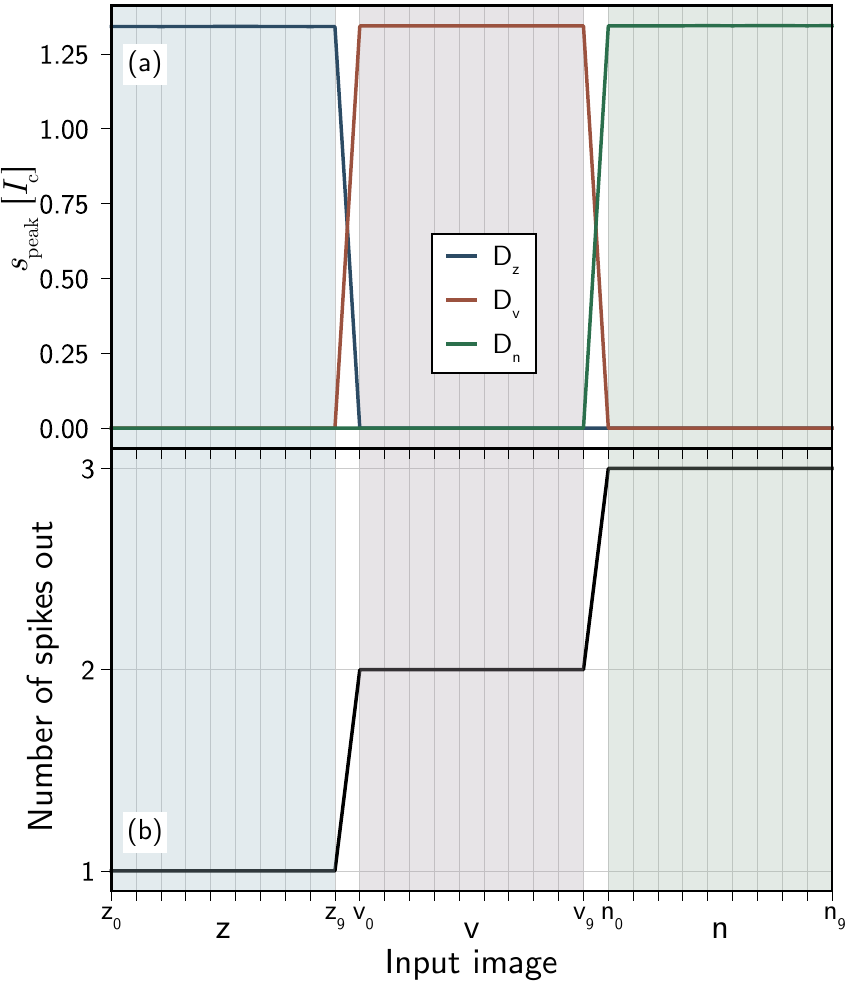}
\caption{\label{fig:neuron__nine_pixel_data}Output of the nine-pixel classifier neuron versus input image. (a) The peak of the integrated signal in the final dendrites that represent the three letters. (b) The number of spikes out as a function of the input image.}
\end{figure}
The full set of 30 inputs is shown in Fig.\,\ref{fig:apx__nine_pixel_drive_matrices} in Appendix \ref{apx:nine_pixel}. The results of all instances of the classification task are shown in Fig.\,\ref{fig:neuron__nine_pixel_data}. The maximum level of integrated signal in all three final dendrites is shown in Fig.\,\ref{fig:neuron__nine_pixel_data}(a) for all thirty presented images. The number of spikes produced by the neuron is shown in Fig.\,\ref{fig:neuron__nine_pixel_data}(b). It can be seen that no signal is generated in any of the dendrites except the one corresponding to the presented letter, even in the presence of one-pixel noise, and the neuron correctly produces the number of spike outputs to communicate the result of its calculation to other neurons that may be present in the network.

\section{\label{sec:discussion}Discussion}
We have seen that the computational circuitry of loop neurons consists of a network of interacting dendrites. Within this framework, a dendrite is a SQUID with output coupled to a current-storage loop. The bias point of the SQUID determines the threshold input flux required to initiate activity, and the storage capacity of the output loop provides a saturating nonlinearity. The various circuit parameters provide means to adjust the response characteristics across broad operating ranges. Learning and homeostatic adaptation can be accomplished with similar circuits providing feedback, dynamically adjusting bias points through coupled flux that can be stored perpetually and adjusted in small or large increments. The final stage of dendritic processing culminates in the soma, the neuron cell body. When this final dendrite reaches threshold, something different happens with regard to physical hardware: an amplifier drives a semiconductor light source. The threshold signal is the final stage of the computational process; the production of light is a binary action potential, and the physical transduction to photons is chosen to enable communication to many destinations across length scales that result in wiring parasitics that render electronic signals impracticable. A neuron is a network of coupled dendrites with local electronic communication. A spiking neural network extends interconnectivity to longer distances with optical communication.

Simulating loop neurons with the phenomenological model brings significant speed increases relative to the conventional method of simulating the first-principles circuit model. The speed of simulation can increase by a factor of ten thousand while retaining a $\chi^2$ of $10^{-4}$ when simulating a single dendrite for 12\,\textmu s. When simulating a large network, the circuit-model approach is impracticable due to the time required to run calculations over even a relatively short duration of network activity. The phenomenological model thus enables design of more components, exploration of a wider range of parameter space, and treatment of larger networks than would otherwise be possible. Still, simulating 12\,\textmu s of activity from a single dendrite requires around 1\,s of CPU time. A physically instantiated network on a small chip (roughly 1\,cm\,$\times$\,1\,cm) would comprise roughly 10,000 neurons and half a million synapses and dendrites. Linearly extrapolating the time requirements, it would take over a year to simulate 1\,ms of activity of this network on the same workstation used in this work. Two conclusions can be drawn: 1) the phenomenological model must be used as a tool to facilitate further increases in speed of design, and 2) constructing the physical hardware of loop neurons could bring a computational platform of immense power and utility. Regarding the first point, by using the phenomenological model to design a variety of dendrites, coupled-dendrite computational blocks, and a variety of neurons, libraries of optimized components can be assembled. The model can be used to efficiently design these components and reduce their behavior to stored transfer functions to achieve a further level of abstraction that can be used to model behavior of larger networks without explicitly stepping through the ODEs, as was done for the logic gates, burst, and rate transfer functions in Sec.\,\ref{sec:dendrites}. Regarding the second point, it would take a year on a workstation with a 3.7\,GHz processor to simulate one millisecond of a network that would fit on a chip of similar size and power consumption to the processor, including cryogenic cooling. The increase in computational speed that would result from realizing SOEN hardware would be 32\,billion for even a modest network. SOENs therefore appear to offer a path to neuromorphic supercomputing.

If the aspiration of loop neurons is to constitute systems of exceptional scale and complexity, might the simplicity of the dendritic building block limit the dynamical repertoire? We hope this model helps answer this question. It is known that even simple systems with simple rules for propagation, such as cellular automata, can give rise to behavior of great sophistication \cite{wolfram1983statistical,wolfram2002new}. It has been argued that similar concepts can be applied as a starting point for physics, with the rich, natural world emerging based on the interactions of fundamental nodes \cite{wolfram2020class}. Similarly, elements as simple as two-state spins interacting with nearest neighbors (Ising models) can give rise to phase transitions and critical phenomena, including crucial long-range correlations \cite{nishimori2010elements,shankar2017quantum} as well as attractor dynamics for associative memory storage and retrieval \cite{little1974existence,hopfield1982neural,amit1985spin,amit1989modeling}. By comparison, the dendrites studied here are multi-dimensional and nuanced. The nonlinear spatio-temporal convolutions occurring in each neuron's synapto-dendritic tree provide a deep repository that can inform the neuron's behavior. Transmission of action potentials to many destinations at light speed enable complex network topologies far beyond nearest-neighbor interactions. The use of superconducting circuits allows signal retention across a broad range of time scales through the choice of the $\alpha$ parameters (the $\alpha$ distribution) that specify the leak rates in Eq.\,\ref{eq:main_equation__ode}, as has been demonstrated experimentally \cite{khan2022superconducting}. The range of integration loop inductances (the $\beta$ distribution) in conjunction with the bias currents will establish the dynamic range of the network. With near and distant connectivity as well as a wide spectrum of dissipation times and an enormous dynamic range, SOENs are likely to achieve the long-range spatiotemporal correlations present in the critical states that optimize neural information processing \cite{kinouchi2006optimal,shew2009neuronal,shew2011information,dambre2012information}. The hierarchy of spatial connectivity in conjunction with the hierarchy of information retention times and response magnitudes appears excellent for enabling the fractal use of space and time that supports information integration and cognition \cite{honey2007network,kitzbichler2009broadband,bassett2010efficient,bressler2010large,he2010temporal,rubinov2011neurobiologically}. Yet the simplicity of constructing the majority of computational grey matter from similar building-block components brings an advantage in modeling and technological implementation. 

The significance of dendrites in neural computation is well documented \cite{mel1994information,poirazi2003pyramidal,london2005dendritic,hawkins2016neurons,sardi2017new,kirchner2022emergence,beniaguev2021single}, yet the incorporation in neuromorphic hardware has received proportionally less attention \cite{elias1993artificial,arthur2004recurrently,schemmel2017accelerated,kaiser2022emulating,mead2022}. The dendrites studied here may provide valuable means for implementing credit assignment in training algorithms that utilize local information in conjunction with population information \cite{guerguiev2017towards,richards2019dendritic,mikulasch2021local,hodassman2022efficient,mikulasch2022dendritic}, potentially leveraging much of what is know from neuroscience about the key role of dendrites in learning \cite{poirazi2001impact,magee2005plasticity,losonczy2008compartmentalized,sjostrom2008dendritic,mel2017synaptic,bono2017modeling,payeur2021burst,d2022compartmentalized} and overcoming a major obstacle to widespread adoption of spiking neural networks for artificial intelligence. The presence of continuous dendritic signals that are continuous in time without erasure following a spike may provide new methods for training spiking neural networks that are not available based on spiking activity alone. Perhaps these continuous signals will be useful for constructing cost functions and training networks with variants of the backpropagation technique, a feat which has been difficult in the case of spiking neurons outside of the rate-coding domain, and a subject which is significant in bridging machine learning and neuroscience \cite{marblestone2016toward}.

While the model presented here is interesting to explore, it is intended as a tool in a larger project. The objective of such research at this stage is to determine if SOENs are indeed as promising for future study and worthy of appreciable investment as our hypothesis contends. We hope this model will help answer several questions: Do the neuromorphic circuit principles demonstrated in these superconducting optoelectronic embodiments bestow systems with the dynamic interplay of structure and function that scaffolds successful neural systems? Do device and circuit features such as the achievable breadth of the $\alpha$ and $\beta$ distributions, the particular dendritic nonlinearities, and the available forms of plasticity bring benefits in support of cognition? Does refraction of the neuron without erasure of information in the dendritic tree offer advantages in information processing, lead to new forms of neural coding, or offer new means of training? How should the structure of the dendritic trees and the structure of the network be co-designed to achieve desired operations? The objective of the model presented here is to enable computational studies that answer these questions.

With the phenomenological model as a guide, we can compare and contrast loop neurons with other complex systems. Spin systems have received considerable attention over decades due to both their simplicity as well as the emergence of interesting phenomena such as phase transitions and spontaneous symmetry breaking. Common spin models include the Ising model, where each spin $s$ takes a binary, scalar value of -1 or 1. Coupling is typically between nearest neighbors and is also a scalar value, $J_{ij}$, typically with the symmetry $J_{ij} = J_{ji}$. The $x-y$ model and the Heisenberg model extend the spin to have two or three vector components, maintaining $|s| = 1$. In the present context, we can think of the current stored in the integration loop of dendrite $i$, $s_i$, as a spin. It is more like an Ising model than an $x$-$y$ or Heisenberg in that it is a scalar value, but in general it is analog instead of binary and can take a large number of values between zero and one. Coupling is through the scalar quantities $J_{ij}$ (Eq.\,\ref{eq:main_equation__coupling}) and is manifest through transformers. In this case, there is no requirement that $J_{ij} = J_{ji}$, and the spatial limitations regarding which spins can couple are relaxed. Coupling is mediated by a transformer circuit rather than an exchange interaction, so flexibility in the adjacency matrix is available. The values of $J_{ij}$ are fixed in time, but the sign can be positive or negative so that excitation or inhibition are possible. While the $J_{ij}$ are fixed in circuit fabrication, the interaction between any two spins is dynamically reconfigurable either by an external influence (experimenter) or through network activity (plasticity) by changing applied bias currents to change the $r$-shell on which the dendrite resides. Furthermore, the magnitude of the spins in loop neurons is not fixed in time, but is a dynamical quantity that grows through interactions and decays passively, following the leaky integrator formalism. In general, the threshold for one spin to induce another is non-zero, an important nonlinearity familiar from neuroscience.

Perhaps the most interesting connection to spin systems is to glasses in which disorder and frustration play central roles. Frustration in spin glasses occurs when any spin cannot achieve a configuration in which the interaction energy to all of its neighbors is minimized. This phenomenon leads to energy landscapes with many peaks and valleys and a large number of metastable configurations \cite{binder1986spin}. Competing interactions are a hallmark of complex systems and result in interesting and often computationally useful dynamics as the system traverses the landscape of metastable states \cite{rabinovich2001dynamical}. In the present case, the dendritic tree of each loop neuron can be compared to a spin glass in which each dendrite has multiple interactions with competing signs of coupling. Frustration can be quantified for dendrite $i$ as
\begin{equation}
\label{eq:frustration_dendrite}
f_i = \frac{ \left( \sum_{j = 1}^{n_+} J_{ij}\,s_j \right) \; \left( \sum_{j = 1}^{n_-} \left|J_{ij}\right|\,s_j \right) }{ \left( \sum_{j = 1}^{n_+} J_{ij}\,s_j^\mathrm{max} \right) \; \left( \sum_{j = 1}^{n_-} \left|J_{ij}\right|\,s_j^\mathrm{max} \right) },
\end{equation}
where the sum to $n_+$ runs over the excitatory inputs, the sum to $n_-$ runs over inhibitory inputs, and $s_j^\mathrm{max}$ is the saturation level of the signal in dendrite $j$. This value for frustration is zero when the sum of excitatory or inhibitory inputs is zero and reaches a maximum value of one when all excitatory and inhibitory inputs are at their maximum values. The frustration of neuron $p$ can be quantified by summing over its arbor, $\mathcal{F}_p = \sum_i f_i$. Each neuron's dendritic tree has qualitative features in common with a dynamical spin glass, with randomness and transitory inhibitory and excitatory interactions leading to ever-shifting states of competition and frustration, as is also evident in biological neural systems \cite{gollo2014frustrated}. The frustration of the network can be obtained by summing once more over neurons, $\mathscr{F}_N = \sum_p \mathcal{F}_p$.  Future work will explore the relationship between frustration, critical phenomena, and useful neural computation in the context of SOENs. 

Above the level of the dendritic tree, at the level of coupled neurons, analogies to spin systems are less relevant, and comparison to pulse-coupled oscillators and their more complex cousins, spiking neurons, are straightforward. Superconducting optoelectronic loop neurons are spiking neurons that produce binary, pulsatile communication signals upon being driven to an internal threshold. The optical communication signals that facilitate inter-neuron interaction are not strictly based on spatial location (nearest neighbors, next-nearest neighbors, etc.), but can be engineered with great flexibility through an adjacency matrix physically realized by photonic interconnects. As compared to typical point neurons, a complex, active dendritic tree appears central to the function and construction of loop neurons \cite{primavera2021active}. Yet at the base of that tree resides a soma that sums inputs and registers a threshold. While the signal in that final integration loop of the soma is purged upon reaching threshold, the signals stored in the loops of the rest of the arbor need not change upon neuronal firing. As compared to archetypal leaky integrate-and-fire neurons, the lack of erasure following firing is distinct. This trait allows a single synapse event to push a neuron into a dynamical orbit and allows information retention regardless of firing state.

In computational neuroscience a list of discrete spike times is sometimes considered a complete description of the behavior of a neuron or population in a given context. Such a list is available for networks of loop neurons, yet continuous variables specifying the states of all dendritic loops can also be used to gain more granular information about the system, much like retaining the time-continuous values of the membrane potentials on all dendrites of all neurons throughout the network. While loop neuron spike times, bursts, and rates are crucial to network activity, the state of any given neuron or population may also be specified by the signal in all loops, providing richer depiction of state space, high temporal resolution even between spike events, and continuous time-series representation suitable for defining distances and overlaps between network states as well as correlation functions between components at different times. The state of dendrite $i$ attached to neuron $p$ is specified by $s_i$. This state description can be extended to neurons as $S_p = [s_1,\dots,s_n]_p$: the state of the neuron is the state of all of its dendrites arranged as a vector. The state of the network can then be defined in full detail as the full vector including the state of all dendrites, $\mathcal{S} = [S_1,\dots,S_N]$, or the state of the each neuron can be reduced to an average for computational efficiency when necessary. The distance between two states can then be defined as the Euclidean distance between two vectors, applicable to single neurons or networks. Such a metric is valuable in many contexts, including in quantifying sensitivity to variation in initial conditions to determine when a network is in a chaotic regime.

We can then discuss correlations between dendrites with themselves at different times, with other dendrites in the same neuronal arbor, and with other dendrites attached to different neurons. For example, a self-correlation function of the form
\begin{equation}
\label{eq:correlation_dendrite_self}
G_i(t,t') = s_i(t)\,s_i(t')
\end{equation}
may be useful for identifying temporal coherence related to recurrent activity. Similarly, a cross-correlation function between two dendrites,
\begin{equation}
\label{eq:correlation_dendrite_cross}
G_{ij}(t,t') = s_i(t)\,s_j(t'),
\end{equation}
may be useful in identifying functional coalitions within a neuron with dendrites $i$ and $j$ within the same arbor, or it may be useful for identifying synchronized activity across the network when they are attached to different neurons. Extending this concept to neurons, we can write down a temporal correlation function of a neuron with itself,
\begin{equation}
\label{eq:correlation_neuron_self}
G_p(t,t') = S_p(t)\,\cdot\,S_p(t'),
\end{equation}
or with other neurons,
\begin{equation}
\label{eq:correlation_neuron_cross}
G_{pq}(t,t') = S_p(t)\,\cdot\,S_q(t').
\end{equation}
In Eqs.\,\ref{eq:correlation_neuron_self} and \ref{eq:correlation_neuron_cross}, the vector dot product can be taken to reduce the information to a single, scalar metric for temporal correlations. The cross-correlation function of Eq.\,\ref{eq:correlation_neuron_cross} contains information about correlations across space and time, as does the dendritic cross-correlation function of Eq.\,\ref{eq:correlation_dendrite_cross} on a shorter length scale. Such functions inform us as to how inputs are represented and stored in correlations, and from these functions we can obtain quantities analogous to the susceptibility which provide information about phase transitions and critical phenomena. These concepts may be further extended to a continuum field theory, potentially conducive to formal analysis that is gaining traction in neural-network applications \cite{helias2020statistical,halverson2021neural}.

In spin \cite{shankar2017quantum}, neural \cite{friston2010free}, and other complex systems, free energy is a powerful concept capturing a system's competing drives to minimize energy and maximize entropy. In an elementary construction, the free energy is given by $\mathcal{F} = \mathcal{U} - T\mathcal{S}$, with $\mathcal{U}$ the total energy of the system, $T$ the temperature, and $\mathcal{S}$ the entropy. Within the present framework, the energy of a single dendrite is given by $L\,I^2/2 \propto \beta s^2$, and the total energy can be obtained by summing over dendrites. The entropy can be obtained from the logarithm of the number of microstates at a given value of total energy, which is a well-defined if tedious problem in combinatorics. Alternatively, an approach in terms of a partition function may be tractable. The temperature of the system may be conceived as the actual, physical temperature in units of kelvins, with the $T\mathcal{S}$ term growing due to stochastic switching of JJs close to $I_c$, or temperature may be engineered as a more abstract quantity introduced via random photonic inputs to synapses. The conceptual foundation of the loop-neuron system in terms of simple component state variables as formulated with this phenomenological model offers multiple routes to connect to a deeper free-energy formulation.

From this list of comparisons and extensions it is evident that loop neurons and superconducting optoelectronic networks have potential to lead to physical systems of immense complexity and useful computational functionality. The phenomenological model developed here may serve in the near term to facilitate design of useful circuits and networks for computation. In the longer term the framework may enable more complete theoretical development of the hardware as a complex physical system, as a tool for artificial intelligence, as a platform for hypothesis testing in neuroscience, and as a means to devise systems with intelligence exceeding our own.

\appendix
\section{\label{apx:circuit_equations_RI}Circuit Equations for the RI Dendrite}
\begin{figure}[tbh]
\includegraphics[width=8.6cm]{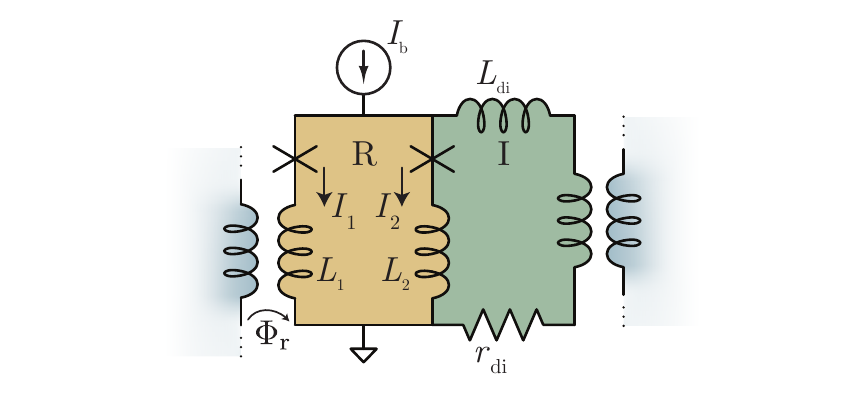}
\caption{\label{fig:apx__circuit_RI}Circuit diagram of the RI dendrite with relevant circuit parameters labeled.}
\end{figure}
A circuit diagram of the RI dendrite is shown in Fig.\,\ref{fig:apx__circuit_RI}. Kirchoff's current law gives $I_\mathrm{b} = I_1 + I_2 + I_\mathrm{di}$. Summing voltages around the R loop gives
\begin{equation}
\label{eq:circuit_equations__RI__1}
\frac{\Phi_0}{2\pi}\frac{d\,\delta_1}{dt} + L_1\frac{d\,I_1}{dt} - L_2\frac{d\,I_2}{dt} - \frac{\Phi_0}{2\pi}\frac{d\,\delta_2}{dt} - \frac{d\,\Phi_\mathrm{r}}{dt} = 0,
\end{equation}
where $\delta_i$ is the phase of the superconducting wave function across the $i$th junction. We introduce the following dimensionless variables \cite{clarke2006squid}:
\begin{equation}
\label{eq:dimensionless_variables}
\begin{split}
i_x &\equiv I_x/I_c, \\[10pt]
\beta_x &\equiv \frac{2\pi L_x I_c}{\Phi_0}, \\[10pt]
\phi &\equiv \Phi_\mathrm{r}/\Phi_0, \\[10pt]
\alpha &\equiv r_\mathrm{di}/r_\mathrm{j}, \\[10pt]
\tau &\equiv \frac{2\pi r_\mathrm{j} I_c}{\Phi_0}\,t \equiv \omega_c t.
\end{split}
\end{equation}
$\omega_c$ is the characteristic Josephson frequency, and $r_\mathrm{j}$ is the Josephson junction shunt resistance in the resistively and capacitively shunted junction (RCSJ) model \cite{van1998principles,kadin1999introduction,tinkham2004introduction,clarke2006squid}. A dimensionless current $i_x$ is defined for each branch of the circuit ($i_\mathrm{b}$, $i_1$, $i_2$, and $i_\mathrm{di}$), while a dimensionless screening parameter, $\beta_x$, is defined similarly for each inductor. We also use the notation $\beta_\mathrm{r} = \beta_1+\beta_2$ and $\bar{\beta} = \beta_1\beta_2 + \beta_1\beta_\mathrm{di} + \beta_2\beta_\mathrm{di}$. We assume all JJs have identical critial current, $I_c$. Algebraic manipulation of Eq.\,\ref{eq:circuit_equations__RI__1} leads to
\begin{equation}
\label{eq:circuit_equations__RI__2}
\frac{d\,i_1}{d\tau} = \frac{1}{\beta_\mathrm{r}} \left( \frac{d\,\delta_2}{d\tau} - \frac{d\,\delta_1}{d\tau} + 2\pi\frac{d\,\phi}{d\tau} \right) + \frac{\beta_2}{\beta_\mathrm{r}}\left( \frac{d\,i_\mathrm{de}}{d\tau} - \frac{d\,i_\mathrm{di}}{d\tau} \right),
\end{equation}
where the relation $d/dt = \omega_c d/d\tau$ has been used. Summing voltages around the I loop gives
\begin{equation}
\label{eq:circuit_equations__RI__3}
\frac{\Phi_0}{2\pi}\frac{d\,\delta_2}{dt} + L_2 \frac{d\,I_2}{dt} - L_\mathrm{di} \frac{d\,I_\mathrm{di}}{dt} - r_\mathrm{di} I_3 = 0.
\end{equation}
Moving to dimensionless units and making use of Eq.\,\ref{eq:circuit_equations__RI__2} we can state the primary equation of motion governing the circuit of Fig.\,\ref{fig:apx__circuit_RI}:
\begin{equation}
\label{eq:circuit_equations__RI__4}
\bar{\beta}\,\frac{d\,i_\mathrm{di}}{d\tau} = \beta_1 \frac{d\,\delta_2}{d\tau} + \beta_2 \frac{d\,\delta_1}{d\tau} - 2\pi \frac{d\,\phi}{d\tau} + \beta_1 \beta_2 \frac{d\,i_\mathrm{b}}{d\tau} - \alpha \beta_\mathrm{r} i_\mathrm{di}.
\end{equation}
Equation \ref{eq:circuit_equations__RI__4} must be solved in a system with an ODE for each of the JJs of the form
\begin{equation}
\label{eq:circuit_equations__RI__5}
\beta_c \frac{d^2\,\delta_x}{d\tau^2} = i_x - \mathrm{sin}(\delta_x) - \frac{d\,\delta_x}{d\tau}.
\end{equation}
Here $\beta_c = 2\pi I_c r_\mathrm{j}^2 c_j/\Phi_0$ is the Stewart-McCumber parameter, with $c_j$ the capacitance of a JJ in the circuit. Such an expression can be derived in the framework of the RCSJ model \cite{van1998principles,kadin1999introduction,tinkham2004introduction,clarke2006squid}. This second-order ODE must be converted to two first order ODEs for each of the JJs. The system of coupled ODEs combining Eqs.\,\ref{eq:circuit_equations__RI__4} and \ref{eq:circuit_equations__RI__5} thus results in a system of five ODEs that can be solved numerically for arbitrary parameter values, bias conditions, and flux drives. After the system of ODEs has been integrated in time, $i_1$ can be obtained as
\begin{equation}
\label{eq:circuit_equations__RI__6}
i_1 = \frac{1}{\beta_\mathrm{r}} \left( \delta_2 -\delta_1 + 2\pi\phi \right) + \frac{\beta_2}{\beta_\mathrm{r}}\left( i_\mathrm{de} - i_\mathrm{di} \right),
\end{equation}
and $i_2 = i_\mathrm{b}-i_1-i_3$. This numerical model has been implemented in Python and the system of ODEs solved with the \code{solve\_ivp} function from SciPy, employing a Runge-Kutta integration method with fifth-order accuracy and an adaptive time grid.

In constructing the phenomenological model, we have essentially replaced Eq.\,\ref{eq:circuit_equations__RI__4} with the leaky integrator Eq.\,\ref{eq:main_equation__ode} and abstracted away the driving terms into the phenomenological rate function. By comparing Eq.\,\ref{eq:circuit_equations__RI__4} and \ref{eq:main_equation__ode} and noting that in the main text we have made the replacement $i_\mathrm{di}\,\rightarrow\,s$, we see that, roughly speaking, 
\begin{equation}
\label{eq:phases_to_rate_array}
\beta_1 \frac{d\,\delta_2}{d\tau} + \beta_2 \frac{d\,\delta_1}{d\tau} - 2\pi \frac{d\,\phi}{d\tau} \, \rightarrow \, r(s,\phi; i_\mathrm{b}),
\end{equation}
aside from differences in $\beta$ prefactors, and ignoring the temporal derivative of the bias current. While the two differential equations tracking the current in the dendritic integration loop are qualitatively similar, by moving to the phenomenological model we no longer require tracking the time derivatives of the phases of the JJs and the applied flux, which evolve on the picosecond time scale, and instead require only the use of previously calculated rate arrays as look-up tables. The demonstrated speed benefits arise from this replacement.

\section{\label{apx:circuit_equations_RTTI}Circuit Equations for the RTTI Dendrite}
The primary equation of motion for the RTTI dendrite is
\begin{equation}
\label{eq:circuit_equation__RTTI}
\beta_\mathrm{di}\frac{d\,i_\mathrm{di}}{d\tau} = \frac{d\,\delta_4}{d\tau} - \alpha\,i_\mathrm{di}.
\end{equation}
Equation \ref{eq:circuit_equation__RTTI} must be solved in a system with an ODE of the for given by Eq.\,\ref{eq:circuit_equations__RI__5} for each of the four JJs in the circuit. Upon reducing the second-order odes to first order, the system comprises nine coupled ODEs. The remaining currents in the circuit can be obtained from
\begin{equation}
\label{eq:RTTI_additional_currents}
\begin{split}
i_4 &= \beta_4^{-1} \left( \delta_3 - \delta_4 \right) + i_\mathrm{b3} - i_\mathrm{di}, \\[10pt]
i_3 &= \bar{\beta}^{-1} \left[ \beta_1 \left( \delta_2-\delta_3 \right) + \beta_2 \left( \delta_1-\delta_3 - 2\pi\phi \right) + \beta_1\,\beta_2\,i_\mathrm{b1} \right] \\ 
&+i_\mathrm{b2} + i_\mathrm{b3} - i_4 - i_\mathrm{di}, \\[10pt]
i_2 &= \frac{1}{\beta_1 + \beta_2}\, \left( \delta_1-\delta_2-2\pi\phi \right) \\ 
&+ \frac{\beta_1}{\beta_1 + \beta_2} \, \left( i_\mathrm{b1} + i_\mathrm{b2} + i_\mathrm{b3} - i_3 - i_4 - i_\mathrm{di} \right), \\[10pt]
i_1 &= i_\mathrm{b1} + i_\mathrm{b2} + i_\mathrm{b3} - i_2 - i_3 - i_4 - i_\mathrm{di}.
\end{split}
\end{equation}

\section{\label{apx:phenomenological_model_dimensionless_units}Obtaining the Phenomenological Model in Dimensionless Units}
In this Appendix we begin with slightly different and more explicit variable names that used in the main text. We proceed with cluttered notation through the derivation to reduce ambiguity, and all notation will be simplified at the end.

\begin{figure}[hbt]
\includegraphics[width=8.6cm]{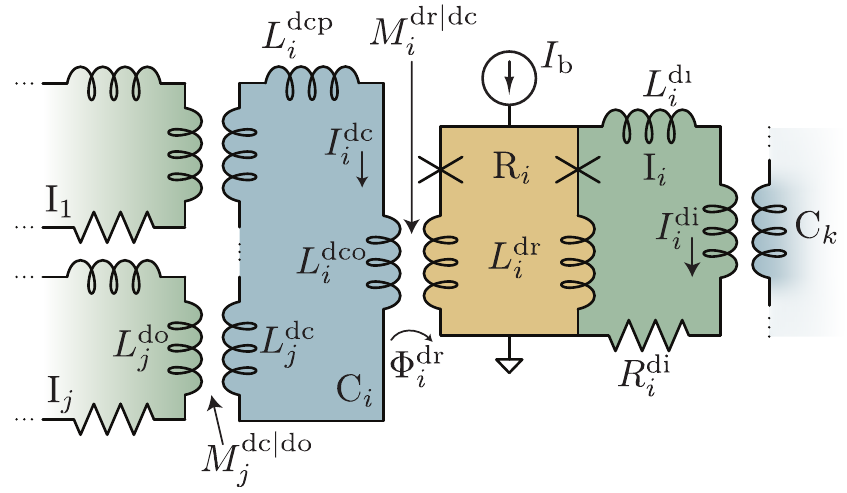}
\caption{\label{fig:apx__circuit_RI__dimensionless}Full circuit diagram for the RI dendrite labeling additional parameters relevant to the derivation of the model in dimensionless units.}
\end{figure}
The key dynamical variable of each dendrite is the integrated current in the DI loop, $I_\mathrm{di}$ (see Fig.\,\ref{fig:apx__circuit_RI__dimensionless}). The foundational postulate of this work is that time-evolution of this quantity can be captured in a phenomenological model of the form
\begin{equation}
\label{eq:phenomenological_model_dimensionless_units__1}
\frac{d\,I_\mathrm{di}}{dt} = I_\mathrm{fq}\,R_\mathrm{fq}\!\left( \Phi_\mathrm{r}, I_\mathrm{di}; I_\mathrm{b} \right) - \frac{I_\mathrm{di}}{\tau_\mathrm{di}},
\end{equation}
where $I_\mathrm{fq} = \Phi_0/L_\mathrm{di}$ is the current associated with each fluxon, $R_\mathrm{fq}$ is the rate of fluxon production as a function of $\Phi_\mathrm{r}$, the flux applied to the receiving loop, and $I_\mathrm{di}$, the current accumulated in the integration loop. $\tau_\mathrm{di} = L_\mathrm{di}/R_\mathrm{di}$ is the time constant governing the leak of the integrated signal. As in Appendix \ref{apx:circuit_equations_RI}, we convert this equation to dimensionless units:
\begin{equation}
\label{eq:phenomenological_model_dimensionless_units__main_ode}
\beta_\mathrm{di}\frac{d\,i_\mathrm{di}}{d\tau} = r_\mathrm{fq} \left( \phi, i_\mathrm{di}; i_\mathrm{b} \right) - \alpha_\mathrm{di}\,i_\mathrm{di}.
\end{equation}
The dimensionless parameters $i_\mathrm{di}$, $\beta_\mathrm{di}$, $\alpha_\mathrm{di}$, and $\tau$ have been defined in Eq.\,\ref{eq:dimensionless_variables}. The dimensionless rate of fluxon production is given by
\begin{equation}
\label{eq:phenomenological_model_dimensionless_units__r_fq}
r_\mathrm{fq} = \frac{R_\mathrm{fq}}{\omega_c/2\pi},
\end{equation}
with $\omega_c = 2\pi r_\mathrm{j} I_c/\Phi_0$. 

Each dendrite in the system will obey an equation of the form of Eq.\,\ref{eq:phenomenological_model_dimensionless_units__main_ode}, and knowledge of the system is complete when the values of $i_\mathrm{di}$ for all dendrites is obtained. However, to solve the complete system of coupled ODEs we need to calculate how the values of $i_\mathrm{di}$ from input dendrites convert to coupling flux in the receiving dendrite. Thus, the next step is to determine the total applied flux to the DR loop of dendrite $i$ in terms of the values of $i_\mathrm{di}$ present in the input dendrites, which we index with $j$. To make space for indices, we move the part labels to superscripts. 

The total applied flux to the DR loop of dendrite $i$ is
\begin{equation}
\label{eq:phenomenological_model_dimensionless_units__total_DR_flux}
\Phi_i^\mathrm{dr} = M_i^\mathrm{dr|dc}\,I_i^\mathrm{dc}.
\end{equation}
The notation $M_i^\mathrm{dr|dc}$ refers to the mutual inductance coupling from the DC loop to the DR loop on dendrite $i$. Explicitly, $M_i^\mathrm{dr|dc} = k_i^\mathrm{dr|dc}\,\left( L_i^\mathrm{dr} \, L_i^\mathrm{dco} \right)^{1/2}$, with $k_i^\mathrm{dr|dc}$ the transformer efficiency. In Eq.\,\ref{eq:phenomenological_model_dimensionless_units__total_DR_flux}, $I_i^\mathrm{dc}$ is the net induced current in the collection coil input to dendrite $i$. This current obeys the differential equation
\begin{equation}
\frac{d\,\Phi_i^\mathrm{dc}}{dt} - L_i^\mathrm{dc}\,\frac{d\,I_i^\mathrm{dc}}{dt} = 0,
\end{equation}
which can be integrated to obtain
\begin{equation}
\label{eq:phenomenological_model_dimensionless_units__I_dc}
I_i^\mathrm{dc} = \frac{1}{L_i^\mathrm{dc}} \, \Phi_i^\mathrm{dc},
\end{equation}
where $L_i^\mathrm{dc} = L_i^\mathrm{dcp} + L_i^\mathrm{dco} + \sum_{j=1}^n L_j^\mathrm{dc}$, with $L_j^\mathrm{dc}$ the inductance of the transformer input to the DC coil connecting dendrite $j$ to dendrite $i$, $L_i^\mathrm{dco}$ the output inductance of the DC loop, which is the input to the transformer to the DR loop, and $L_i^\mathrm{dcp}$ any additional parasitic inductance on the DC coil. The sum is over the $n$ inputs to the DC loop of dendrite $i$. The total flux applied to the DC loop is 
\begin{equation}
\label{eq:phenomenological_model_dimensionless_units__Phi_dc}
\Phi_i^\mathrm{dc} = \sum_{j=1}^n M_j^\mathrm{dc|do}\,I_j^\mathrm{di},
\end{equation}
with $M_j^\mathrm{dc|do} = k_j^\mathrm{dc|do}\,(L_j^\mathrm{dc}\,L_j^\mathrm{do})^{1/2}$. Inserting Eq.\,\ref{eq:phenomenological_model_dimensionless_units__Phi_dc} into Eq.\,\ref{eq:phenomenological_model_dimensionless_units__I_dc} we have
\begin{equation}
\label{eq:phenomenological_model_dimensionless_units__I_dc_alt}
I_i^\mathrm{dc} = \frac{1}{L_i^\mathrm{dc}}\,\sum_{j=1}^n k_j^\mathrm{dc|do}\,(L_j^\mathrm{dc}\,L_j^\mathrm{do})^{1/2}\,I_j^\mathrm{di}.
\end{equation}
Now inserting Eq.\,\ref{eq:phenomenological_model_dimensionless_units__I_dc_alt} into Eq.\,\ref{eq:phenomenological_model_dimensionless_units__total_DR_flux} we obtain
\begin{equation}
\Phi_i^\mathrm{dr} = k_i^\mathrm{dr|dc} \frac{(L_i^\mathrm{dr}\,L_i^\mathrm{dco})^{1/2}}{L_i^\mathrm{dc}}\sum_{j=1}^n k_j^\mathrm{dc|do} (L_j^\mathrm{dc}\,L_j^\mathrm{do})^{1/2}\,I_j^\mathrm{di}.
\end{equation}
We now define the dimensionless quantities
\begin{equation}
\label{eq:phenomenological_model_dimensionless_units__final_dimensionless_quantities}
\begin{split}
s_i &\equiv i_i^\mathrm{di} = \frac{I_i^\mathrm{di}}{Ic}, \\[10pt]
\phi_i &\equiv \phi_i^\mathrm{dr} = \frac{\Phi_i^\mathrm{dr}}{\Phi_0}, \\[10pt]
J_{ij} &\equiv \frac{k_i^\mathrm{dr|dc}\,k_j^\mathrm{dc|do}}{2\pi}\,\frac{(\beta_i^\mathrm{dr}\,\beta_i^\mathrm{dco}\,\beta_j^\mathrm{dc}\,\beta_j^\mathrm{do})^{1/2}}{\beta_i^\mathrm{dc}},
\end{split}
\end{equation}
where the various $\beta$ parameters are related to the inductances as in Eq.\,\ref{eq:dimensionless_variables}. For simplicity of notation we also specify $\beta_i^\mathrm{di} \rightarrow \beta_i$ and $r_\mathrm{fq} \rightarrow r$, and we drop the subscript $i$ for notational simplicity. We can now recast the phenomenological model of Eq.\,\ref{eq:phenomenological_model_dimensionless_units__main_ode} in its final form:
\begin{equation}
\label{eq:phenomenological_model_dimensionless_units__final_ode}
\beta \frac{d\,s}{d\tau} = r \left( \phi, s; i_b \right) - \alpha\,s,
\end{equation}
where the coupling flux from the $n$ dendrites indexed by $j$ into dendrite $i$ is given by
\begin{equation}
\label{eq:phenomenological_model_dimensionless_units__final_coupling_flux}
\phi_i = \sum_{j=1}^n\,J_{ij}\,s_j.
\end{equation}

Equations \ref{eq:phenomenological_model_dimensionless_units__final_ode} and \ref{eq:phenomenological_model_dimensionless_units__final_coupling_flux} are the complete phenomenological model of a system of coupled superconducting loop dendrites. The $J_{ij}$ are coupling parameters that can be chosen across a broad range in the circuit layout. To extend the model to account for neurons, certain dendrites are assigned to be somas, wherein a transmitter circuit is activated and light is produced when the value of $s$ reaches a specified threshold. The times of these threshold events are spike times and lead to synapse events on that neuron's downstream synapses. To extend the model to account for synapses, certain dendrites are assigned to receive input flux from an SPD as in Eq.\,\ref{eq:spd_response} at a time corresponding to the input neuron's spike times plus a delay to account for the activity of the transmitter circuits and light sources, as discussed in Appendices \ref{apx:source} and \ref{apx:transmitter}.

\section{\label{apx:soen_ode_comparison_additional_data}Additional Data Comparing Phenomenological and Circuit Models}
\begin{figure}[tbh]
\includegraphics[width=8.6cm]{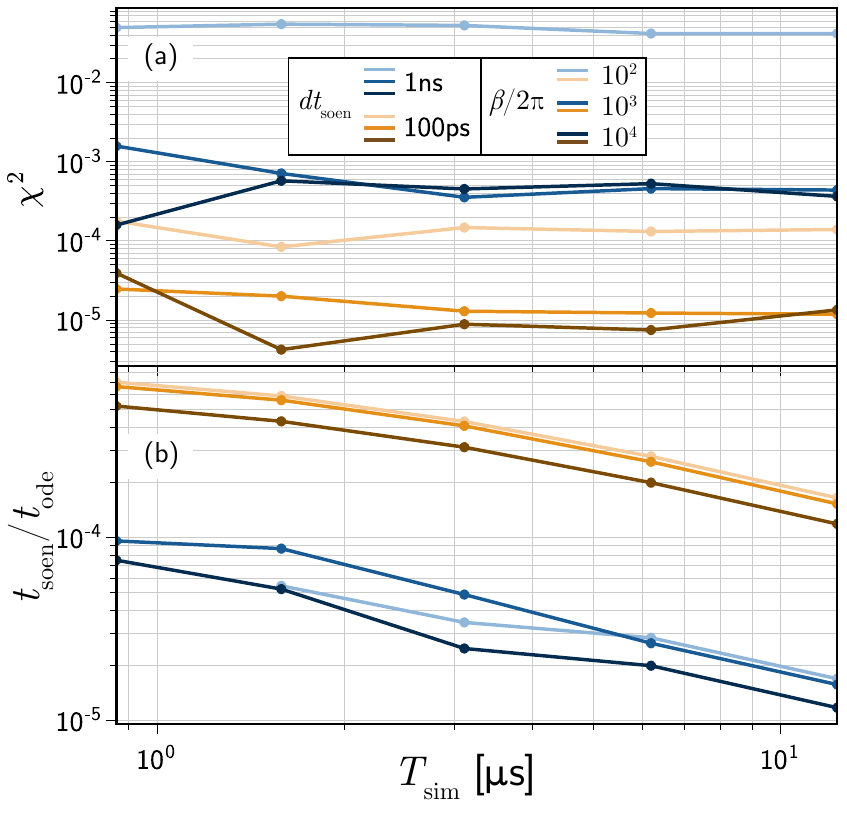}
\caption{\label{fig:apx__error_vs_dt__summary}Summary of errors in comparing the phenomenological model to the circuit equations as a function of the simulated time, $T_\mathrm{sim}$. In these simulations, $\tau_\mathrm{di} = 250$\,\textmu s, and the drive signals were random square-pulse inputs with 10, 20, 40, 80, and 160 pulses. (a) The values of $\chi^2$ for several values of time step and integration loop inductance. (b) The ratio of required time for the computation. $t_\mathrm{soen}$ is the time required to step through the phenomenological model, while $t_\mathrm{ode}$ is the time required to solve the system of circuit ODEs. The longest simulation treated $T_\mathrm{sim} = 12.32$\,\textmu s of simulated time.}
\end{figure}

Here we tabulate the $\chi^2$ and simulation time ratio data from Figs.\,\ref{fig:dend__ode_comparison__lin_ramp__100ps}, \ref{fig:dend__ode_comparison__sqr_pulse__100ps}, and \ref{fig:dend__ode_comparison__sqr_pulse__1ns}. Figure \ref{fig:apx__error_vs_dt__summary} shows further data from square pulse driving functions. The data points come from cases of 10, 20, 40, 80, and 160 square pulses. The $x$-axis is the duration of the simulated time.
\begin{table}[tbh]
\centering
\caption{Values obtained from the simulation of Fig.\,\ref{fig:dend__ode_comparison__lin_ramp__100ps}, the case of the linear ramp input with 100\,ps time step.}
\label{tab:comparison__lin_ramp__100ps}
\renewcommand{\arraystretch}{1.5}
\begin{tabular}{ cccc  }\toprule[1.5pt] 

\cellcolor{lightgreygreen}$\beta_\mathrm{di}/2\pi$ & \cellcolor{lightgreygreen}$\tau_\mathrm{di}$ & \cellcolor{lightgreygreen}$\chi^2$ & \cellcolor{lightgreygreen} $t_\mathrm{ode}/t_\mathrm{soen}$ \\ \midrule

\cellcolor{lightestgrey}$10^2$ & \cellcolor{lightestgrey} 10\,ns & \cellcolor{lightestgrey}$6.76\times 10^{-5}$ & \cellcolor{lightestgrey}$1.40\times 10^3$ \\
\cellcolor{lightestgrey}$10^2$ & \cellcolor{lightestgrey} 50\,ns & \cellcolor{lightestgrey}$8.01\times 10^{-5}$ & \cellcolor{lightestgrey}$1.29\times 10^3$ \\
\cellcolor{lightestgrey}$10^2$ & \cellcolor{lightestgrey} 250\,ns & \cellcolor{lightestgrey}$9.95\times 10^{-5}$ & \cellcolor{lightestgrey}$1.17\times 10^3$ \\
\cellcolor{lightestgrey}$10^2$ & \cellcolor{lightestgrey} 1.25\,\textmu s & \cellcolor{lightestgrey}$8.14\times 10^{-5}$ & \cellcolor{lightestgrey}$1.16\times 10^3$ \\

\cellcolor{lightgrey}$10^3$ & \cellcolor{lightgrey} 10\,ns & \cellcolor{lightgrey}$3.83\times 10^{-6}$ & \cellcolor{lightgrey}$1.78\times 10^3$ \\
\cellcolor{lightgrey}$10^3$ & \cellcolor{lightgrey} 50\,ns & \cellcolor{lightgrey}$5.07\times 10^{-6}$ & \cellcolor{lightgrey}$1.59\times 10^3$ \\
\cellcolor{lightgrey}$10^3$ & \cellcolor{lightgrey} 250\,ns & \cellcolor{lightgrey}$1.35\times 10^{-6}$ & \cellcolor{lightgrey}$1.71\times 10^3$ \\
\cellcolor{lightgrey}$10^3$ & \cellcolor{lightgrey} 1.25\,\textmu s & \cellcolor{lightgrey}$9.78\times 10^{-7}$ & \cellcolor{lightgrey}$1.37\times 10^3$ \\

\cellcolor{lightestgrey}$10^4$ & \cellcolor{lightestgrey} 10\,ns & \cellcolor{lightestgrey}$2.44\times 10^{-6}$ & \cellcolor{lightestgrey}$3.29\times 10^3$ \\
\cellcolor{lightestgrey}$10^4$ & \cellcolor{lightestgrey} 50\,ns & \cellcolor{lightestgrey}$1.37\times 10^{-6}$ & \cellcolor{lightestgrey}$2.90\times 10^3$ \\
\cellcolor{lightestgrey}$10^4$ & \cellcolor{lightestgrey} 250\,ns & \cellcolor{lightestgrey}$7.26\times 10^{-7}$ & \cellcolor{lightestgrey}$2.48\times 10^3$ \\
\cellcolor{lightestgrey}$10^4$ & \cellcolor{lightestgrey} 1.25\,\textmu s & \cellcolor{lightestgrey}$8.75\times 10^{-7}$ & \cellcolor{lightestgrey}$2.20\times 10^3$ \\

\bottomrule[1.5pt]
\end{tabular}
\end{table}

\begin{table}[tbh]
\centering
\caption{Values obtained from the simulation of Fig.\,\ref{fig:dend__ode_comparison__sqr_pulse__100ps}, the case of the square pulse inputs with 100\,ps time step.}
\label{tab:comparison__sqr_pulse__100ps}
\renewcommand{\arraystretch}{1.5}
\begin{tabular}{ cccc  }\toprule[1.5pt] 

\cellcolor{lightgreygreen}$\beta_\mathrm{di}/2\pi$ & \cellcolor{lightgreygreen}$\tau_\mathrm{di}$ & \cellcolor{lightgreygreen}$\chi^2$ & \cellcolor{lightgreygreen} $t_\mathrm{ode}/t_\mathrm{soen}$ \\ \midrule

\cellcolor{lightestgrey}$10^2$ & \cellcolor{lightestgrey} 10\,ns & \cellcolor{lightestgrey}$3.32\times 10^{-4}$ & \cellcolor{lightestgrey}$1.66\times 10^3$ \\
\cellcolor{lightestgrey}$10^2$ & \cellcolor{lightestgrey} 50\,ns & \cellcolor{lightestgrey}$1.86\times 10^{-4}$ & \cellcolor{lightestgrey}$1.48\times 10^3$ \\
\cellcolor{lightestgrey}$10^2$ & \cellcolor{lightestgrey} 250\,ns & \cellcolor{lightestgrey}$1.17\times 10^{-4}$ & \cellcolor{lightestgrey}$1.69\times 10^3$ \\
\cellcolor{lightestgrey}$10^2$ & \cellcolor{lightestgrey} 1.25\,\textmu s & \cellcolor{lightestgrey}$7.87\times 10^{-5}$ & \cellcolor{lightestgrey}$1.49\times 10^3$ \\

\cellcolor{lightgrey}$10^3$ & \cellcolor{lightgrey} 10\,ns & \cellcolor{lightgrey}$6.53\times 10^{-5}$ & \cellcolor{lightgrey}$2.31\times 10^3$ \\
\cellcolor{lightgrey}$10^3$ & \cellcolor{lightgrey} 50\,ns & \cellcolor{lightgrey}$2.20\times 10^{-5}$ & \cellcolor{lightgrey}$2.00\times 10^3$ \\
\cellcolor{lightgrey}$10^3$ & \cellcolor{lightgrey} 250\,ns & \cellcolor{lightgrey}$1.74\times 10^{-5}$ & \cellcolor{lightgrey}$1.53\times 10^3$ \\
\cellcolor{lightgrey}$10^3$ & \cellcolor{lightgrey} 1.25\,\textmu s & \cellcolor{lightgrey}$1.82\times 10^{-5}$ & \cellcolor{lightgrey}$1.62\times 10^3$ \\

\cellcolor{lightestgrey}$10^4$ & \cellcolor{lightestgrey} 10\,ns & \cellcolor{lightestgrey}$4.06\times 10^{-5}$ & \cellcolor{lightestgrey}$2.77\times 10^3$ \\
\cellcolor{lightestgrey}$10^4$ & \cellcolor{lightestgrey} 50\,ns & \cellcolor{lightestgrey}$1.12\times 10^{-5}$ & \cellcolor{lightestgrey}$2.47\times 10^3$ \\
\cellcolor{lightestgrey}$10^4$ & \cellcolor{lightestgrey} 250\,ns & \cellcolor{lightestgrey}$1.04\times 10^{-5}$ & \cellcolor{lightestgrey}$2.11\times 10^3$ \\
\cellcolor{lightestgrey}$10^4$ & \cellcolor{lightestgrey} 1.25\,\textmu s & \cellcolor{lightestgrey}$9.18\times 10^{-7}$ & \cellcolor{lightestgrey}$1.87\times 10^3$ \\

\bottomrule[1.5pt]
\end{tabular}
\end{table}

\begin{table}[tbh]
\centering
\caption{Values obtained from the simulation of Fig.\,\ref{fig:dend__ode_comparison__sqr_pulse__1ns}, the case of square pulse inputs with 1\,ns time step.}
\label{tab:comparison__sqr_pulse__1ns}
\renewcommand{\arraystretch}{1.5}
\begin{tabular}{ cccc  }\toprule[1.5pt] 

\cellcolor{lightgreygreen}$\beta_\mathrm{di}/2\pi$ & \cellcolor{lightgreygreen}$\tau_\mathrm{di}$ & \cellcolor{lightgreygreen}$\chi^2$ & \cellcolor{lightgreygreen} $t_\mathrm{ode}/t_\mathrm{soen}$ \\ \midrule

\cellcolor{lightestgrey}$10^2$ & \cellcolor{lightestgrey} 10\,ns & \cellcolor{lightestgrey}$3.60\times 10^{-2}$ & \cellcolor{lightestgrey}$5.09\times 10^3$ \\
\cellcolor{lightestgrey}$10^2$ & \cellcolor{lightestgrey} 50\,ns & \cellcolor{lightestgrey}$2.34\times 10^{-2}$ & \cellcolor{lightestgrey}$5.43\times 10^4$ \\
\cellcolor{lightestgrey}$10^2$ & \cellcolor{lightestgrey} 250\,ns & \cellcolor{lightestgrey}$2.70\times 10^{-2}$ & \cellcolor{lightestgrey}$1.41\times 10^4$ \\
\cellcolor{lightestgrey}$10^2$ & \cellcolor{lightestgrey} 1.25\,\textmu s & \cellcolor{lightestgrey}$2.45\times 10^{-2}$ & \cellcolor{lightestgrey}$1.48\times 10^4$ \\

\cellcolor{lightgrey}$10^3$ & \cellcolor{lightgrey} 10\,ns & \cellcolor{lightgrey}$2.84\times 10^{-3}$ & \cellcolor{lightgrey}$1.75\times 10^4$ \\
\cellcolor{lightgrey}$10^3$ & \cellcolor{lightgrey} 50\,ns & \cellcolor{lightgrey}$9.25\times 10^{-4}$ & \cellcolor{lightgrey}$1.41\times 10^4$ \\
\cellcolor{lightgrey}$10^3$ & \cellcolor{lightgrey} 250\,ns & \cellcolor{lightgrey}$2.50\times 10^{-4}$ & \cellcolor{lightgrey}$>4\times 10^4$ \\
\cellcolor{lightgrey}$10^3$ & \cellcolor{lightgrey} 1.25\,\textmu s & \cellcolor{lightgrey}$1.20\times 10^{-4}$ & \cellcolor{lightgrey}$1.37\times 10^4$ \\

\cellcolor{lightestgrey}$10^4$ & \cellcolor{lightestgrey} 10\,ns & \cellcolor{lightestgrey}$2.32\times 10^{-3}$ & \cellcolor{lightestgrey}$2.54\times 10^4$ \\
\cellcolor{lightestgrey}$10^4$ & \cellcolor{lightestgrey} 50\,ns & \cellcolor{lightestgrey}$6.62\times 10^{-4}$ & \cellcolor{lightestgrey}$2.82\times 10^4$ \\
\cellcolor{lightestgrey}$10^4$ & \cellcolor{lightestgrey} 250\,ns & \cellcolor{lightestgrey}$3.07\times 10^{-4}$ & \cellcolor{lightestgrey}$2.14\times 10^4$ \\
\cellcolor{lightestgrey}$10^4$ & \cellcolor{lightestgrey} 1.25\,\textmu s & \cellcolor{lightestgrey}$1.72\times 10^{-4}$ & \cellcolor{lightestgrey}$1.92\times 10^4$ \\

\bottomrule[1.5pt]
\end{tabular}
\end{table}

\section{\label{apx:source}Rate Equations for the Transmitter Light Source}
To accurately model a superconducting optoelectronic loop neuron, the formalism of the computational circuitry\textemdash the dendrites\textemdash must be accompanied by a treatment of the transmitter circuit and light source that produce photons when a soma reaches threshold. We begin by treating the light emitters themselves in a rate-equation framework and subsequently consider the transmitter circuit that will provide current to the emission medium. We consider two types of light emitters: quantum dots and silicon emissive centers. The quantum dots we have in mind are group III-V self-assembled dots grown by molecular beam epitaxy or metallorganic chemical vapor deposition. Such emitters have the advantages of high efficiency and fast radiative lifetime. However, for the application at hand scalable manufacturing and integration with semiconductor and superconductor electronics are crucial considerations, and III-V light emitters have a long history of difficulty integrating with silicon. The quantum dot model is also applicable to group IV, Ge-based quantum dots \cite{zinovyev2021si}, which may prove adequate for the present application. We additionally consider a rate-equation model of silicon emissive centers, as they are easy to fabricate and integrate with electronics, although to date their efficiency is low.

\subsection{Quantum Dots}
The areal density of quantum dots is $n_\mathrm{D}$ (600/\textmu m$^2$). We define the concentration (number per area which can be converted to volume) in the uncharged, unpopulated, ground state as $n_0$. The concentration that have trapped a hole is $n_1^\mathrm{p}$, while the concentration that have a trapped electron is $n_1^\mathrm{n}$. The concentration that have trapped a hole and an electron is $n_2$. Assuming quantum dots are not created or destroyed during operation we have
\begin{equation}
n_\mathrm{D} = n_0 + n_1^\mathrm{p} + n_1^\mathrm{n} + n_2.
\end{equation}
The population of quantum dots in these states is modeled by the following rate equations:
\begin{equation}
\label{eq:rate_qd}
\begin{split}
\frac{d\,n_0}{dt} &= -n_0\,p\,c_{01}^\mathrm{p} - n_0\,n\,c_{01}^\mathrm{n} + n_2\,e_{20}, \\[10pt]
\frac{d\,n_1^\mathrm{n}}{dt} &= n_0\,n\,c_{01}^\mathrm{n} - n_1^\mathrm{n}\,p\,c_{12}^\mathrm{p}, \\[10pt]
\frac{d\,n_2}{dt} &= n_1^\mathrm{n}\,p\,c_{12}^\mathrm{p} + n_1^\mathrm{p}\,n\,c_{12}^\mathrm{n} - n_2\,e_{20}.
\end{split}
\end{equation}
In Eqs.\,\ref{eq:rate_qd}, $c_{01}^\mathrm{p}$ is the capture coefficient for an unpopulated dot to obtain a hole, and $c_{01}^\mathrm{n}$ is the capture coefficient for an unpopulated dot to obtain an electron. The quantities $c_{12}^\mathrm{p}$ and $c_{12}^\mathrm{n}$ are defined analogously, and $e_{20}$ is the emission probability, related to the radiative lifetime by $\tau_\mathrm{rad} = 1/e_{20}$. The populations of free electrons and holes in the semiconductor medium are given by
\begin{equation}
\label{eq:rate_np__qd}
\begin{split}
\frac{d n}{dt} &= \frac{\eta_\mathrm{inj}\,I}{q\,V_{\mathrm{LED}}} - n_0\,n\,c_{01}^\mathrm{n} - n_1^\mathrm{p}\,n\,c_{12}^\mathrm{n} \\
&-\frac{v_\mathrm{s}\,A_{\mathrm{LED}}}{V_{\mathrm{LED}}}\, n - (C_n\, n^2\,p + C_p\, p^2\,n) , \\[10pt]
\frac{d p}{dt} &= \frac{\eta_\mathrm{inj}\,I}{q\,V_{\mathrm{LED}}} - n_0\,p\,c_{01}^\mathrm{p} - n_1^\mathrm{n}\,p\,c_{12}^\mathrm{p} \\
&-\frac{v_\mathrm{s}\,A_{\mathrm{LED}}}{V_{\mathrm{LED}}} p - (C_n\, n^2\,p + C_p\, p^2\,n).
\end{split}
\end{equation}
In Eqs.\,\ref{eq:rate_np__qd} $\eta_\mathrm{inj}$ is the injection efficiency, which could be determined by a spatial carrier transport model, $v_\mathrm{s}$ is the surface recombination velocity, $A_{\mathrm{LED}}$ is the surface area of the LED, $V_{\mathrm{LED}}$ is its volume, and $C_n$ and $C_p$ are the Auger recombination coefficients. Throughout Eqs.\,\ref{eq:rate_qd}, $n_1^\mathrm{p}$ can be eliminated with $n_1^\mathrm{p} = n_\mathrm{qd} - n_0 - n_1^\mathrm{n} - n_2$.

The rate of photon production from electroluminescence under current injection is given by 
\begin{equation}
\frac{dN_\mathrm{ph}}{dt} = V_\mathrm{LED} \, e_{20} \, n_2(t).
\end{equation}
The total number of photons emitted is obtained by integrating:
\begin{equation}
N_\mathrm{ph} = V_\mathrm{LED} \, e_{20} \, \int_{t_0}^{t_f}n_2(t) dt.
\end{equation}
The primary quantity of interest for our application is the efficiency of light production. In the context of the present model, we can define the efficiency as the number of photons generated divided by the number of electron-hole pairs injected into the structure:
\begin{equation}
\label{eq:efficiency}
\eta_i = \frac{N_\mathrm{ph}}{N_{eh}},
\end{equation}
where $N_{eh}$ is the total number of electron-hole pairs injected into the intrinsic region of the diode.

To simulate the QDs we use a radiative lifetime of 1\,ns, giving $e_{20} = 1$\,GHz. The capture coefficients are expected to be quite fast for quantum dots, and we take $c_{01} = c_{12} = 10^{-10}$m$^3$/s. The precise values of $c_{01}$ and $c_{12}$ have only minor impact on the numerical results as the capture rates are always much smaller than the recombination times. For the parameters governing non-radiative recombination we $v_\mathrm{s} = 2.5\times 10^3$cm/s \cite{higuera2017ultralow} and $C_n = C_p = 7\times 10^{-30}$cm$^6$/s \cite{schroder2015semiconductor}.

The system of ODEs given by Eqs.\,\ref{eq:rate_qd} and \ref{eq:rate_np__qd} was again solved using \code{solve\_ivp}. Complete analysis of these light sources is beyond the scope of the present work. The important take-away is that the efficiency of the source can be quite high, provided the number of carriers injected is commensurate with the number of dots in the diode. For this spiking neuromorphic application, we are interested in cases wherein an initial pulse of electrons and holes is injected into the diode and left to decay. If the pulse of injected carriers contains appreciably more electrons and holes than the number of quantum dots, the excess carriers recombine non-radiatively through surfaces or Auger before the populated dots have time to emit photons and trap additional carriers. Thus, for a neuron to efficiently produce light upon reaching threshold, the light source must have a number of quantum dots that is chosen in accordance with the number of synapses that are intended to receive photons, and the transmitter circuit that injects the light source must be designed to inject the appropriate number of electrons and holes in a brief burst of charge. Such a transmitter circuit is summarized in Appendix \ref{apx:transmitter}. Time-domain simulations of quantum dots being driven by such a driver circuit will be presented in that Appendix. First, we briefly introduce the rate equation model of silicon emissive centers. 

\subsection{Silicon emissive centers}
We can expand the quantum dot model to treat silicon emissive centers. A similar model was presented in Ref.\,\onlinecite{recht2009temperature}, but in that work steady state emission was investigated to seek a continuous-wave laser. He we require the full dynamical equations to investigate transient behavior. The total concentration of the particular emissive center of interest is $n_\mathrm{D}$. We define the concentration (number per volume) in the uncharged, unpopulated, ground state as $n_0$. The concentration that have trapped a hole is $n_1$. The concentration that have trapped a hole and an electron is $n_2$. Following Ref.\,\onlinecite{recht2009temperature} we focus on W centers \cite{buckley2017all} and assume a hole is always trapped before an electron. Assuming emissive centers are not created or destroyed during operation we have
\begin{equation}
n_\mathrm{D} = n_0 + n_1 + n_2.
\end{equation}
The equations governing the population of emissive centers are
\begin{equation}
\label{eq:rate_ec}
\begin{split}
\frac{d n_0}{dt} &= -n_0\,p\,c_{01} + (n_\mathrm{D}-n_0-n_2) c_{01} k_{01} + n_2\,e_{20} \\
\frac{d n_2}{dt} &= -n_2\,(e_{20} + c_{12}\,k_{12}) + (n_\mathrm{D}-n_0-n_2)\,n\,c_{12}.
\end{split}
\end{equation}
In Eqs.\,\ref{eq:rate_ec}, $c_{01}$ is the capture coefficient for an unpopulated center to obtain a hole. As was argued in Ref.\,\onlinecite{recht2009temperature}, W centers appear to always trap a hole before an electron. The capture coefficient for trapping an electron after the center has been populated with a hole is $c_{12}$. $e_{20}$ is the emission coefficient. The emission coefficients are related to the capture coefficients through the relations
\begin{equation}
\begin{split}
e_{10} &= c_{01}\, k_{01} = c_{01} \frac{n_0^* n_c^*}{n_1^*} \\
e_{21} &= c_{12}\, k_{12} = c_{12} \frac{n_1^2 n_c^*}{n_2^*}.
\end{split}
\end{equation}
Asterisks refer to equilibrium values. The relations can be further specified by
\begin{equation}
\begin{split}
k_{01} &= \frac{n_0^* n_c^*}{n_1^*} = \frac{N_v}{g_n} \mathrm{e}^{-E_h/kT} \\
k_{12} &= \frac{n_1^* n_c^*}{n_2^*} = \frac{N_c}{g_e} \mathrm{e}^{(E_e-E_g)/kT}.
\end{split}
\end{equation}
Here $N_v$ and $N_c$ are the valence and conduction band densities of states, $g_n$ and $g_e$ are the degeneracies of the states of the emissive center, $E_g$ is the band gap, and $E_h$ and $E_g - E_e$ are the hole and electron binding energies. The energy of the emitted photon is $E_e-E_h$. For the case of the W center, this equals 1.018\,eV. The top of the valence band is set to zero energy. See Ref.\,\onlinecite{pierret1987advanced}, Ch. 5 for further explanation.

In the second of Eqs.\,\ref{eq:rate_ec} the term $e_{20} + c_{12}\,k_{12}$ plays a prominent role in determining the efficiency of emission. This term represents competition between emission ($e_{20}$) and release of the electron in the bound exciton back to the conduction band ($c_{12}\,k_{12}$). The ratio $e_{20}/(c_{12}\,k_{12})$ is roughly $10^5$ at 4.2\,K, so emission is much more likely than exciton dissociation at this temperature.

In addition to the rate equations for the emissive center populations, the model for Si light sources should include the effects of a population of non-radiative centers that will inevitably be produced when the implants for the emissive centers are performed. Currently knowledge of the number of different types of non-radiative centers, their capture coefficients, and their lifetimes is limited. For simplicity, we assume there is one type of dominant non-radiative recombination center, we assume it can be populated first by an electron or hole, we assume the capture cross section is $c_{01}$, and we assume the secondary capture of a hole or electron is governed by $c_{12}$. We also assume after capture of an electron, hole, or both, no carriers are ever re-emitted to the bands, which is reasonable at the low temperatures necessary for the superconducting electronics present in the system. The total concentration of non-radiative defects is $n_\mathrm{D}^\mathrm{nr} = n_0^\mathrm{nr}+n_1^\mathrm{nr}+n_2^\mathrm{nr}$. The population rate equations for non-radiative centers then can be written as:
\begin{equation}
\label{eq:rate_nr}
\begin{split}
\frac{d n_0^\mathrm{nr}}{dt} &= -n_0^\mathrm{nr}\,c_{01}(p+n) + n_2^\mathrm{nr}\,e_{20}^\mathrm{nr} \\
\frac{d n_2^\mathrm{nr}}{dt} &= -n_2^\mathrm{nr}\,e_{20}^\mathrm{nr} + c_{12}\,(n_\mathrm{D}^\mathrm{nr}-n_0^\mathrm{nr}-n_2^\mathrm{nr})\,(p+n).
\end{split}
\end{equation}
The coefficients $c_{ij}$ are different here than in the emissive center equations, but we expect them to be fast, so this is of little consequence. The lifetime of the non-radiative transition of the center (which may be a trap-assisted Auger process or, more likely, phonon recombination) is given by $\tau_\mathrm{nr} = 1/e_{20}^\mathrm{nr}$.

Equations \ref{eq:rate_ec} and \ref{eq:rate_nr} are coupled to the charge-carrier rate equations. For the electron concentration we have
\begin{equation}
\label{eq:rate_n}
\begin{split}
\frac{d n}{dt} &= \frac{\eta_\mathrm{inj}\,I}{q\,V_{\mathrm{LED}}} - (n_\mathrm{D}-n_0-n_2)\, n\, c_{12} + n_2\, c_{12}\, k_{12} \\
&-n_0^\mathrm{nr}\,n\,c_{01} - (n_\mathrm{D}^\mathrm{nr}-n_0^\mathrm{nr}-n_2^\mathrm{nr})\,n\,c_{12}\\
&-\frac{v_\mathrm{s}\,A_{\mathrm{LED}}}{V_{\mathrm{LED}}}\, n - (C_n\, n^2\,p + C_p\, p^2\,n) , 
\end{split}
\end{equation}
and for the hole concentration we have
\begin{equation}
\label{eq:rate_p}
\begin{split}
\frac{d p}{dt} &= \frac{\eta_\mathrm{inj}\,I}{q\,V_{\mathrm{LED}}} - n_0\, p\, c_{01} \\
&-n_0^\mathrm{nr} \, p \, c_{01}  - (n_\mathrm{D}^\mathrm{nr}-n_0^\mathrm{nr}-n_2^\mathrm{nr})\,p\,c_{12} \\
&-\frac{v_\mathrm{s}\,A_{\mathrm{LED}}}{V_{\mathrm{LED}}} p - (C_n\, n^2\,p + C_p\, p^2\,n).
\end{split}
\end{equation} 

To numerically solve Eqs.\,\ref{eq:rate_ec}, \ref{eq:rate_nr}, \ref{eq:rate_n}, and \ref{eq:rate_p}, we need to specify $c_{01}$, $c_{12}$, and $e_{20}$. For the process of photon emission, related to $e_{20}$, we take $e_{20} = \tau_{ec}^{-1} = 1/40$\,ns\,$= 2.5\times 10^{7}$ \cite{buckley2017all,chartrand2019hunt}. To our knowledge, $c_{01}$ and $c_{12}$ have not been measured. We take $c_{01} = 5\times 10^{-14}$m$^3$/s and $c_{12} = 1.8\times 10^{-11}$m$^3$/s. These numbers are motivated by a simple geometrical model as well as a hydrogenic model of an impurity screened by the dielectric \cite{lannoo2012point}. Numerical studies show that the results are not sensitive to these values across several orders of magnitude.

To specify the nonradiative terms in the rate equations, we take the Auger recombination coefficients as $C_n = 2.8\times 10^{-31}$\,cm$^6$/s and $C_p = 1\times 10^{-31}$\,cm$^6$/s. These are the values at 300\,K \cite{schroder2015semiconductor}, and they will reduce slightly at low temperature (scaling with temperature as $T^{0.6}$ \cite{svantesson1979temperature}), but this is not appreciable given the low level of accuracy in the present model, so we ignore the temperature dependence of Auger and use 300\,K values. The surface recombination can be made quite low in properly treated Si and Si-SiO$_2$ interfaces, with values as low as 0.25\,cm/s in the literature \cite{yablonovitch1986unusually}. Throughout these simulations we use $v_\mathrm{s} = 2.5$cm/s.

Time-domain simulations of silicon emissive centers based on Eqs.\,\ref{eq:rate_ec}, \ref{eq:rate_nr}, \ref{eq:rate_n}, and \ref{eq:rate_p} provides a similar message to that learned for quantum dots: the emitters will populate rapidly, but if the level of carrier injection is too large, excess carriers will go to waste. In the case of silicon emissive centers, loss is not due to surfaces or Auger, but rather to nonradiative centers. The relative concentration of nonradiative centers formed during the ion implantation used to realize W centers is likely to be a crucial quantity determining the ultimate efficiency of such a light source. Given the competing populations of emissive and nonradiative centers, it is not sufficient to inject the same number of carriers as emissive centers and desired photons; one must inject enough carriers to populate both the nonradiative centers and the emissive centers. However, if the concentration of nonradiative centers can be limited to something like 10 times the concentration of emitters, total light-production efficiency may stay within a useful range.

We next describe transmitter circuits capable of providing brief pulses of injected current to produce the desired number of photons necessary to serve a neuron's downstream synaptic connections.

\section{\label{apx:transmitter}Circuit Equations for the Transmitter Driver Circuit}
\begin{figure*}[tbh]
\includegraphics[width=17.2cm]{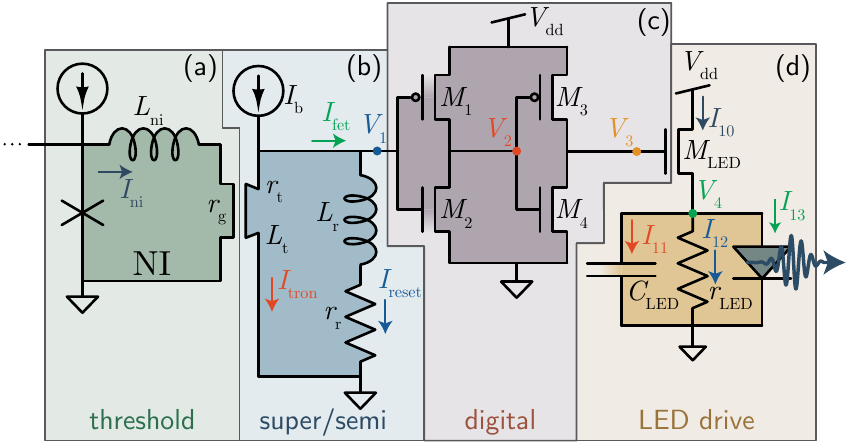}
\caption{\label{fig:apx__circuit_transmitter}Diagram of the transmitter circuit with relevant components labeled. (a) The neuronal integration loop ($\mathsf{NI}$) and tron thresholding element. (b) The superconductor-semiconductor interface where the tron switches the first inverter. (c) The inverter pair that produces the digital drive signal. (d) The LED driver MOSFET and the LED itself.}
\end{figure*}
The transmitter circuit under consideration is shown in Fig.\,\ref{fig:apx__circuit_transmitter}. The circuit combines JJs, an hTron thresholding element \cite{mccaughan2019superconducting}, MOSFETs (PMOS and NMOS), and an LED. The hTron serves as the interface between the superconducting and semiconducting domains, enabling low-voltage JJs to switch a CMOS inverter. The basic operation of the circuit is as follows. Signal moves from left to right. When the JJ in the neuronal integration (NI) loop has added sufficient current to drive the hTron gate above threshold, the hTron channel switches from zero resistance to high resistance. The current bias to the hTron channel ($I_\mathrm{b}$) is thus shunted from the hTron channel ($I_\mathrm{tron}$) to the passive reset branch ($L_\mathrm{r}$, $r_\mathrm{r}$), temporarily inducing a voltage at node $V_1$ sufficient to switch the first CMOS inverter ($M_1$, $M_2$). The two inverters serve to produce a digital signal with voltage $V_\mathrm{dd}$ applied to MOSFET $M_\mathrm{LED}$ when the neuronal threshold is reached and the hTron switches. This digital behavior ensures the superconducting components are well separated from the LED, providing consistent current biasing of the LED decoupled from the performance of the thresholding element. The LED driver MOSFET ($M_\mathrm{LED}$) can then be designed with a width-to-length ratio and doping level commensurate with the number of photons desired from the LED, which is chosen based on the number of synaptic connections made by the neuron. $M_\mathrm{LED}$ will deliver a pulse of current to the LED while the hTron is keeping the voltage $V_1$ high, and when that voltage drops as the passive reset circuit performs its operation, the current to the LED will cease. As is shown below, the MOSFETs are active for a few nanoseconds. With this qualitative description of the circuit operation in mind, we now present the circuit model used in this work to provide a phenomenological treatment of the behavior.

To treat the thresholding component where the current added to the integration loop drives an hTron above threshold, we model the hTron as a device that switches from a zero resistance to a high-resistance state when a certain current is reached. Accurately modeling the electrothermal dynamics of the hTron is difficult and will do little to improve the model for the present purpose, so a simple resistive switch is employed.

To simulate the behavior of the MOSFETs, we use a charge-control model \cite{ytterdal2003device} with current-voltage characteristics given by
\begin{equation}
\label{eq:mosfet__sccm_model__IV}
\begin{split}
I_{\mathrm{ds}}(V_\mathrm{ds}, V_\mathrm{gs}) &= \frac{W \mu_n c_\mathrm{i}}{L} \\
&\times
\begin{cases} (V_\mathrm{gt}-V_\mathrm{ds}/2)V_\mathrm{ds}, & \text{for } V_\mathrm{ds} \le V_\mathrm{sat} \\
V_\mathrm{gt}^2/2, & \text{for } V_\mathrm{ds} > V_\mathrm{sat}.
\end{cases}
\end{split}
\end{equation}
$V_\mathrm{gt}$ is the voltage above threshold: $V_\mathrm{gt} = V_\mathrm{gs} - V_\mathrm{t}$. We assume the drain-source current is zero for $V_\mathrm{gs} < V_\mathrm{t}$, i.e., we ignore subthreshold behavior. In this model, $V_\mathrm{sat} = V_\mathrm{gt}$.

\begin{figure*}[htb]
\includegraphics[width=17.2cm]{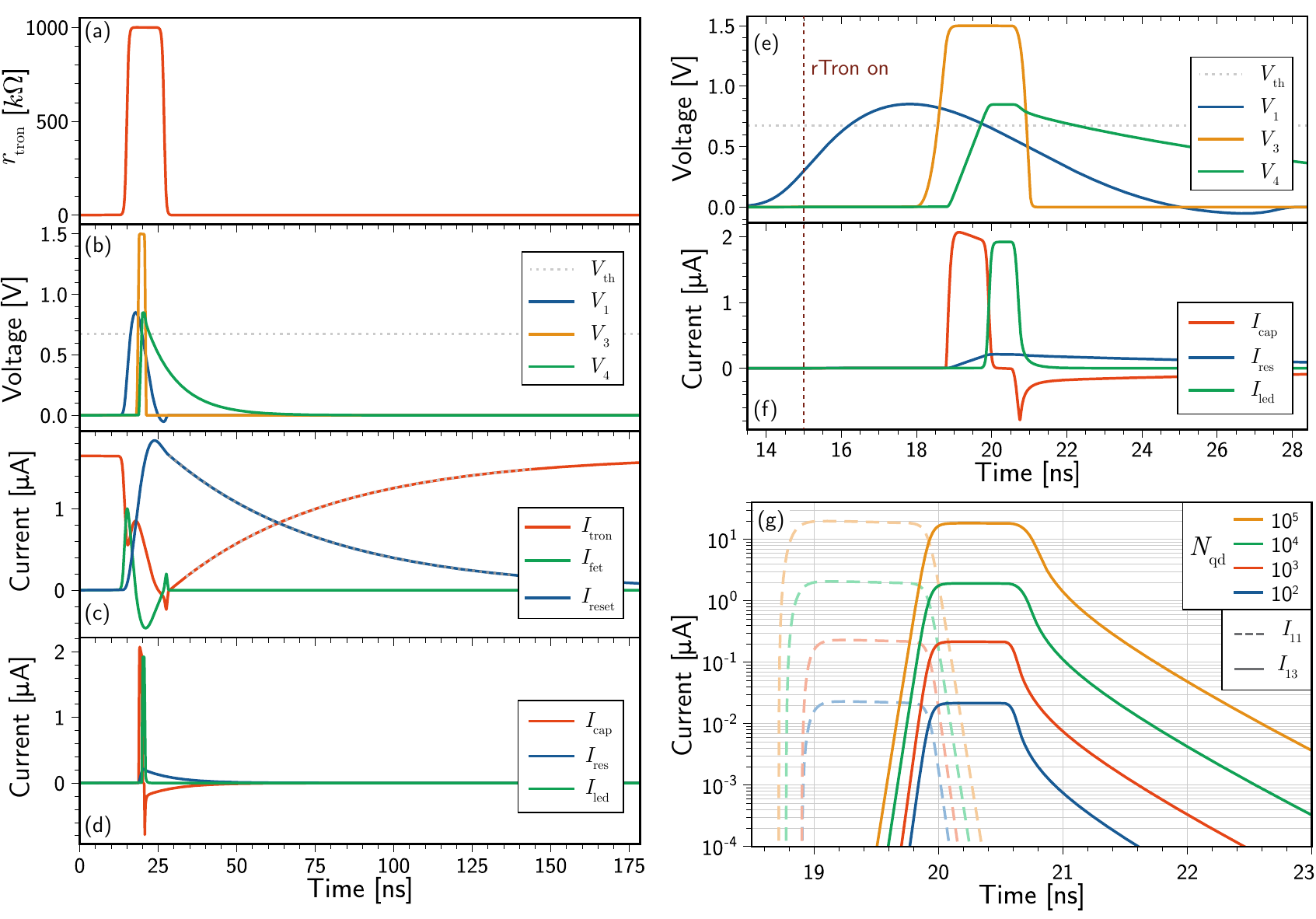}
\caption{\label{fig:apx__transmitter_time_traces}Time traces from the transmitter circuit. (a) The resistance of the tron as a function of time. (b) The voltages of the MOSFETs. (c) The currents in the tron and related circuit elements. (d) Currents through the LED. (e) Temporal zoom of the MOSFET voltages. (f) Temporal zoom of the LED currents. (g) Currents into the LED and capacitor when different numbers of quantum dots are present in the light source.}
\end{figure*}
We treat the LED with a conventional diode model of the form
\begin{equation}
\label{eq:led__IV}
\begin{split}
I_\mathrm{LED} = & \,eA\left[ (D_p/L_p)p_n + (D_n/L_n)n_p \right] \\
& \times\left[ \mathrm{exp}\left(eV/k_\mathrm{B}T\right) - 1 \right].
\end{split}
\end{equation}
In the present work we have considered two classes of LEDs. One is based on silicon emissive centers \cite{buckley2017all}, as have been considered in the context of SOENs due to their unique process compatability and feasibility for enabling low-cost manufacturing of large systems. The other is based on III-V quantum dots, specifically with material parameters for InGaAs quantum dots on a GaAs platform. For integration with SOENs, wafer bonding of such substrates may be employed. The model presented here can take into account either of these light sources. Both will obey an equation of the form given by Eq.\,\ref{eq:led__IV}, with the diode placed in parallel with a capacitor and a resistor as shown in Fig.\,\ref{fig:apx__circuit_transmitter}. In reality the capacitance and parasitic shunt resistance will both be functions of applied voltage, but for simplicity we fix these values. Based on the source rate equation model in Appendix \ref{apx:source} we know that we would like to have one emitter (quantum dot or emissive center) per photon to be generated by the light source per neuronal firing pulse. We assume the emitters will have a specified density $\rho_\mathrm{ec}$, so the value of the capacitance will depend on the number of emitters through the geometry of the junction. The number of emitters is in turn specified by the number of synaptic connections made by the neuron. Therefore, we consider two values of capacitance per unit area ($C_\mathrm{a} = 1\times 10^{-7}$F/\textmu m$^2$ and $C_\mathrm{a} = 1\times 10^{-5}$F/\textmu m$^2$) to cover a range of values that may be found in various devices. The value of the capacitor, $C_\mathrm{LED}$, is determined by $C_\mathrm{LED} = C_\mathrm{a}\,N_\mathrm{ph}/\rho_\mathrm{ec}$ with $\rho_\mathrm{ec}$ specified as a number of emitters per unit area.

\begin{figure*}[htb]
\centering
\includegraphics[width=17.2cm]{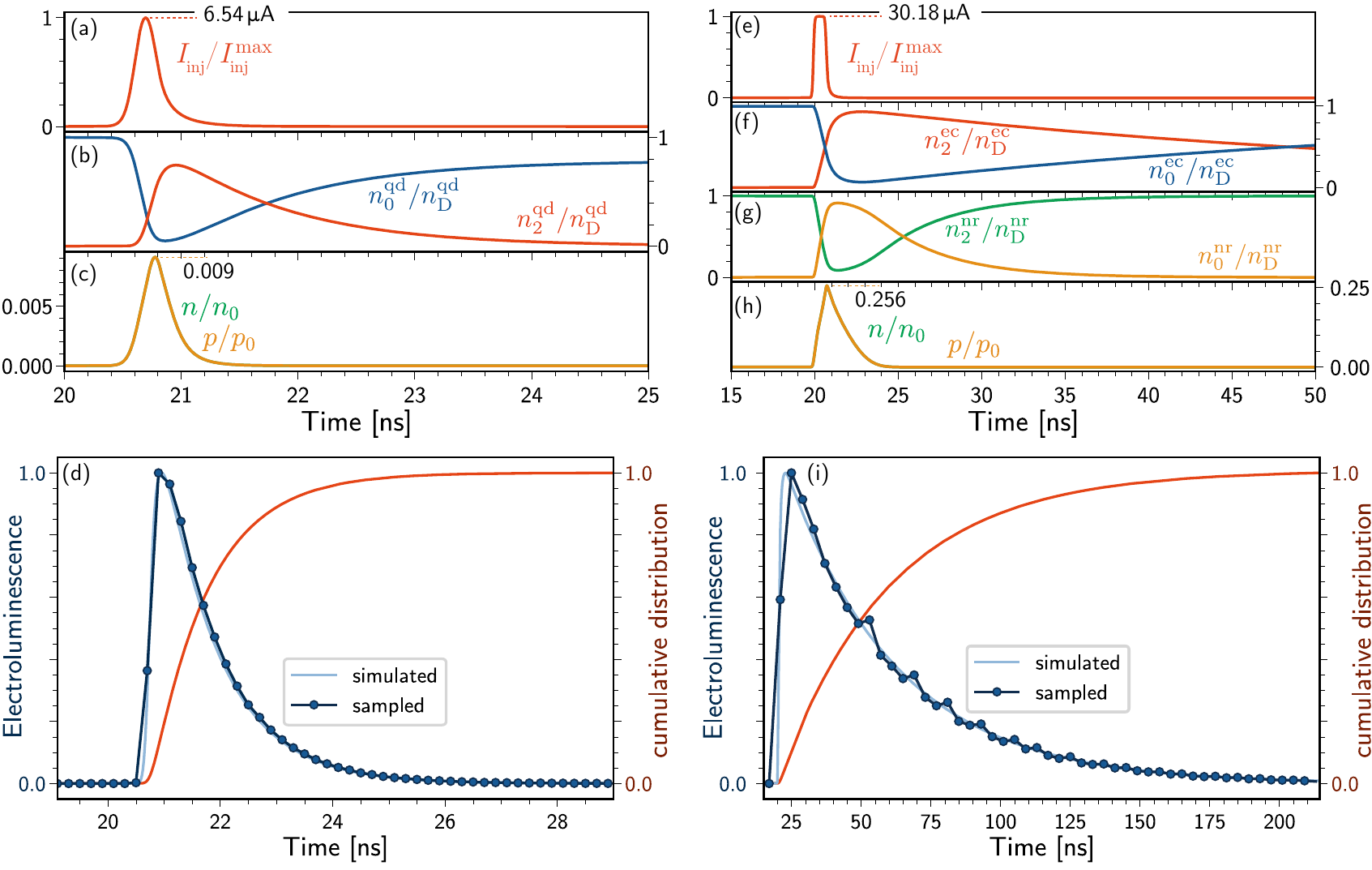}
\caption{Time traces for the light source model. (a)-(d) The quantum-dot model. (e)-(i) The emissive center model. (a) The current pulse injected from the transmitter circuit. (b) The populations of quantum dots in the ground and excited states normalized to the concentration of quantum dots in the LED volume. (c) The concentrations of free electrons and holes normalized to the total number injected into the LED volume. (d) The simulated electroluminescence signal, the cumulative distribution, and a histogram of sampled emission times. (e) The current pulse injected from the transmitter circuit. (f) The populations of emissive centers in the ground and excited states normalized to the concentration of emitters in the LED. (g) The populations of non-radiative recombination centers in the ground and excited states normalized to the concentration of non-radiative recombination centers in the LED. (h) The concentrations of free electrons and holes. (i) The simulated electroluminescence signal, the cumulative distribution, and a histogram of sampled emission times.}
\label{fig:apx__sources}
\end{figure*}
Given these models for the hTron, MOSFETs, and LED the following circuit equations can be derived:
\begin{equation}
\label{eq:transmitter}
\begin{split}
\frac{d\,V_1}{dt} &= \frac{I_b - I_1 - I_3}{C_\mathrm{I}}, \\
\frac{d\,V_2}{dt} &= \frac{-1}{C_\mathrm{I}} \left[ I_\mathrm{p}(V_2-V_\mathrm{dd},V_1-V_\mathrm{dd}) + I_\mathrm{n}(V_2,V_1) \right], \\
\frac{d\,V_3}{dt} &= \frac{-1}{C_\mathrm{D}} \left[ I_\mathrm{p}(V_3-V_\mathrm{dd},V_2-V_\mathrm{dd}) + I_\mathrm{n}(V_3,V_2) \right], \\
\frac{d\,V_4}{dt} &= \frac{1}{C_\mathrm{LED}} \left[ I_\mathrm{n}(V_\mathrm{dd}-V_4,V_3) - I_\mathrm{LED}(V_4) - \frac{V_4}{r_\mathrm{LED}} \right], \\
\frac{d^2\,I_1}{dt^2} &= \frac{I_b-I_1-I_3}{L_\mathrm{t} C_\mathrm{I}} - \frac{\dot{r}_\mathrm{t}}{L_\mathrm{t}}\,I_1 - \frac{r_\mathrm{t} I_1}{L_\mathrm{t}}\,\frac{d\,I_1}{dt}, \\
\frac{d\,I_3}{dt} &= \frac{r_\mathrm{t}}{L_\mathrm{r}}\,I_1 + \frac{L_\mathrm{t}}{L_\mathrm{r}}\,\frac{d\,I_1}{dt} - \frac{r_\mathrm{r}}{L_\mathrm{r}}\,I_3.
\end{split}
\end{equation}
Here, $I_\mathrm{p}(V_\mathrm{ds},V_\mathrm{gs})$ and $I_\mathrm{n}(V_\mathrm{ds},V_\mathrm{gs})$ are the PMOS and NMOS forms of Eq.\,\ref{eq:mosfet__sccm_model__IV} (i.e., current flows from source to drain in PMOS and drain to source in NMOS), $C_\mathrm{I}$ is the capacitance of an inverter, and $C_\mathrm{D}$ is the capacitance of the MOSFET driver to the LED. The time-dependent resistance of the hTron is $r_\mathrm{t}$ and its temporal derivative is $\dot{r}_\mathrm{t}$.

\begin{figure*}[htb]
\includegraphics[width=17.2cm]{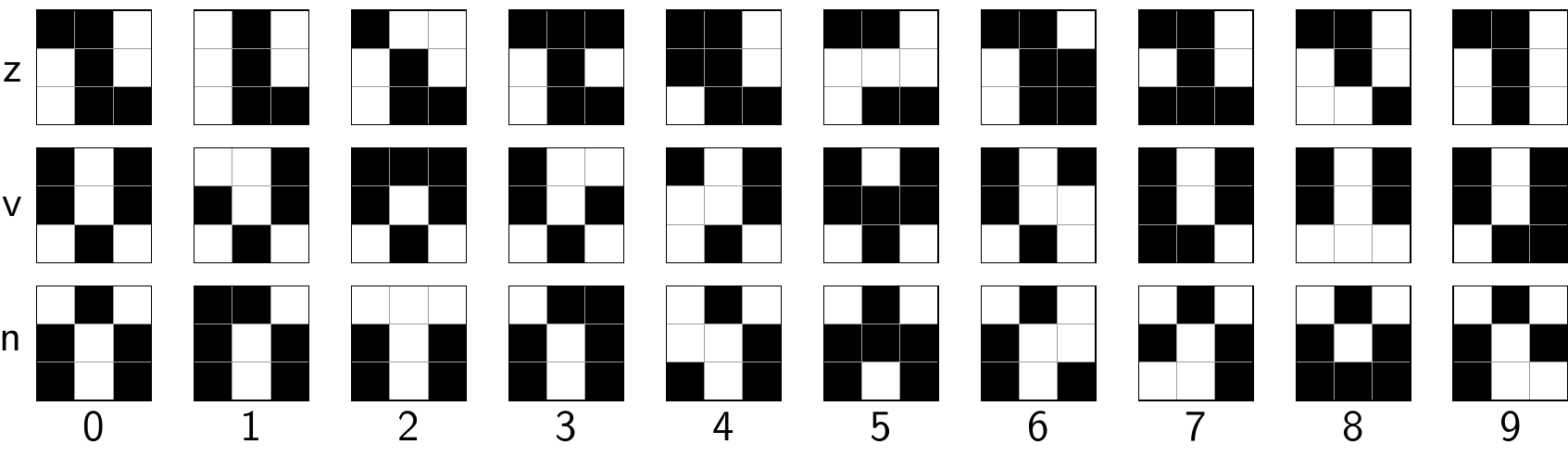}
\caption{\label{fig:apx__nine_pixel_drive_matrices}All 30 inputs to the nine-pixel classifier.}
\end{figure*}
Time domain circuit simulations of the transmitter model given by Eqs.\,\ref{eq:transmitter} have been carried out using \code{solve\_ivp}. A sample of results is given in Fig.\,\ref{fig:apx__transmitter_time_traces}. The resistance pulse of the hTron is shown in Fig.\,\ref{fig:apx__transmitter_time_traces}(a), and the relevant voltages are shown in Fig.\,\ref{fig:apx__transmitter_time_traces}(b). The currents are shown in Fig.\,\ref{fig:apx__transmitter_time_traces}(c) and (d). A temporal zoom of the voltage traces is shown in Fig.\,\ref{fig:apx__transmitter_time_traces}(e), where the MOSFET threshold is shown by the grey dotted line. The gates of the first inverter stage quickly rise above threshold, and a nearly ideal square pulse is delivered to the gate of $M_\mathrm{LED}$, the driver MOSFET to the light source, as can be seen in the temporal zoom of the LED currents of Fig.\,\ref{fig:apx__transmitter_time_traces}(f). It is for this reason that the two-inverter digital circuit is employed. After the voltage to the first inverter drops below threshold, $M_\mathrm{LED}$ quickly shuts off, and the voltage across the LED block decays with the RC time constant of that circuit. The resistance, $r_\mathrm{LED}$, has been chosen to establish a 100\,ns time constant, and the actual value in a fabricated circuit will need to be investigated. 

In the temporal zooms of Fig.\,\ref{fig:apx__transmitter_time_traces}(e) and (f), the time at which the hTron switches to the resistive state is labeled (15\,ns), and the total delay of the circuit from the time of the resistance to the time that current begins to enter the diode is 5\,ns, with delays accrued due to charging up the various MOSFET and LED capacitances. The MOSFET model treated here assumed a 1\,\textmu m minimum feature size, compatible with the cleanroom at NIST, and shorter delays are achievable with the reduced capacitance of advanced CMOS. Full circuit simulations have not been run for MOSFET models corresponding to contemporary CMOS nodes, but estimates indicate the delay will reduce to 2-3\,ns.

In running these simulations, the number of quantum dots or emissive centers was specified ($N_\mathrm{qd}$), and the width-to-length ratio of $M_\mathrm{LED}$ as well as the channel doping of that MOSFET were iterated to obtain the appropriate current injection to populate that number of emitters. For each value of $N_\mathrm{qd}$, the area of the diode was calculated based on the density of emitters, and from this area the capacitance was approximated using a parallel-plate model. Thus, the capacitance increases linearly with the number of photons required from the source. Figure \ref{fig:apx__transmitter_time_traces}(g) shows the currents in the LED on a fine temporal scale with a logarithmic $y$-axis for four values of the number of emitters ranging from $10^2$ to $10^5$, covering a broad range of neuron types with varying degree of connectivity. The dashed curves at early times in the plot correspond to the current being driven into the LED capacitance ($I_{11}$ in Fig.\,\ref{fig:apx__circuit_transmitter}), while the solid traces at slightly later times correspond to the current into the active region of the LED ($I_{13}$ in Fig.\,\ref{fig:apx__circuit_transmitter}). It is evident that for all cases current must be delivered to the capacitor to bring the voltage across the LED above threshold before current is driven into the diode itself, resulting in approximately 1\,ns of additional delay beyond the MOSFET stages. 

In the phenomenological model that is the subject of this work, the current through the diode is used as an input to the source rate equations. $I_{13}$ from the transmitter model is input as the driving current, $I$, in Eqs.\,\ref{eq:rate_np__qd}, \ref{eq:rate_n}, and \ref{eq:rate_p}. The results of these calculations are shown in Fig.\,\ref{fig:apx__sources}. Figure \ref{fig:apx__sources}(a)-(d) shows results from the quantum dot rate equations, while Fig.\,\ref{fig:apx__sources}(e)-(i) shows results from the emissive center rate equations. In Fig.\,\ref{fig:apx__sources}(a), the current injection pulse is shown, while Fig.\,\ref{fig:apx__sources}(b) shows the ground- and excited-state populations as a function of time. The sum of these two quantities does not return to unity at the end of the simulation because a finite fraction of the dots remains in an excited state with a trapped electron or hole. At low temperature, without a mechanism for decay of these states, the consequence would be that slightly less injected charge is required on subsequent pulses to populate the ensemble of quantum dots with excitons. Figure \ref{fig:apx__sources}(c) shows the populations of electrons and holes normalized to the total number of carriers injected in the simulation. The simulated electroluminescence is shown in Fig.\,\ref{fig:apx__sources}(d). Similar time traces for the silicon emissive center model are shown in Fig.\,\ref{fig:apx__sources}(e)-(i), with the addition of the populations of the non-radiative centers in Fig.\,\ref{fig:apx__sources}(g). The result is very similar to the quantum dot case, except the longer lifetime of the emissive centers results in an electroluminescence signal that extends further in time. In both Figs.\,\ref{fig:apx__sources}(d) and \ref{fig:apx__sources}(i), a histogram of $10^6$ samples drawn from the respective cumulative distributions are shown by the dark blue dots. 

Only the electroluminescence output from the simulations enters the phenomenological model. From the electroluminescence signal, the cumulative distribution is formed, and from this, photon delay times can be sampled. To use these simulations in the phenomenological model under consideration, it is helpful that the four curves in Fig.\,\ref{fig:apx__transmitter_time_traces}(g) are essentially scaled versions of each other. When input to the source rate equations, the output electroluminescence spectrum is very similar. When implemented in the phenomenological model code, a neuron will have a given number of downstream synapses, and a multiplicative factor is specified to determine how many photons are generated when the neuron reaches threshold, which we refer to as $N_\mathrm{ph}$. Each time a neuron spikes, $N_\mathrm{ph}$ samples are drawn from the electroluminescence distribution giving $N_\mathrm{ph}$ values of time delay. These $N_\mathrm{ph}$ values of time delay are randomly assigned across the receiving synapses. Typically, $N_\mathrm{ph}$ is larger than the number of synapses, so for each synapse the earliest time from its list is chosen as the spike time. This behavior is justified based on the binary response of the single-photon detectors \cite{buckley2020integrated} that form the receivers at each synapse.

While Appendices \ref{apx:source} and \ref{apx:transmitter} have been somewhat involved, the take-away message is simple: when a neuron reaches threshold, its synapses all receive spike events with times randomly sampled from the numerically determined probability distributions. The probability distributions depend on which light source is chosen, with the primary difference being the exponential decay time constant set by the lifetime of the radiative emission process. This transmitter circuit and source design completes the phenomenological treatment. The phenomenological model of the computational circuits---the dendrites, synapses, and somas---is given by a leaky-integrator equation, Eq.\,\ref{eq:main_equation__ode}, with a nonlinear driving function, $r(\phi,s;i_b)$, and a simple form for coupling between dendrites, Eq.\,\ref{eq:main_equation__coupling}. The phenomenological model of communication between neurons is simply a probabilistic delay time drawn from a specified distribution obtained through circuit and source simulations. 

\section{\label{apx:nine_pixel}Nine-pixel drive matrices}
The full set of 30 inputs to the nine-pixel image-classification task are shown in Fig.\,\ref{fig:apx__nine_pixel_drive_matrices}. The images are generated by starting with the ideal letter instances for $z$, $v$, and $n$. Subsequent variants are generated by letting one pixel at a time switch its state.

\bibliographystyle{unsrt}
\bibliography{phenomenological}

\end{document}